# Autonomy and Unmanned Vehicles

**"**Augmented Reactive Mission–Motion Planning
Architecture for Autonomous Vehicles**"**

# Authors:

**1. Somaiyeh MahmoudZadeh**

Faculty of Information Technology, Monash University, VIC 3800, Australia
Email: Somaiyeh.mahmoudzadeh@monash.edu

**2. David M.W. Powers**

School of Computer Science, Engineering and Mathematics
Flinders University, Adelaide, SA 5042, Australia
Email: David.powers@flinders.edu.au

**3. Reza Bairam Zadeh**

Fleet Space Technology, Adelaide, SA 5009, Australia
Email: Reza.bairamz@fleet.space

# Editors:

# Preface

Advances in hardware technology have facilitated more integration of sophisticated software toward augmenting the development of Unmanned Vehicles (UVs) and mitigating constraints for onboard intelligence. As a result, UVs can operate in the complex missions where continuous transformation in environmental condition calls for higher level of situational responsiveness and autonomous decision making. This book is a research monograph that aims to provide comprehensive survey of UVs autonomy and its related properties in internal and external situation awareness toward robust mission planning in severe conditions. An advance level of intelligence is essential to minimize the reliance on the human supervisor, which is a main concept of autonomy. A self-controlled system needs a robust mission management strategy to push the boundaries towards autonomous structures, and the UV should be aware of its internal state and capabilities to assess whether current mission goal is achievable or find an alternative solution. In this book, the AUVs will become the major case study thread but other cases/types of vehicle will also be considered. Indeed the research monograph, the review chapters and the new approaches we have developed would be appropriate for use as a reference in upper years or postgraduate degrees for its coverage of literature and algorithms relating to Robot/Vehicle planning, tasking, routing, and trust.

To share learned lesson and highlight some of the latest advances in AUVs and Unmanned Air vehicles (UAVs) as well as emphasize on some foreseeable direction, this book will take an overview on UVs mission planning and management systems introduction in Chapter 1. Accordingly, various methodologies will be discussed and worked out in the literature of Chapters 2 and 3 in two disciplines of UVs mission and motion planning.

A certain degree of autonomy is demanded for an AUV to fulfill any particular underwater mission objectives and ensure vehicle's safety in all stages of the operation. Although a majority of today's AUVs are competent to carry out unsupervised missions, in most of the cases the operators remain a close-up control on vehicle's operation, which is an expensive process and infeasible in some situations. As regards, existing AUVs are associated with battery capacity and endurance restrictions, which entails having a systematic mission time, task and resource management strategy to perform persistent deployment in longer missions. Another critical concern of mission success is safety and reliability of a vehicle's deployment. In this respect, having an efficient motion planning system facilitates the AUV to cope with marine uncertainties and sudden environmental changes.

The idea of Chapters 4 to 7 is about designing a new reactive control architecture to cover both requirements of higher level decision autonomy and providing a safe manoeuvre in the face of periodic disturbances in a turbulent and highly uncertain environment. It can be interested for its comprehensive and new perspective to the autonomous vehicle's real-time decision making in critical situations. The proposed framework encompasses a reactive execution layer for decision making and a deliberative action generator that can simultaneously plan the complete mission and carry out a fruitful mission while allowing higher degrees of efficiency in propulsion and dynamic maneuverability. Parallel execution of the higher and lower level systems accelerates the computation process, in which total operation time is in the range of seconds that is a remarkable achievement for such a real-time system. The main reason for the prevalent performance of this system is the mechanism of mixing and matching two different strategies and their accurate synchronization. A significant benefit of such modular model is that the modules can employ different methods or their functionality can be upgraded without manipulating the system's structure. This advantage specifically increases the reusability and versatility of the control architecture and eases updating/upgrading its functionalities to be compatible with different applications, and particularly applicable for other platforms of autonomous systems. The operation diagram and schematic of the framework's different components are represented in detail using figures, tables, and flowcharts.

# Acknowledgements

This book arose from the outcome of the previous works initiated by the first Author *Somaiyeh MahmoudZadeh* and had been directed and supervised by the second author *David M.W Powers*. I would like to thank the third author *Reza Bairam Zadeh* for his technical support and guidance in the preparation of this book.

I would like to express my sincere gratitude to *Adham Atyabi* for his continuous support of related research without whose efforts this book could not have been written. My appreciation also extends to my family for their wise counsel and supports through my life.

The writing team has been expanded the content of this book to induct benefits of our previous works in the scope, which has been written in an appropriate order to tell a coherent story to a more general audience. This book does not contain any material previously published or written by another person except where due reference is made in the text.

Melbourne, 21 June 2018

Somaiyeh MahmoudZadeh
David M.W Powers
Reza Bairam Zadeh

# Contents

## Chapter 1

## Introduction to Autonomy and Applications

S. MahmoudZadeh[1], A. Atyabi, D.M.W. Powers, R. Bairam Zadeh

[1] Faculty of IT, Monash University, Clayton, VIC 3800, Australia
Email: Somaiyeh.mahmoudzadeh@monah.edu

**Abstract.** Increasing the level of autonomy allows reducing reliance on the human supervisor. Addressing autonomy in the real world with unknown events, it is strongly critical to instantaneous adaptation to the continuously changing situations. Autonomous adaptation relies on the understanding of the surrounding environment. Complicated missions that cannot be accurately defined in advance will need to be resolved through intermittent communication with a human supervisor. This will restrict the applicability and accuracy of such vehicles. A fully autonomous vehicle should have capability to consider its own position as well as, its environment, to properly react to unexpected or dynamic circumstances. This chapter aims to provide a general background of some Unmanned Vehicles (UVs) and existing difficulties on the way of having a true autonomy in their applications. Assessing the levels of autonomy and its related properties in internal and external situation awareness toward robust mission planning are also discussed in this chapter.

### 1.1 Background and Challenges over the Autonomous Unmanned Vehicles

The advancement and application of UVs showed a rapid increase during the last decade especially after recent improvements in the hardware that allowed the incorporation of more complex and resource demanding software that reduced the limitations regarding the level of onboard intelligence. Improvements in UVs' degree of intelligent allowed their use in more sophisticated missions that require a higher level of situational responsiveness in persistently changing environments. The subject of autonomy and mission planning have been comprehensively investigated in various frameworks and different environments over the past decades. Hence, due to the distinct particularity of operational environments, the discussion over the UVs and barriers in achieving sufficient autonomy is worked out separately in two disciplines of underwater and aerial vehicles.

### 1.1.1 Autonomous Underwater Vehicles (AUVs)

Autonomous robotic platform has become increasingly popular. AUVs are considered as part of a wider group called Unmanned Underwater Vehicles (UUVs), which also includes non-autonomous, semi-autonomous and remotely controlled vehicles. The earliest AUV, named SPURV, was designed in 1957 at Washington University for the purpose of simple underwater explorations and acoustic transmission [1]. Previously, AUVs were only able to proceed limited number of dictated tasks. With advancement of high yield energy supplies and conducting powerful processors facilitates today's AUVs to handle more complex tasks and missions. Numerous types and classes of these vehicles have been designed over past decades for different purposes and they are constantly evolving over time. Their size is varying from portable types to larger diameters (e.g. over 32 feet length), where each class of these vehicles have their own advantages and applications. Larger vehicles have stronger sensor payload capacity longer endurance, while smaller vehicles are advantageous to lower logistics. As AUVs have proven their cost effectiveness, they are widely used in underwater exploration up to thousands of meters, far beyond what humans can reach. The following is one of the most recent sample of an AUV developed by a Norwegian team for detecting gas leaks and chemical underwater discharges in a cost-effective way [2].

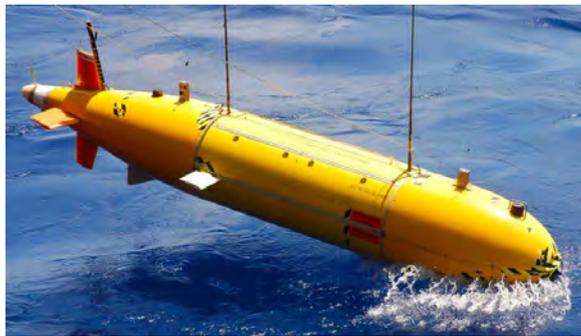

**Fig. 1.1.** An autonomous underwater vehicle just prior to launch [2].

Nowadays, AUVs serve different purposes in the areas of commercial offshore, military and scientific approaches such as underwater scientific exploration, coastal areas monitoring, turbulence measuring, and offshore mining [3]-[5]. The oil industry applies AUVs for detailed seafloor map-ping prior to developing subsea infrastructure. Afterward, the produced infrastructures and pipelines can be also installed by these vehicles in the most cost-effective way with minimum disruption to the environment. The survey companies also take advantage of AUVs to carry out precise surveys of undersea where the conventional bathymetric surveys tend to be too costly or less effective. Despite the capabilities of extended cost-effective operations, AUVs function is restricted by their limited autonomy and low operational bandwidth [6]. Due to these reasons, the only possible time for them to exchange the data and communicate to an operator is before and after a mission. AUVs failure is inadmissible due to expensive maintenance. Thus, an AUV needs to occur higher levels of intelligence to carry out complex missions efficiently and to be reliable for unsupervised operations, where prompt reaction to the raised changes is necessary in unknown and uncertain environment [7]. Accurate awareness of the plausible environmental situations and making efficient decisions are key properties of autonomy.

Time and battery restriction is another critical challenge for AUVs' operations, which is even further problematic in long-ranged missions and complex mission scenarios. Current vehicles have limited energy supplier and confined endurance. Therefore, they should be intelligent enough to wisely manage available resources and persistently deploy in more extended diverse missions [7].

### 1.1.2 Unmanned Aerial Vehicles (UAVs)

UAVs are a type of aircraft operating without a pilot and usually called drone. There is another similar group known as Unmanned Aircraft System (UAS), which operates by a ground-based controller. Similar to what we discussed about AUVs, the UAVs with onboard processors also appear with different levels of autonomy. These group of vehicles are beneficial for dangerous mission that are too risky to include human pilot. UAVs extensively employed for various purposes such as recreational, commercial, agricultural, surveillance, military, or other applications. The following is one sample of a great variety of these vehicles called "Tiburon", which is a modifiable drone made for scientific applications. This vehicle is designed to investigate the changes of Antarctic ice shelf over the time, which is a critical concern for climate change and global warming [8].

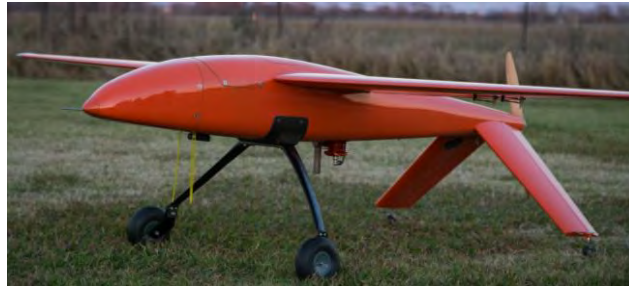

**Fig. 1.2.** Sample of a scientific UAV developed in the center of "Houston-based company Intuitive Machines" [8].

Employing UAVs over alternatives offers many advantages such as: boosting flight performance, cost reduction, and applicability in hazardous and risky missions [9]. Today's technological improvements in the scope of UAVs have propelled progression towards autonomous structures with a higher level of onboard intelligence requiring less human involvement. However, different concepts can be conceived for 'intelligence', which is hard to measure and formalize. Meystel and Albus [10] used the following definition: "*Intelligence is the ability of a system to act appropriately in an uncertain environment, where an appropriate action is that which increases the probability of success, and success is the achievement of behavioral sub-goals that support the system's ultimate goal.*"

Veres et al. [11] proposed five fundamental features to achieve autonomy in UVs, in which the first four items are also shared by remotely controlled vehicles:

   1) Structural hardware,
   2) Efficiency of energy suppliers,
   3) Sensors and actuators,
   4) Computing hardware,
   5) Autonomous software,

Ideally, autonomous systems should be independent of human assistance, and they should operate relying on their own collected sensory data. However, today's UAVs still have a long journey to be fully autonomous and reliable. To have a fair comparison of proposed ideas over the autonomy and intelligence, and for assessing the levels of UVs autonomy having a common and standard base is necessary.

## 1.2 Automation vs. Autonomy

There is a considerable difference between concept of autonomous and automatic operations. In automatic systems, the vehicle/machine precisely executes the pre-programmed commands without any functionality for choosing or making decisions, while autonomous systems are capable of recognizing various circumstances and making a decision respectively. Therefore, advancing the level of intelligence for autonomous systems is a fundamental requirement facilitating them with the ability to reconfigure on diverse situations and autonomous mission planning/re-planning under the new circumstances, which is the area of interest for many researchers in the field and system designers [12]. Modern flight control systems have taken benefits of automatic models, which play an important role in elaborating them in terms of comfort, efficiency, and safety of the motion; however, it is different with what we know as autonomous systems. The following complementary explanation is proposed by Stenger et al. [13] to distinguish the difference between these two categories: "*an automatic system is designed to fulfil a pre-programmed task. It cannot place its actions into the context of its environment and decide between different options. An autonomous system, on the other hand, can select amongst multiple possible action sequences in order to achieve its goals. The decision which action to choose is based on the current knowledge, that is, the current internal and external situation together with internally defined criteria and rules.*"

Cognitive systems are another prospective to analyse and explain autonomy. In this respect, the dependency of autonomy and cognition is explained by Vernon [14] as "*One position is that cognition is the process by which an autonomous self-governing agent acts effectively in the world in which it is embedded. As such, the dual purpose of cognition is to increase the agent's repertoire of effective actions and its power to anticipate the need for future actions and their outcomes*".

Following these principles, the ability of sensing, monitoring, comprehending of operational contexts and properties, and probable circumstances are substantial requirements for an autonomic system. An autonomous UV should constantly adapt to a continuously changing environment independent of human involvement. This level of consciousness strictly tightened up to the definition of Endsley for situation awareness [15] (explained in Sec. 1.2.1).

### 1.2.1 What is Situation Awareness?

The ability of comprehension and managing to deal with highly dynamic and uncertain conditions is characterized as Situation Awareness (SA), and it is a substantial necessity for frameworks that are committed to responding to dynamic and time-varying conditions [15]. SA outlines a way toward sensing, detecting, comprehending and operating in partially known or unknown environments. Enhancing the SA level of UVs can advance their capacities from full human control to completely self-governing (autonomous) control [16]. Unforeseen and uncertain circumstances can enforce a mission to terminate before completion and in worst cases may even result in loss of the vehicle, as happened for Autosub2 that was lost under the Fimbul ice shell in the Antarctic [17].

The present condition of a vehicle and substantial incoming load of surrounding information should be incorporated and considered simultaneously to form an appropriate mental model of the existing situation. This amalgamated picture creates the central organizing pattern from which all decisions and action selections take place. SA along these lines incorporates three parts of observation, perception of the surroundings; comprehension and understanding of the phenomenon/events; and projection of plausible events that may occur in the future (see Fig. 1.3).

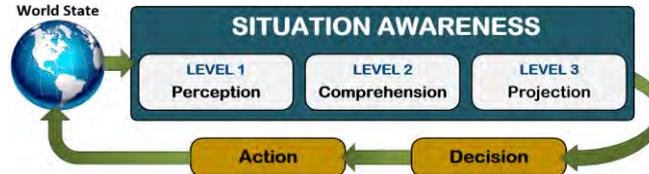

**Fig. 1.3.** Endsley definition of SA steps and decision making [16].

Framework's impression of surroundings can advance its ability to manage unstructured and unexpected occasions. Accordingly, objects, environmental components and their characteristics should be processed simultaneously through pre-attentive sensory supply. Recently perceived data is aggregated with present knowledge in the working memory to form a new updated mental image of the evolving circumstance. These chunks of data utilized to make projections of what may occur in the future. These predictions, thus, enable the UV to choose what actions (ongoing and real-time) to take as a response. SA, therefore, is a standout amongst the most basic required components for advancing the upcoming classes of AUVs and UAVs.

A semantic world model framework is introduced by Patron et al. [18] to improve the global (system level) and local (agent level) SA through a hierarchical illustration of the extracted sensory knowledge. This mechanism applied a declarative goal-based mission planning approach that could improve the mission parameterization, execution, handling the internal issues, and dynamic adaptation relying on existing knowledge of platform's capabilities. Later on, Patron et al., [19] consolidated the goal-based mission planning and knowledge-based structure to furnish agent oriented embedded decision making. They have proposed adaptive planning mechanism through the autonomous coordination of agents. This structure has specifically influenced by the knowledge representation schema and proper distribution of information among embedded agents, which provided an improved SA for interoperability of the agents in autonomous platforms. The outcomes acquired particularly impacted mission flexibility, robustness, and autonomy.

In later years, another hierarchical ontological approach has been proposed by Patron et al., [20] to build a platform capable of autonomous re-planning of missions to adapt new circumstances during the operation. A goal-oriented system was conducted for parameterizing a mission based on available knowledge and vehicle's capabilities, and a semantic knowledge representation framework is designed as the core of the architecture for improving the overall SA of the UUVs' service-oriented agents.

A new framework of SA is designed by Chai and Du [21] comprising event extraction and correlation, diagnosing the force structure, expectation inference and prediction in which event-rule extraction accomplished according to the expert knowledge system. Respectively, Rashaad et al., [22] introduced a beneficent way to model an actionable SA applying Fuzzy Cognitive Maps (FCM) that embraced all three SA levels of observation, comprehension, and projection. The given SA-FCM cognitive model is created straightforwardly from the objectives, choices, and fundamental information associated with appropriate decision-making in a particular area, which is known as a computational naturalistic decision-making model that replicates human cognition as it relates to SA.

### 1.2.2 What is Cognition?

Cognition is the joined capacity to comprehend and anticipate how the things may plausibly behave or change now or in one step forward, and encountering these intuitions to decide the necessary actions/moves to make. Cognitive systems are categorized into two main groups of the 'Cognitivist' and 'Emergent' frameworks.

Cognitivist frameworks use a symbolic expression of processing information, while the emergent frameworks depend on the self-organizing principles mainly embraced by connectionist, dynamic, and enactive systems. Considering a behaviour-based representation of a system, cognitivist approaches are characterized by sets of actions directed by experts to deal with earlier known conceivable states and sensitive to the inaccurate perception of the environmental situations. On the other hand, emergent models have the potential of learning and advancing behaviours to deal with unpredicted and unforeseen environmental circumstances efficiently. A contained model is proposed by Vernon et al. [23] in which physical instantiation of the framework does not impact the cognition model while the emergent-based cognition models incorporate an essential role for physical instantiation. Figure 1.4 depicts the developmental components of a general cognitive model to indicate that in such a framework, the system behaviours are impacted by perception (guidance), objectives (forming and coordinating), and triggering motives.

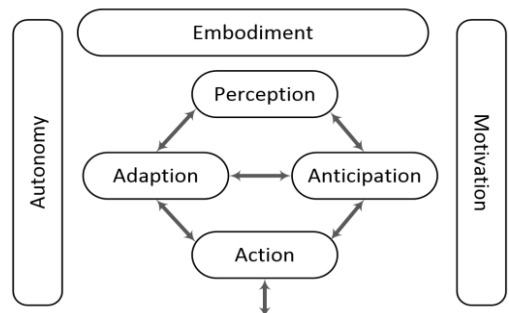

**Fig. 1.4.** Developmental components of cognitive systems [16].

Considering the necessity of predicting outcomes and adapting to dynamic situations, it is noteworthy to mention that the cognition also facilitates the system to form explanations and making assumptions over unforeseen events. This potency helps to broaden the agent's repertoire of actions and improves its capability of interacting with the world around it while enhancing its level of autonomy. Vernon [14] classified some of the existing popular cognitive-based architectures into three main categories:

1) Emergent (Global workspace, Autonomous Agent Robotics (AARs), Self-Aware Self-Effecting (SASE), Self-Directed Anticipative Learning (SDAL), DARWIN)
2) Cognitivist (Executive Process Interactive Control (EPIC), Soar, The ICARUS Cognitive Architecture, Adaptive Control of Thought-Rational (ACT-R), ADAPT),
3) Hybrid (Cog: theory of Mind, HUMANOID, Kismet, Cerebus).

Between all employed cognitive architectures in robotic studies, Soar is the popular one often applied for the purpose of mission management, which is a rule-based system that processing with the two cycles of production and decision. In the production cycle those productions that match the declarative memory contents fires which result in alteration of the declarative memory and further firing of other productions. In the decision cycle, a single action is chosen out of all possible actions. Soar uses a sub-goal mechanism whenever an impasse situation occurs. The impasse is the situation in which the actions are ambiguous, or no action is available. In the sub-goal mechanism, the impasse is resolved by setting the new state in a new problem space. This change of state in Soar system is the only form of learning that exists in this architecture [14].

Enactive structures are another type of Emergent models that enhance emergent capacities by assuming cognition as a procedure whereby necessary requirements for the persistent existence of a cognitive agent are provided [23], [24]. In enactive models, having a symbolic representation is not an essential requirement as there is no obligation to have access to present information or rule and the system operates using an enactive interpretation. For example, considering a UV in the context of cognitive robotic systems, the main idea of enaction that the vehicle builds up its mental picture of its surrounding world by interacting the environment. Hence, enaction enables the UV's autonomous operation according to its mental model of how the world works. The followings are the essential elements when dealing with enactive systems [16]:

*i.* ***Autonomy:*** Survivability is a critical concern in developing UVs, which entails self-protection and recovering from faults (if possible) in a diverse environment, which improves performance of the operations. The UVs are associated with some organizational self-maintaining characteristic similar to intelligent creatures, which are using their own capacities to manage their interactions with the environment to survive. In other words, the system tends to be entirely self-regulating and self-governing, so that takes actions that best benefit their own requirements and the swarm's objectives. Therefore, the system is fully self-controlled, allowing independent operations. It does not mean that the system is not influenced by its surrounding world, but rather these influences are brought about through interactions that do not threaten the autonomous operation of the system.

*ii.* ***Embodiment:*** In the enactive cognitive category, the system needs to physically exist in the world and interact with the environment and its included objects (animate or inanimate, cognitive or not), while influ-enced by and react to these stimuli. In such a system, different types of embodiment can be considered including Structural Coupling (two-way perturbation of a system and the environment), Organismic (autopoietic living systems) embodiments, Historical (history structure coupling), Physical (capability to have forcible action), and Organismoid (for humanoid or rat-like robots).

**iii. Emergence:** The emergence addresses the way cognition arises within the individual and team of UVs. This element refers to the mechanisms and set of rules that govern the behaviour of the components of the UV(s). The behaviours (cognitions) in the emergence system arise from the dynamic interplay of defined rules and interactions with the other components of the cognitive system. It means the internal dynamics (as the essential requirement for maintaining the system's autonomy) emerge from the controlling behaviours, where the objective is to develop an architecture with a multi-layered hierarchical level of autonomy to facilitate the sense-making requirements and emergence in a complex cognitive system.

**iv. Experience:** It refers the UVs history of interactions with the surroundings, where these interactions trigger structurally determined state changes rather than controlling the system. These changes rely on the system's structure regarding the embodiment of the self-organizational principles that are essential for the system's autonomy.

**v. Sense-making:** It refers to the relationship between the knowledge encapsulated by the UV and the interactions which gave rise to it. The system provides emergent knowledge to capture some principles or lawfulness, while the sense it makes depends on the form of interaction (its actions and its perceptions of the environment's reaction).

The level of autonomy can be assessed in different ways according to the expectation of the vehicle's performance in high and low-level operations.

### 1.2.3 Assessment of the Autonomy Levels

The difference between intelligence and autonomy is the common confusions in autonomous systems that has been debated by many researchers. Intelligence can be defined as the ability knowledge discovery and using that to do something. For an agent, autonomy can be introduced as the agent's capability of generating its own goals and objectives without requiring outside instructions. Clough [25] explained it by stating: "*They are* *not the same. Many stupid things are quite autonomous (bacteria) and many very smart things are not (my three year old daughter seemingly most of the time)*". According to Clough discussion, the important criteria for assessing a vehicle's autonomy is how well the vehicle carries out the associated tasks toward its goals without operators involvement in monitoring/supervising the moments of a mission. Hence, it might be conceived that the system's intelligence is not important as long as it is capable of performing the assigned tasks. Thus, the autonomy and its level can be investigated from different perspectives.

Nowadays, there are still highly autonomous platforms relying on levels of experts' observation, knowledge, and control in handling complex approaches. To make a vehicle genuinely autonomous and reliable, more advanced SA is demanded [15]. For a successful mission and maximize mission performance, the autonomous system should continuously monitor its resources and compromise on tasks' priority/risk and vehicle's constraints. Advancement of the autonomy can improve UVs' operation in the following approaches:

- **Optimizing the energy consumption:** this can include improving the power sources and reducing vehicle's power consumption to accommodate long endurance missions.
- **Optimal risk-free navigation:** advancing the navigation system with accurate localization and minimum position calibration error.
- **Accurate and robust decision-making:** improving the capacity of sensing, interpreting, acting/reacting to unexpected environmental changes, and making proper decisions when facing various challenging situations.

Autonomous decision-making and navigation accuracy are related by various standpoints. One essential concern for autonomy is advancement of navigation system (including path/trajectory planning) to be robust to the strong environment variability. Chapter 3 will provide a comprehensive overview on the state of the art in navigation improvement. Some of the most well-known autonomy assessment methods for UAVs are discussed in the following subsections.

### i. Mobility, Acquisition, and Protection (MAP) Method

This method of assessment has been developed by Mark Tilden at Los Alamos National Lab [26] and considers mobility, acquisition, and protection as the factors of UAV's autonomy level assessment. To be more precise, to address mobility, $M_0$ reflects no motion ability and $M_x$ representing the ability to move in $x$ dimensions (with maximum $x$ of 5); acquisition in-cludes the ability to extract, store, and utilize energy; and protection in terms of self-defend-ability against hostile stimuli. This autonomy classification is more desirable for army use due to employing a radar view interface. However, it has restrictions in dealing with multi-UAV scenarios due to the following reasons:

- Weak discrimination between various levels of autonomy.
- It is not able to handle multi-UAV's interactions.
- It is not able to address operational characteristics of multiple UAVs.
- The acquired fixation on some of the vectors in the radar view caused by the assessment of multiple UAVs at the same time.

### *ii.* **Draper Three Dimensional Intelligence Space**

This metric is introduced by Stark's metric at the Draper Laboratory [27] to evaluate and measure the level of autonomy and intelligence. The Stark's metric provides useful features such as three metrics allowing radar view presentation of the outcomes while taking the operational issues into account for the assessment procedure. This measurement is applicable to the multi-UAV autonomous control system; however, it has some other restrictions including:

- The poor resolution of the metric causes max-out in the radar view for multi-UAV scenarios.
- SA is quantified and measured according to the number of sensors in the system, and then a fusion method is applied rather than evaluating whether the system is capable of understanding the on-going surrounding events, which cause the changes in sensors' outputs.

### *iii.* **Autonomous Control Level Chart (ACL)**

Another taxonomy/classification is introduced by Clough [25] to measure the autonomy of UAVs, which includes 9 autonomy levels in which radio-controlled drones are considered as the lowest autonomy level (level 0). Each higher level achieved through the addition of an extra set of functionalities to the previous level, and the highest level (level 9) corresponds to the fully autonomous system. These functionalities are listed as follows:

- Executing a pre-planned mission (Autonomy level 1).
- Switching between a set of pre-generated mission plans concerning the UAV's SA (Autonomy level 2).
- Adding sophisticated capabilities such as contingency management and fault mitigation for single UAV scenarios (Autonomy level 3-5).
- Developing the cooperation and coordination capabilities.

### *iv.* **Sheridan Scale for Autonomy**

Sheridan [28] provided a new scale metric with ten levels of autonomy assessment in which the lowest level corresponds to an entirely human-controlled entity, and the highest corresponds to a fully intelligent entity that operates without human supervision, detailed as follows:

- Fully controlled by an operator.
- The computer suggests action alternatives.
- Computer narrows down the choices.
- The computer offers a specific action.
- The computer executes that action upon the operator's approval.
- The operator has restricted time to decline the computer's decision before its automatic operation.
- The computer performs commands automatically; afterward, the operator will be informed.
- When the automatic execution completed, Computer notifies the operator upon a request.
- Computer notifies the operator after automatic execution based on its own decision.
- Entirely controlled by the computer (ignoring the operator).

Further to the mentioned measurement metrics, Veres et al. [11] introduced a simple 3 level of autonomy scale, where the first level refers to the vehicles capable of tracking their self-generated trajectories to particular points; the second level includes vehicles capable of navigating towards self-defined intermediate points while the human guidance is limited to defining the global targets; and the third level refers to those vehicles with capabilities of two previous levels along with ability to onboard knowledge processing, receiving mission goals from the operator, and performing mission-related decision makings. This metrics is very general, and it is not sufficient for accurate classification of various existing autonomous systems.

### *Example of Measuring the Autonomy in a Simplistic Scenario*

Generally, autonomy for UVs can be considered in three primary levels of fully human-operated, semi-autonomous and fully autonomous systems. A hypothetical case study is being presented here for better understanding of the distinction between various autonomy measures. Consider a simple scenario where a UUV is requested to travel from waypoint *a* to *d* while visiting *b* and *c* on its path. The vehicle should also deal with some obstacles and harsh water current in its way. The vehicle's performance is discussed below concerning the given three levels of autonomy:

**Human operated UUV:** In such a system, all decisions are taken by an expert operator. The whole mission is monitored and controlled by a human, concerning the given inputs, previous experiences, and primary knowledge of the environment. The operator drives the vehicle through the points *a*, *b*, *c*, and plans to reach point *d* before running out of battery. In a case of occurring any unforeseen situation (e.g., facing a buoyant obstacle or water turbulent), the severity of the condition is identified by the operator, and then he/she decides how to guide the vehicle toward point *d* to cope with the raised situation safely. In these systems, the operator specifies high-level mission plan, and the vehicle accommodates inclusive mission-related information. One of the main disadvantages of the humanoid systems is that a human may get easily distracted due to tiredness, miss precaution, and vagueness or improper cognitive capabilities.

**Semi-autonomous UUV:** this vehicle is advanced with SA and capability of path planning. It operates under human expert's oversight and the operator can interrupt when the vehicle is unable of performing contingency management, path re-planning or task re-scheduling. In the given simplistic scenario, the UUV's navigation system indicates the intermediate waypoints, and it is capable of recognizing the situation, path planning between points, autonomous departure/return, and collision avoidance. However, the operator re-plans the mission scenario and decides how to re-arrange order of visits to the waypoints to compensate the lost time.

**Fully autonomous UUV:** The vehicle can locate its position on a map and generate its trajectory to the assigned waypoints. The system receives a prior knowledge of environmental information including candidate sequence of waypoints, and the system provides trajectories to safely drive through the points and reach the destination on-time concerning its available resources. The spent time to reach a point is compared with the expected time and the system re-plans its mission if the UUV is behind its predetermined schedule. The identical aspect of this system is that the vehicle is capable of independent operation as it is advanced with a high-level SA, real-time path planning, the ability of contingency management, and task scheduling.

After all, the existing mechanisms for measuring and validating the level of autonomy, in general, is unreliable or inapplicable for all ranges of vehicles. Such measures are not adequate to capture multi-facet nature of autonomous systems, mission scenarios. This insufficiency is clearly reflected in 2012 final report of the task force to DoD [29], which states:

*"The Task Force reviewed many of the DoD-funded studies on "levels of autonomy" and concluded that they are not particularly helpful to the autonomy de-*

*sign process. These studies attempt to aid the development process by defining taxonomies and grouping functions needed for generalized scenarios. They are counter-productive because they focus too much attention on the computer rather than on the collaboration between the computer and its operator/supervisor to achieve the desired capabilities and effects. Further, these taxonomies imply that there are discrete levels of intelligence for autonomous systems, and that classes of vehicle systems can be designed to operate at a specific level for the entire mission. These taxonomies are misleading both from a cognitive science perspective and from observations of actual practice. […] In practice, treating "levels of autonomy" as a developmental roadmap has created a focus on machines, rather than on the human-machine system. This has led to designs that provide specific functions rather than overall resilient capability."*

## 1.3 Summary of Chapter

Instant adaptation to diverse real-world situations and unexpected events is essentially demanding for improving the level of autonomy. Autonomous adaptation depends on vehicle's perception of the surrounding environment. Some complex missions with sophisticated requirements cannot be precisely specified in advance and will need to be settled through intermittent interaction with a human supervisor. Apparently, this will make some restrictions on the concept of autonomy. A fully autonomous UV is expected to have the capacity of understanding its situation as well as its surroundings, to perform an appropriate response to unforeseen circumstances. To update UVs' SA, an extensive number of sensors and data analysis tools are required. Conceptually, decision autonomy can be divided into two categories: The capability to cover interior malfunctions (sustainability), and the capability to handle probable outside happenings (adaptivity). Experience can also result in a particular level of automaticity in mental processing of situations that lead automatic reaction in facing a similar situation to previous experience. The pattern-recognition and automatic action-selection can be considered as primary steps to mimic human cognitive strategy achieving fully autonomous self-controlled systems, which can also improve SA for more demanding tasks. Humans have the general cognitive capability to concentrate on the relevant events and ignore the others. This chapter provided some information over the definition of autonomy and how it is related to vehicle's cognitive system and its awareness of the various situations. The difference between concept of autonomous and automatic operations, along with some previous attempts over determining the levels of autonomy for various UVs are emphasized in this chapter. Two next chapters will be precisely discuss on benefits and defects of an extensive number of research over the UV's mission planning/management, task scheduling approach, and motion planning strategies, which are considered as the main aspects of a UV's autonomy.

# Chapter 2

# State-of-the-art in UVs' Autonomous Mission Planning and Task Managing Approach

S. MahmoudZadeh[1], A. Atyabi
[1] Faculty of IT, Monash University, Clayton, VIC 3800, Australia
 Email: Somaiyeh.mahmoudzadeh@monah.edu

**Abstract.** The purpose of this chapter is to review some of the recent advancements in the autonomous mission management and mission planning systems in UV studies concentrating on UAVs and AUVs individual and swarm operations. In recent years, increasing attention has been concentrated on extending the ranges of missions and UVs' endurance, increasing UVs' applicability, improving vehicles' autonomy to manage longer missions without human guidance, and decreasing operating costs and many other aspects of autonomy [1]. Apparently, having the common ground condition(s) is an essential requirement for comparing or judging the level of autonomy achieved by various studies in the scope. A standard autonomous mission management system involves some components such as:

- Mission planning component, which includes task/resource allocation mechanism, action prioritizing process, and planning a general overview of the vehicle(s) motion in the terrain.

- Mission execution, which includes procedures such as navigation, trajectory planning, task execution, intelligent action selection.

- Mission monitoring including SA, mission progress evaluation, and anomaly detection.

- Mission re-planning procedures such as re-tasking, resource reallocation, reprioritizing and re-routing.

## 2.1 Recent Achievements in UAV's Mission Management Systems

Based on Johnson et al. argument in [2] the existing UAV's do not fully address the safety requirement for operating in populated and civilian airspaces. The research bolds out existing shortcomings associated with current UAV technologies such as vehicles intelligence and perception, con-current adaptivity, and reflexive reaction resulting in a poor performance compared to the piloted systems. Although a failure or inefficiency might be acceptable in some cases given that there is no loss of human life, this can restrict the application of this technology to non-populated areas. They investigated the reasons of mission failure in UAVs operations and reported most of the problems occurs due to shortcoming in testing (16%), evaluating components quality/suitability (16%), faults in software configurations (13%), design and assembly errors (9%), poor emergency procedures (26%), and redundancy of critical systems (10%). The authors believe it is feasible to reduce almost 30 percent of errors by improving mission management system. In this regard, Johnson et al. [2] introduced an autonomous mission management structure called TRAC that facilitated UAVs with a robust software architecture to monitor vehicle's current situation, predict its future states and to deal with the present and probable problems. The TRAC architecture is engaged with the stepwise managing of the real-time mission plan execution through the following functionalities:

- Generating commands for each mission segment;
- Tracking the segment completion;
- Responding to unforeseen events in a mission segment;
- Managing mission segment's start and termination;
- Recording (maintaining) new experiences and mission events;

An Intelligent Mission Management (IMM) strategy has been developed by Sullivan et al. [3] advancing UAVs goal-directed autonomy, allowing the vehicle to redirect its path according to the variation in environmental conditions and mission goals. The proposed IMM method was helpful in reducing the mission risk and improving UAVs chance of success. To this end, a ground-based Collaborative Decision Environment (CDE) was employed aiming to provide collaboration, SA, and decision-making tools and to improve the capacity of planning, schedule monitoring, integrating sensing and visualization. An automated 'mixed initiative activity planning generator' has been employed as a ground-based decision support system allowing the operators to monitor the automatically generated plan and to interrupt in cases of emergency to maintain the mission within the defined boundaries and available resources. This package corresponds to level 6 of Sheridan's scale of autonomy[1].

In mission planning, the significance of contingency management was discussed by Frank et al. [4] who argues this term as compensation and recovery capabilities from consequences caused by unpredictable circumstances and conditions possessing potentials to impact plan implementation negatively. The study introduces the holistic contingency management's key components including multi-level assessment; plan-based assessment; predictive assessment; capability-based assessment; and team-based assessment as well as a comprehensive view over contingency management in 2 to 3 structural layers. The comprehensive view, in one dimension, contains short, medium, and long-time span; additionally, five levels of mission management functions are presented, including closed loop response, reflexive/reactive response, and plan-dependent monitoring.

---

1. Refer to Chapter 1 for Sheridan's autonomy assessment metrics

Aligning the operator's conceptualization of the mission planning was discussed as the most challenging obstacle for developing intellectual autonomy by Linegang et al. [5]. Traditionally, the human operators' role includes specifying mission's goals and limitation, reviewing, approving, executing, and overriding the planned actions, in necessary circumstances, for satisfying the mission's goals in both offline and online phases.

A hybrid supervisory control framework, making 3D leader-follower formation control possible for UAVs, were presented by Karimoddini [6]. In addition to utilizing multi-affine functions, the framework uses a spherical abstraction of the state space. The hybrid system, in addition to a logic supervisor used for providing necessary control over the formation of the UAVs during the mission, makes use of a final state model bi-similar to the original continuous-variable dynamic system.

In scenarios based on searching for a point of interest in a time-limited style, a distributed version of self-coordinated and cooperative UAVs was utilized by Guerriero [7]. For optimizing the objectives of the study (maximizing customer satisfaction, minimizing UAVs travel distance, and minimizing the number of utilized UAVs in the mission), two strategies of a rolling horizon and the ε-constraint were considered by the author. In addition to addressing the multi-objective optimization problem by providing a set of ε-constraints for each objective separately at each state/iteration, complete knowledge of the environment, the points of interest, and their related timing factor are assumed by the ε-constraint method. Based on the availability of partial information regarding each event's location and the related time instant, the rolling horizon strategy is designed, and the UAV routing is performed at each iteration, based on the partial information available at that iteration and an optimized route that best suits the objectives of the mission.

To provide on-board time-optimal trajectory management, Keller [8] employed B-spline curves, developed by a set of algorithms transcribing sets of points to a continuous path. Aiming at providing minimum execution time, dynamic adaptation to acquired sensory-feedback, and adherence to mission constraints, the approach post-processes the point-by-point path planner's output to generate minimal representation. The approach extends B-Spline to involve features such as real-time trajectory interruption and redirection.

The networking aspects of mission management were studied by Kopeikin [9], and the importance of being able to maintain the communication, which relays between a fleet of UAVs and the ground station, was highlighted. The networking aspect is handled through distribution task allocation that employs UAVs to satisfy the role of network relayers to maintain communication to the base, which could be facilitated by coupling task assignments and creating relay process, resulting in solutions with the potential of addressing realistic network communication dynamics. The constraints of networking on mission planning were also studied by Manousakis [10] who noted that defining the performance of the required networking is part of the mission planning process. The investigation concentrated on employing a Tactical Information Technologies for Assured Network (TITAN) to manage mission networking requirements and generating maintenance strategies based on network plan and mission goals. In this context, the TITAN role is to dynamically re-plan the network to fulfil the mission objectives in which the process is facilitated using a Mission to Policy Translation (MPT) mechanism that utilizes the existing resources in each stage in a way to satisfy the mission objectives.

A Cooperative Mission Management System (CMMS) with a distributed architecture for a team of cooperative agents, were designed by Xu, Yang and Zhang [11]. The architecture distributes the control sub-systems between the leading agent and the ground control station. The system includes two management layers where mission objectives are assigned to team members by CMMS without causing any conflict with onboard autonomous management system, which its main focus is on resource allocation and scheduling. In the context of this system, the ground control station assigns the mission objectives prior to agents' deployment, monitors agents status and back feeds to operators during the operation. In contrast, the team-leader is in charge of coordinating the sole autonomous management and other team members. Different functionalities for a team of heterogeneous agents such as SA, resource scheduling, mission assignment, team forming, and health management are provided by the proposed CMMS architecture in [11].

A two-layer cognitive structure for resource management has been developed by Boehm et al. [12], where the top-level of the proposed hierarchical controlling system contains a cognitive approach to manage the mission. The system used knowledge-based elements such as goal-driven decision making, inference-based situation interpretation; and pattern-based task execution. Lower layer of the structure facilitated automation by providing the data linkages and route-planning and the model has been tested in a real flight with a fixed wing motor-glider UAV.

Stenger et al. [13] presented another cognitive approach for UAV mission planning, which used Soar-based cognitive architecture to elaborate UAV's autonomous capabilities within a dynamically changing environment. The higher-level planner in this architecture operates based on a familiar environment assumption and use these assumptions to develop the overall mission plan, while the cognitive agent uses the knowledge about the partially known environment to provide the required reactions to unforeseen circumstances. The learning in a Soar architecture occurs based on interaction with the environment that results in evolving actions and requiring a small amount of pre-programmed task sequences, which is a benefit of this system as claimed by the authors. In a simulated environment, under missions with different stress levels, the feasibility of the proposed architecture is evaluated, and the results highlighted the potential of the Soar in addressing the required level of autonomy in UAVs involved in high stressed missions.

The capability of cognitive task/work analysis in development of UAVs autonomy algorithms in dealing with wilderness search and rescue missions (called WiSAR) was investigated by Adams et al. [14]. The WiSAR missions are complex in which a hierarchy of human decision making involves in different stages of the mission. The UAV tends to search the unknown environment for a missing object and back feeds the report or live video to the operator, the expert operator then analyzes signs of the missing object and inform the incident commander about the new evidence. The authors argued that the application of UAV in WiSAR mission cannot turn a manual human-based search process into a fully autonomous operation, but it advances the manual search with a cognitive process. The study suggested combinations of goal-directed task analysis and cognitive work analysis to enhance UAV's autonomy in supporting WiSAR missions.

The problem of task allocating with heterogeneous natures emphasized by Binetti [15], in which some of the tasks are considered to be crucial for a set of heterogeneous agents capable of performing a limited number of tasks. A decentralized critical task allocation algorithm with a three-phase iterative structure was employed, which in the first stage, the algorithm allocates tasks either by forcing or adding, in the second stage it provides procedures for conflict resolution, and in the third stage, it removes tasks from agents that exceeded their task capacity.

The resource management aspects of the mission planning were addressed by Braman and Wagner [16] where the impact of the efficient energy management on human safe operation in space missions has been investigated. The study discussed that how and why space mission planning and monitoring is a complicated procedure as that is dependent on factors such as the way the resources are utilized, the activities to be planned and uncertainty. The authors introduced an energy advisory mechanism providing not only energy efficient plans for space exploration vehicles, but also SA and real-time reports to astronauts and mission operators. Based on the operation plans and the system models, the advisory system predicted the desired energy consumption and based on created energy models it provides the ground operator with a higher level of freedom in safety margins. The mission management becomes even further challenging in the scope of underwater explorations due to severity and extreme uncertainties of the environment and vehicles' communicational and operational restrictions.

## 2.2 Recent Achievements in AUV's and UUV's Mission Management Systems

In recent years, there has been increasing attention to the autonomous application of AUVs to conduct various tasks in the dynamic and continually varying environment, which has led to increasing the range of mission and improving the autonomy of the vehicles to take over more extended mission without supervision [1]. The underwater is unpredictable vast 3D environment rarely populated with obstacles. The difficulties related to such a volatile environment turns out to be significantly more prominent in long-range missions with the extension of the operation zone. AUVs' robustness, in strong environment variability, is the critical factor in mission performance and safe execution, which is hard to achieve due to inadequate information about prior and later condition of such an uncertain environment.

Similar to the UAVs, the mission planning of undersea is usually considered as combinatorial problem of tasking, time management, and/or routing. Each of these sub-problems are analogous to a Non-deterministic Polynomial-time (NP) hard problem of dynamic knapsack or traveller's salesman problems [17, 18]. To deal with the NP-hard complexity of this problem, Yan et al. [19] designed an integrated method combing a velocity synthesis approach with the branch and bound algorithm for mission planning in the form of targets assignment and routing along the assigned targets, in which reducing the total energy cost in presence of ocean stream was the main focus of this research. A Lagrangian-based K-tree approach
has been applied by Martinhon et al. [20] for route planning in an undirected graph-like operation field. Even though the study used linear programming for variable estimation to reduce the mission costs, the approach
does not appear competitive in computational time and cost evaluation.

A new adaptive track-spacing algorithm was developed by Williams
[21] to provide optimal routing for an AUV in mine countermeasures missions and maximizing underwater mines detection. This approach has not
incorporated with any on-board processing capability, which is a drawback
for adaptivity of a routing system to the dynamic changes.

Chow [22] suggested K-means method to tackle AUVs' task assignment problem, and discussed on how considering the dynamics of the
ocean stream can reduce multiple AUVs travel time. For route planning
and controlling the AUVs to steer along the planned direction, the proposed approach combined AUV dynamic model with adapted Dubins
model as well as ocean current factor. However, the approached is implemented in a 2D environment, which tends to be insufficient for resembling
the AUV's motion with six degree of freedom and 3D ocean properties.
Also, static ocean current has been considered in this proposal which may
contradict with the real environments in the ocean.

The large-space operations overwhelm with excessive computational
burden that increases exponentially with the problem size. This is often a
problematic issue for the deterministic methods such as Mixed Integer
Linear Programming (MILP) that is proposed by Yilmaz et al. [23] for
governing multiple AUVs. Although the deterministic strategies tend to
give better quality solutions, they are computationally expensive and inappropriate in the real-time handling mission updates in on-board routing
systems (which all calculations take place in embedded processor(s)) [18].

Meta-heuristics are another approach to these problems, which fast
computation in producing optimal or near-optimal solutions is one of the
outstanding characteristics of these methods. The meta-heuristics also offer
better scalability with the large sized problems compared to the deterministic methods, and their advantages on vehicle routing and task assignment
have been investigated vastly in recent years. Sharma et al. [24] compared
the performance of Dijkstra, as a deterministic method, and Genetic Algorithm (GA), as a meta-heuristic approach, in addressing a vehicle routing
problem in a graph like operation field. The evaluations stated that the
same solution had been produced by both algorithms but with different execution time as GA tended to be faster than Dijkstra in the same examination.

Another attempt to deal with the AUV's routing, task assignment, and risk management joint problem has been carried out by MahmoudZadeh et al. [25], which the heuristic search nature of GA and Particle Swarm Optimization (PSO) has been investigated in finding optimum sequences of to-do tasks and waypoints guidance in a large-scale static terrain. Later on, the approach was integrated to be applicable on semi-dynamic networks taking the use of Biogeography-Based Optimization (BBO) and PSO algorithms [26]. The merit of the proposed solutions by [25, 26] is that being independent of the graph size and complexity. However, the dynamicity of ocean terrain and Kino-dynamic of the AUV remained unaddressed, which is a remarkable shortcoming prevent the platform from resembling real-world situations.

Generally speaking, an important issue with all studies outlined above is that they mostly focus on routing/tasking problem in which task assignment is the principal direction of these studies and quality of deployment (motion) and uncertainty of the environment have not been addressed. The vehicle's safe and confident deployment in the hazardous marine environment is a critical factor that should be taken into consideration at all stages of the mission.

A new synchronous hierarchal model was introduced by [27] for handling AUV's mission planning and task-time management problem. The approach provided vehicle's decision autonomy through the designing route planning system for ordering the execution of tasks and guiding the AUV toward the target location; coupled with a local path planner to ensure a safe and efficient deployment according to vehicles awareness of the situation. The authors in [27] mostly focused on increasing mission productivity and managing battery lifetime applying some population-based meta-heuristic approaches. Although the research emphasized a new point of view to autonomous mission planning, some details such as effect of water current force as an effective factor on AUV's deployment is not addressed.

Later on, MahmoudZadeh et al. [28] provided a mature version of the previous study [27] involved with developing a modular control architecture with two deliberative and reactive execution layers. The proposed interactive modular approach manages the concurrent execution of several tasks with different priorities through the deliberative layer, while the reactive layer controls vehicle's real-time responsiveness to critical events. The subject area is one that is of importance, as the authors point out, steps to reduce the reliance on expert operators that contribute to the scalability of AUV's application, reliability of operations and enhance AUVs self-management characteristics. However, this research is developed and evaluated through some simulations, which again may not be sufficient for real experiences.

Niaraki and Kim [29] modelled vehicle route planning with a multi-criteria decision-making mechanism and applied a generic ontology-based architecture taking advantage of a hierarchical analytic process to determine the choice of criteria for using an impedance function in the route finding algorithm. In this study, the domain-specific ontology provides a foundation for maintaining or extending the domain knowledge. The problem associated with knowledge-based (ontology-based) systems is that consistent information about the correct action and environmental situations is not always available.

Woodrow et al. [30] reported the development of concepts that are designed to improve the level of AUVs' autonomous planning and mission management. The study is focused on development of *i)* an intuitive user interface that moves away from specifying missions as a detailed scriptive approach toward a set of military goals, *ii)* an autonomous on-board mission (re)planning software that help AUVs to compensate for the changes in the mission goals, environment, or status that demand re-planning of the mission, and *iii)* transit planning software that makes the autonomous route planning possible with respect to environmental conditions, possible risks, and defined weights. The planning module of the package employs a hierarchical approach in which first a plan based on simple task models is generated, and the simple plan is refined in later levels to incorporate more detailed task plans. The re-planning level is split into two layers of mission re-planning and task planning layers. The developed transit planning software considers some environmental and risk-related factors for planning a transit path. These factors include *i)* Time-varying and non-uniform sub-surface currents, *ii)* Time-varying water levels, *iii)* Areas or times of high physical risk, *iv)* Exclusion zones, and *v)* Risk of detection.

Albiez et al. [31] proposed an adaptive AUV mission management model that utilizes a plan management mechanism applying both behaviour-based and predictive approaches to control the AUV and handle under-informed situations. The introduced hybrid reactive deliberative architecture uses behaviour-based methods to manage the tasks and uses an elaborate plan manager to control the deployment, activation, and deactivation of the behaviours in order to maintain the progress and fulfil the missions. The architecture includes blocks of sensor processing, vehicle management and safety, behaviour pool, and plan management. The feasi-bility of the architecture is assessed using AUV AVALON in a pipeline leak detection scenario.

Bian et al. [32] have been studied the implication of intelligent decision-making algorithm in AUVs tasking to perform in-depth oceanic survey in which the environmental constraints such as obstacles and forbidden zones have been considered. An intelligent decision algorithm constitutes a global path optimizer, and a speed optimization method along with a hierarchical on-board architecture is utilized for performing the mission. The mission management architecture utilized in the study is based on increasingly detailed Petri nets in which the separation of mission operations in different Petri nets increases the feasibility of adding new operations. In a similar scenario, Rajan et al. [33] dealt with abstracted diagnosis and failure recovery aiming to provide solutions for the robust, continues deployment of AUV's in severe underwater missions such as in-depth oceanic exploration. Investigating and analyzing current AUV operations model, the study identified some critical issues to be addressed in the design of AUV-based autonomous mission management system including: failure to turn up on time; uncertainty in onboard consumable; sensor failure conditions; impaired mobility; lack of built-in SA; limited adaptability to opportunistic events.

Pfuetzenreuter [34] tackled the mission re-planning issue in AUVs task management to survey in in-depth oceanic scenarios in which communication with human operators and requesting mission update is impossible. The study described the architecture in functional modules of mission control, mission plan handling, mission monitoring, mission re-planning, and chart server. Mission monitoring module is responsible for mission/plan initiation and observing mission execution while mission re-planning module is responsible for detecting the necessity of plan modification and generating mission updates that optimize the outcome concerning mission objectives. The chart server module checks the modified plan against digital charts to confirm its consistency with terrain constraints, and in case of detecting violation(s) the system considers alternative plans.

Brito et al. [35] presented the key obstacles for the integration of adaptive mission planning techniques for AUVs based on the exchanged opinions between experts in the field and the results gathered from some questionnaires. The study identified various reasons for the failure to adopt adaptive mission planning software including:

1) Technology is not understood (20.7%)
   - Technology is not well explained (55.4%)
   - Technology is too complex (44.6%)

2) Uncertainty with regards to vehicle responses (39.7%)
- Insufficient demonstration trials (65.8%)
- Lack of risk assessment (34.2%)

3) Technology is too expensive (21.5%)
- Development costs are not tangible (50.7%)
- No predefined development life-cycle (49.3%)

4) Benefits are not significant (15.0%)

5) Uncertainty with regards to legal limitations (3.1%)

Based on the gathered opinion from experts in the field, in the first level, uncertainty with regards to vehicle response is recognized as the most probable cause of failure to use adaptive mission planning software in AUV community. Within the second level, insufficient demonstration trials (26.1%) and the lack of risk assessment (13.5%) are considered as the probable cause of the failure to use these type of methodologies in AUV community.

## 2.3 Autonomous Mission Management Systems for Swarm Robotic Scenarios

Developing swarm based solutions for teams of robots that tackle problems through cooperation, coordination, and collaboration have several benefits over the use of singular robots with high intelligence and abilities but expensively complex. Using the swarm solution, however, requires a change in the way that a mission is planned and a proper task distribution mechanism to take the best use of the capacity of all contributed robots. There are several similarities between UAVs and AUVs in the context of mission planning problem and autonomous decision making, so we are also going to discuss the way that these problems are addressed over the multiple underwater and aerial vehicles collaborative operation.

Planning problems that are concerned with coordinating actions of multiple assets in which these assets are subject to failure are also referred to as "Multi-Agent Health Problem" [36]. Raghvendra et al. [37] counted faster completion time, and ability to cover larger areas as some of the advantages of using multiple UAVs in complex missions over the use of single UAV. They addressed this issue through the use of a temporary command control that allows any UAV in the group to take over the control of the team and ask other UAVs to readjust their position to allow sufficient maneuvering space whenever a re-planning is required. The approach keeps the human operators in the loop by letting them overrule any necessary action.

Sariel and Balch [38] introduced a generic framework named Distributed and Efficient Multi-Robot Cooperation Framework (DEMiR-CF) for distributed task assignment and to handle a complex mission for a group of robots requiring concurrent execution of some tasks. In the area of multi-agent transportation planning, Dominik et al. [39] implemented a new framework for route planning and decision making. In this study, the routing problem is considered with distinct time windows so that cluster-dependent tour starts and the agent chooses between traversing edges, distributing orders to customers, competing vendors, and so on. The outlined studies [38, 39] concentrated on task allocation among multiple robots, their move toward the destination, and managing the loose of any of agent in the group by reassignment of its tasks into the nearest agents. Nevertheless, these studies assumed the ideal environment, while in reality many uncertainties and considerations exist that should be taken into account. Simplistic assumptions or lack of existing crucial environmental factors can cause many problems in practical applications especially in the case of autonomous unmanned robots.

Combination of high-level structures and a bottom-up procedure is suggested by Zhu et al. [40] for navigating a swarm of robots. Using such a bottom-up layer provides some degree of autonomy in individual robots, while the higher level layer of intelligence assures the convergence in critical situations. This approach is advantaged to incorporate different algorithms designed for managing certain circumstances. In contrast to the outlined multi-layer platform of intelligence in [40], Pimenta et al. [41] considered a single software/algorithm approach to control the swarm that excludes individuality and independent decision making within the swarm. The principle idea in this research is to direct the swarm of robots towards the area of interest, which is beneficial since the swarm is no longer dependent on individual members. The Smoothed-Particle Hydrodynamics (SPH) simulation method is employed to produce an abstract representation of the swarm in the sense of incompressible fluid. Such representation provides a loose way for controlling the swarm.

A heterogeneous task selection approach in multi-robot coordination is investigated by De Lope et al. [42] with the idea of producing decentralized solutions by allowing robots individually and autonomously selecting tasks with the constraint of maintaining optimal task distribution. The constraint is facilitated using Learning Automata (LA) probabilistic algorithms and the Response Threshold Model (RTM). The potential of this approach is assessed through simulation study using random and maximum principle approaches for task selection. The results indicated that the maximum principle is a better fit for the task selection process. Furthermore, it is concluded that the RTM method is robust to noise and it is more time-intensive in comparison with the LA-based probabilistic approach. The results indicated the suitability of this method for addressing issues relating to task allocation within a swarm of robots in the absence of any central task scheduler unit.

An Ant Colony System (ACS) coupled with GA is applied by Geng et al. [43] for multiple UAVs mission planning and their collaboration in urban surveillance missions. In the study, GA is employed at the early stage for the mission planning and identifying the number of required agents to be involved in a mission to assure constant surveillance over the 3D environment. ACS is applied then to plan the shortest route and minimize the required changes in the driving altitude of each UAV in the team.

Jameson et al. [44] developed a general autonomous architecture with an onboard mission planning module for collaborative team operations considering aspects of awareness, intelligence, responsiveness, collaboration, and agility. At this framework, the highest level plans the missions for multiple teams of UAVs and the lower levels plans for each group of UAVs in the swarm and also individual UAV. The collaboration components consider information sharing, task coordination, task and responsibility allocation, dynamic team forming, and interacting with external assets. This architecture takes advantage of Lockheed Martin's MENSA technology for contingency management that incorporates routines for identifying plan dependencies and constraints.

Atyabi et al. [45, 46] investigated the problem of controlling a small swarm of simple robots in search and rescue scenarios with different degrees of complexity in terms of tasks' time dependency, dynamism, heterogeneous skill distribution between swarm members. The study investigated the potential of two modified PSO approaches of Area Extended PSO (AEPSO) and Cooperative AEPSO (CAEPSO), which shared some heuristics commonalities, but different ability in learning and handling a combinatorial type of noise. AEPSO performed a better local search in comparison with a variety of approaches including basic PSO, random search and linear search in both dynamic and time-dependent scenarios. The robustness of the AEPSO and CAEPSO to noise and task-time-dependency is considered to be the result of having better cooperation achieved by knowledge sharing and balancing the use of exploration and exploitation behaviors of the swarm.

Alighanbari and How [47] proposed a combinational approach of Robust Filter Embedded Task Assignment (RFETA) for multiple UAVs path planning/re-planning and dynamic SA-based task assignment in an uncertain variable environment. Tulum et al. [48] presented a situation aware agent-based approach for UAVs proactive route planning problem, in which the A* algorithm has been applied for finding the best order of waypoints and optimizing the distance and battery cost over the large-scale operation network. The study highlighted the relevance of route planning and SA; however, the proposed approach accommodates only a restricted tactical situations that is not sufficient to furnish all real-world possibilities. The technique needs a pre-computation to analyze the situations due to a high computational cost of A*, which is a substantial deficiency for a real-time platform.

Liu and Shell [49] also handled the challenges regarding the dynamic task assignment to multiple robots in which the change of location of tasks is managed applying a top-down partitioning strategy; however, solution quality is not perfectly satisfied in the simulation results. A neural network based self-organizing map has been integrated by Zhu et al., [50] with a velocity synthesis approach to managing swarm AUVs' dynamic task assignment and motion planning. Multiple target locations and time-varying water current in a three-dimensional operating field have been considered by this study, and the approach is capable of keeping the vehicle on the desired track during a mission [51]. However, the environment is assumed to be ideal without taking real-world uncertainties into account (e.g., any kind of static or buoyant obstacles, etc.).

Acknowledging the relevant studies in the scope of ground and aerial vehicles, the mission management becomes even further challenging for AUVs considering the severity and uncertainties of the underwater environment. The underwater environment is generally a vast 3D area, which is rarely populated with obstacles and usually unknown in advance. The challenges associated with such an uncertain environment becomes even more significant in long-range missions with enlargement of the operation area. Robustness of AUV to strong environmental variability such as severe turbulent or uncertain no flying zones, is a crucial concern for the mission performance and safe deployment. Restrictions of a priori knowledge about later conditions of the environment attenuate AUVs autonomy and robustness.

## 2.4 Summary of Chapter

This chapter provided a survey on some of the most recent developments in the field of UVs mission planning and mission management. Although several developing methodologies have been conducted to furnish the requirements of the mission planning and management problems, the impact of the human operators is usually ignored. This is the main criticism of the mission planner/manager for the possible failure of the operations while operators also have a significant impact on the whole progress of a mission. This issue becomes even more critical when multiple UVs are being controlled by very few operators especially in the environments with limited available information. In such cases, changes to the dynamism of the environment should be constantly applied to the pre-planned mission. Moreover, a higher level of SA for operators and the UVs should be provided along with advancing the level of workload on operators which had failure in making proper adjustments to the plan. Considering the given problems, it seems to be essential to develop an adaptive mission planner and management framework that not only automate the plan readjustments but also reduces the operators' workload (both physical and mental) while advances their SA.

# Chapter 3

# State-of-the-art in UVs' Autonomous Motion Planning

S. MahmoudZadeh[1], D.M.W. Powers, R. Bairam Zadeh

[1] Faculty of IT, Monash University, Clayton, VIC 3800, Australia
Email: Somaiyeh.mahmoudzadeh@monah.edu

**Abstract.** Autonomous mission management is closely related to the accuracy of the navigation system. Path planning is an essential component in the UV's development, which determines the vehicle's level of autonomy in dealing with environmental changes and it is considered as a premise of mission reliability and success [1]. One primary concern for autonomy is the advancement of the navigation system, including trajectory/path planning, to be robust to the extreme environmental variability.

From the robotics viewpoint, the path planning or in general the motion planning is considered as a multi-objective constraint optimization problem aiming to generate a feasible and efficient trajectory to autonomously guide the robot towards the target of interest in the corresponding operating area [2, 3]. As a matter of fact, optimization in any specific problem is a process defined over some essential criterion and constraints to produce a best-matched solution to the specified conditions.

Motion planning optimization problems usually concern traveling time/distance, safe deployments in terms of threat avoidance, and energy consumption as essential criteria for evaluation of the optimal path [4]. The identical aspect of motion planning systems is that they are obligated to perform prompt react to continuous changes of a variant environment. Therefore, realistic modelling of the environment, including the use of actual-practical assumptions, and efficient searching strategy are the primary steps in all motion planning approaches. This chapter provides a comprehensive review of the prior investigations in the area of individual and swarm UVs motion planning and its dependencies to autonomy.

## 3.1  Path Construction Techniques

Selecting a suitable and efficient method of path construction is a fundamental step toward achieving a satisfactory operation. The quality of the generated trajectory, such as continuity and smoothness, also needs to be taken into account [5]. Commonly, a potential path is formed by correlating a set of distributed control points, which are initialized in advance in the defined two or three-dimensional operation area. The most popular techniques for connecting these control points are as follows:

− **Straight-Line Method:** this method is used by [6] for the first time and is criticized due to insufficient flexibility in providing a smooth transition in the connections of piece lines.

− **Dubins-Based Methods:** this technique caught much interest in mobile robot path planning, which its mechanism is intercalating circles around the linkage so that the path gets smoother in connections of control points [7, 8]. However, these methods are criticized for their high computational cost, which defects the real-time performance of the system especially when path re-planning is required [9].

− **Spline and Piecewise Polynomial Strategy:** This is another popular category of path construction methods that uses just a few variables of control points' peculiarities to produce a comparatively flexible path curve. Obviously, dealing with small number of variables boosts the computation speed and eases the optimization process. Other improved versions of this method such as Cubic Hermite Spline [10] and B-splines [11] are developed, which tend to be more efficient in terms of computational time and smoothness [3].

Comparing the given path construction methods, the B-Spline is advantageous due to a mechanism of adding fixed points after start point and before ending point, which eases controlling the direction of the curve (vehicle) at these points. This functionality of B-Spline method makes it appropriate for application of vehicles with more physical constraints such as AUVs and UAVs.

## 3.2  Methodological and Technical Review on Autonomous Vehicles' Motion Planning

Motion planning in robotics framework has been investigated comprehensively over the last two decades, and different techniques have been suggested to handle this problem in different environments (e.g., aerial, ground, surface, and underwater). In general, the challenges associated with path planning can be studied in two main aspects: the methodologies being utilized, and the competency of the method in fast operation for real-time applications. Respectively, the performance of the path planning techniques comprises two perspectives: methods that mainly emphasize on resolution and accuracy of the solutions; and the group of techniques mainly concerned on minimizing the computational cost in producing feasible and probabilistically near-optimal solutions. Motion planners with the focus on solution resolution usually use graph search approaches, while the group with the time concern usually employs evolutionary algorithms aiming to satisfy the real-time constraints of autonomous operations. Each of these approaches is appropriate for a purpose respecting the problem demands of quality resolution or speed of the computation.

### 3.2.1 Graph Search Methods

Several graph search methods such as A*, Dijkstra, and D* algorithms have been employed in recent years to deal with the autonomous vehicles path planning problem that use a discrete optimal planning on a graph-like search space [12]. The identical characteristic of these algorithms is that the heuristic function estimates a solution with the minimum cost and then the algorithm keeps track of the potential solutions to determine the best possible cost compared to the estimated cost. To make these algorithms applicable on path planning problem, the search space should be transformed to a graph feature in order to define the forbidden sections and collision boundaries. Afterward, the path is constructed by patching the free segments upon the defined cost function. The nature of these algorithms is well suited on vehicle routing problems in which the default form of the search space is a weighted directional or bidirectional graph [13].

A modified Dijkstra Algorithm was employed by Eichhorn [14] for routing of a SLOCUM Glider (a kind of AUV) in a dynamic environment, in which the algorithm was facilitated with a time-variant cost function to simplify the search for a time-optimal route in the geometrical graph. The Dijkstra is a useful and popular algorithm that usually applied on routing problems and shortest path computation. However, this algorithm is computationally expensive due to calculating all possible paths in a graph until finding a best-fitted answer. This issue becomes even further challenging when the search space or the graph size increases and calculations need to be repeated for several times, which makes the Dijkstra computationally unsuitable for real-time applications [15].

A* is another graph search algorithm which is known as a best-first heuristic search method. This algorithm acts more efficiently comparing to Dijkstra due to its heuristic searching capability [16]. A* is applied by Carroll et al. [17] to obtain efficient paths with minimum cost in a quadtrees operation space. Perdomo et al. [18] used A* method on an offline dataset to determine a time efficient path with minimum risk for a gliders in a regional ocean model. The A* also suffers from expensive computational cost in larger search spaces. D* is a modified version of A*, which operates using a linear interpolation-based mechanism and has been vastly applied to the UVs' path planning problem. Using such a mechanism eases updating vehicle's heading directions; however, D* did not propose a considerable improvement to A* computational complexity in high-dimensional problems [19].

The Fast Marching (FM) algorithm is another derivative of the A* that applies a first-order numerical estimation of the nonlinear Eikonal equation. Petres et al. [20] employed FM algorithm to obtain a time efficient path for an AUV where the dynamic of the underwater environment such as water current field has been considered. Although FM is an accurate method but it is also computationally expensive than A* [21]. Afterward, a heuristically guided FM known as FM* is emphasized in [21] to deal with the same problem for UAV operation. FM* keeps the advantages of A* and FM in terms of efficiency and accuracy, but it is restricted to the use of linear anisotropic cost to improve the computational efficiency. Hence FM* has been further developed by Soulignac [22] applying a wave-front expansion method to determine the time optimum path from start point to the target.

Addressing the underwater environmental influence on AUV's deployment, Petres et al. [23] applied another derivative version of the FM called Anisotropic Fast Marching (AFM) method, which performed a better computational efficiency and guaranteed convergence in generating collision-free path. Although the effectiveness of this method has been mathematically proven in Mirebeau [24], it is still computationally expensive to be applied to online/real-time path planning systems. Moreover, this strategy may not be appropriate in dynamic environments that is continuously updated during the operation.

To deal with computational challenges, Sequeira and Ribeiro [25], confined the search space to a graph form with fewer connected nodes considered as obstacle regions, in which only static sea currents in a very simplistic fashion has been taken into account. The authors in [25] weighted the graph connections by energy costs and addressed the AUV's motion plan-ning using Dijkstra algorithm. Garau et al. [26] suggested the similar idea but restricted the motion to be in a two dimensional (2D) environment and applied A* algorithm to find the minimum cost path. In this approach, the operation area was split into a large and abrupt grid of connected nodes assigned with energy cost calculated upon journey time assuming a constant velocity. Although the simulation results performed a reasonable outcome, assuming a 2D search space is not sufficient to embody all the information of a 3D marine environment and vehicle's six degrees of freedom. Furthermore, the study has not consider the desirable currents for vehicles motion, which is important for saving mission time.

The grid-search-based methods are criticized due to their discrete state transitions, which restrict the vehicle's motion to a discrete set of directions. The graph-based search algorithms are popular in finding the shortest path in a network (graph); however, they are inefficient in large and complex workspaces, which is their main drawback. Their time complexity increases exponentially with enlargement of the search space and rendering the data becomes intractable. Reaching to a reasonable solution takes too long due to redundant iterative computations, while many iterations are required in these types of optimizers [27].

For a group of problems in which the deterministic techniques and heuristic-grid search approaches are not capable of satisfying real-time requirements, the artificial intelligence and evolution-based methods are good alternatives with faster computation, specifically in dealing with multi-objective optimization problems.

### 3.2.2 Artificial Potential Field Method

Artificial Potential Field (APF) method is another popular approach suggested to deal with the complexity of path planning problem that applies potential fields over the obstacles in a terrain to prevent the vehicle from crossing the collision boundaries. Sorbi et al. [28] applied APF on the multi-AUV motion-planning problem considering static obstacles in testing the collision avoidance capability of the method. However, assuming only the static obstacles is not sufficient to model the uncertain ocean. The APF with a combinatorial cost function has been employed by Witt [29], Warren [30] and Karuger et al. [31] for AUV path planning problem considering decision factors of time/distance, collision regions, and energy consumption. APF with a velocity synthesis method used by Cheng et al. [32] to deal with ocean current variations in AUV path planning; however, dynamic objects in the environment such as mobile/motile obstacles have not been considered in this research. Although APF is computationally fast and efficient, but it is susceptible to the local minima (the vehicle can be easily trapped in a U-shaped obstacle) [33].

To address the drawback of these approaches, Wu et al. [34] have suggested an improved APF method to enhance the performance of path planning. To address the APF local minima problem, the algorithm has been integrated with a wall-tracking technique, allowing the vehicle to get out of a trapped area by following the collision edge of any U-form obstacle. Moreover, the APF has been combined with Ant Colony Optimization (ACO), in which the ACO is applied for a global routing according to the generated path being used as the primary guidance route. Respectively the APF was in charge of producing the local path and avoiding collisions. Although the applied strategy gives a promising solution to handle the given challenges; however, integrating the APF with wall-tracking method increase the computational burden of the algorithm, which can cause delay in performing real-time actions.

It should be noted that the majority of the studies mentioned above concentrated on collision avoidance aspect of the path planning and ignored the environmental impacts on vehicles motion. Taking the influence of environmental dynamics into account becomes even more crucial in marine applications as the undersea environment is an uncertain, complex and unpredictable space while AUVs suffer from lack of communication and accessing localization facilities such as GPS. Such discrepancy can put vehicles' missions at risk, especially when the vehicle has restricted operating speed and relatively small dimension. This issue will be discussed in detail in Chapter 4.

### 3.2.3 Meta-Heuristics and Evolutionary Algorithms

Evolutionary algorithms are another approach employed successfully in dealing with NP-hard complexity of path planning problems. There is always time restrictions for fully autonomous vehicles on performing the real-time actions as all calculations are carried on by the onboard processor(s) according to changes appeared in the environment. In such systems, the path-planning problem in a large-scale operational field, satisfying time/energy, and collision constraints is more critical than providing an exact optimal solution; hence, having a quick acceptable path that satisfies all constraints is more desirable than taking a long computational time to find the best path. With respect to this priority, Meta-heuristic optimization algorithms are a desirable approach in such platforms with the ability to run on parallel machines with multiple processors and providing faster computations [35].

Alvarez et al. [4], applied GA to address AUV's path planning problem and highlighted energy efficiency as a factor of path optimality. This study considered the influence of strong time-varying ocean current on vehicles motion and applied a grid partitioning method on the search space to avoid convergence to local cost minima. Zhang [36] investigated a hierarchical GA-based path planning approach for AUVs, which relies on the decomposition of the operational space to deal with challenges of the large-scale operations and then searching for the best path at each level.

Rubio and Kragelund [37] also applied GA for UAVs' optimal path replanning according to the estimation of wind velocities. To reduce the energy consumption for the entire mission, the cost function is defined according to the vehicles fuel consumption characteristics along with desirable relative wind speed. Later on, an improved version of GA called Improved-Niche-GA (INGA) is utilized for carrying out the multiple UAVs real-time motion planning in a 3D environment [38]. The δ-field perturbation operator has been applied to enhance the local search capabilities of INGA and also an adaptive k-means method is used as a crowding mechanism to generate coverage paths in the area of interest. Ataei et al. [39] proposed an offline pre-generative three-dimensional path planner based on the Non-Dominated Sorting GA (NSGA-II) for a small size AUV, in which the study concerned on four criteria of path smoothness, length, the gradient of diving, and margin of safety. Such an offline planner gets into serious problems when the environmental situation is constantly updating and re-planning is required.

GAs uses genetic operators to pursue optimal results in a population of possible paths which are evolved iteratively. GA has shortcomings such as lack of consistency, which makes the vehicle's trajectories difficult to track, and lack of convergence, which means the generated path may be suboptimal. In contrast, graph search techniques such as Dijkstra's and A* methods perform better consistency and convergence due to using a discretised representation of the environment, known as a grid map. However, as a result of the non-holonomic constraint of some vehicles like AUVs, further path smoothing procedure is required [21]. Moreover, the computational time can be potentially high as it is proportional to the number of grids on a map, which is in turn dependent on the resolution of the graph.

Roberge et al. [35] developed an energy efficient path planner using meta-heuristic search nature of the PSO algorithm, where time-varying ocean current is encountered in the path planning process. GA and PSO share a similar population-based search mechanism in which a set of probabilistic and deterministic rules are applied to enhance the solutions' quality. PSO differs from GA in the sense of having a stochastic evolution mechanism that does not guarantee the best-fitted solution while GA incorporates this problem using mutation and crossover operators. PSO retains particles in each iteration and only adjust their characteristics applying update rules on their velocities and positions to best fit them to the cost function. Further comparison of PSO and GA can be found in [40].

A real-time Differential Evolution (DE) based motion planner is designed by MahmoudZadeh et al. [11] to provide a time/battery efficient operation for a single AUV in a dynamic ocean environment. Zamuda et al. [41] also applied a DE-based path planner for an underwater glider used in sampling of dynamic mesoscale ocean structures, where the dynamic ocean structure is also taken into consideration. Later on, another elaborated evolution-based online path planner/re-planner is proposed for AUV path planning and rendezvous problem in an uncertain underwater terrain to ensure AUV's safe deployment and secure docking [2]. Performance of four evolutionary algorithms of DE, PSO, Biogeography-Based Optimization (BBO) and Firefly Optimization Algorithm (FOA) is precisely investigated by this study to test and evaluate the robustness of the planner to current variability, and terrain uncertainties concerning the vehicular and the environmental constraints. The authors considered and compared vehicles ability in performing the SA-based autonomous reactions to different scenarios. Although suggested evolution based approaches perform promising performance in satisfying the path planning constraints for autonomous vehicles; however, evolutionary algorithms may quickly converge to a suboptimal solution. Thorough investigations on capability and efficiency of utilizing the Inverse Dynamics in the Virtual Domain (IDVD), a pseudo spectral method was proposed by Yazdani et al. [42, 43] to provide real-time updates of feasible trajectory for a single AUV.

It is conceived from the given literature in this chapter that UVs' autonomous navigation in unfamiliar and dynamic environments is a complicated process, especially when it is required to be fast in responding to environmental changes, where usually *a priori* information is not available. There are several publications in the area of UVs path planning; however, only a limited number of them addressed convincing field trail results of vehicles operation in an uncertain variant environment. Thus, judgment about efficiency and reliability of the developed technologies in dealing with dynamic environment is difficult and yet is dependent to complex considerations and evaluations.

## 3.3 Summary of Chapter

The subject of autonomy and motion planning have been comprehensively investigated on various frameworks and different environments over the past decades. The primary step towards increasing endurance and range of the vehicle's operation is improving the vehicle's ability to save energy. More advanced development to this end aims to increase the vehicle's autonomy in robust SA-based responsiveness for handling mission objectives that is directly influenced by navigation system performance. It has been discussed in this chapter that why having an efficient motion planning is principle requirement toward advance autonomy and facilitate the vehicle to handle long-range operations.

The challenges and difficulties associated with the motion planning strategies can be synthesized form two perspective: first, the mechanism of the algorithms being utilized; and second, competency of the techniques for real-time applications. The recent investigations over the UVs' path planning encountering variability of the environment has been discussed in three methodological categories of graph search approaches, artificial potential field method, and meta-heuristic evolutionary approaches.

The recent investigations on path planning encountering variability of the environment have assumed that planning is carried out with perfect knowledge of probable future changes of the environment, while in reality, environmental events are usually difficult to be predicted accurately. Although useful methods have been suggested for autonomous vehicles path planning, they still face several difficulties when operating across a large-scale geographical area. The computational complexity grows exponentially with enlargement of search space dimensions. A huge data load from whole environment should be analysed continuously every time that path re-planning is required, which is computationally inefficient. Additional to addressed problems in motion-planning realm, many technical challenges still remained unaddressed that can be the potential area of interest for further research.

# Chapter 4

# Advancing Autonomy by Developing a Mission Planning Architecture (Case Study: Autonomous Underwater Vehicle)

S. MahmoudZadeh[1], D.M.W. Powers, R. Bairam Zadeh
[1] Faculty of IT, Monash University, Clayton, VIC 3800, Australia
Email: Somaiyeh.mahmoudzadeh@monah.edu

**Abstract.** The subject of autonomy and mission planning have been comprehensively investigated on various frameworks over the past decades. It has been briefly discussed in previews chapters that among attempted scopes of autonomous operations, the underwater exploration remained still restricted to particular tasks with a low-level of autonomy. Underwater robotic platforms arouse more interest and broadly used in maritime approaches to achieve routine and permanent access to the undersea environment. Autonomous underwater operations including high and low-level mission-motion planning for an AUV is targeted as a case study in this book to precisely investigate challenges that a vehicle can face in long-range missions in harsh environment. These classes of autonomous vehicles are largely employed for numerous purposes such as scientific marine exploration, surveys, sampling and monitoring undersea biodiversity, offshore mapping, installations and mining, etc., [1, 2].

The reason for selecting this case study is remarkable restrictions of AUV's communication and maneuverability in a comparatively hard and uncertain environment. AUV's communication is restricted by high latency and excessively low bandwidth, which makes problem for the data exchange. Due to this reason, a sufficient level of autonomy should be achieved to make them trustable and accurate in carrying out the complicated mission [3]. This chapter will comprehensively discuss the difficulties of undersea missions, why this approach arose relatively more interest among other robotics application and will explain the scope and motivation of this study toward addressing a part of existing problems. An augmented reactive mission planning architecture is going to be introduced by the following chapters to accommodate an AUV's task-time management and mission planning in a top level, and to incorporate dynamics of underwater using a synchronic motion planner in a lower level.

## 4.1 AUV as a Case Study in This Research

Although AUVs are largely employed for various purposes, a majority of the existing AUVs make use of a pre-programmed mission scenario in which all parameters are specified beforehand, and human interaction is required at all stages of a mission. For any of mining, scientific, military, or surveillance application of AUV, a series predefined commands and tasks get fed to the vehicle through the mission scenario, which is a defect restricting its operation [4]. Accordingly, a superior degree of autonomy and an advanced decision maker is demanded to counterbalance the importance of the task and existing restrictions.

Another aspect of autonomy is to provide a safe and reliable deployment while handling uncertainties and changes of the surrounding environment that should be addressed. Abrupt changes of the terrain and unpredictable situations can enforce the mission to abort, and in worst case scenario might even cause the loss of the vehicle, as happened to Autosub2, which was lost under the Fimbul ice shell in Antarctic [5]. AUV's failure is not admissible due to expensive maintenance; hence a vehicle should be able to deal with unexpected events appropriately. The environment usually is unknown or partly known for an AUV, so having a precise definition of the present situation along with the accurate prediction of possibilities in one step forward can facilitate a safe operation and practical decision making. For this purpose, the next generation of AUVs should be facilitated with a high level of SA as a process of sensing, comprehending and dealing with extremely dynamic and intricate environments. Some of the important replacements toward advancing the vehicle's autonomy is given as follows:

− The vehicle should operate independent of human involvement.

− The vehicle should be able to detect and take instant measurements of oceanic properties and various situations.

− The vehicle should be to detect anomalies in the ocean. It should be noted that detecting these anomalies is a tough job that relies of SA capability of the vehicle.

− AUVs have limited battery life, so it should be able to manage a mission and deal with detected anomalies in a cost-effective way.

− The vehicle should be able to balance the relative costs between prioritizing the tasks in a mission and managing energy for traveling long distances.

Relying on a proper awareness of the situations, the AUV receives information from its surroundings, make decisions and takes actions according to the predefined knowledge and collected information. Ultimately, it considers the feedback from the environment in response to a particular action and updates its knowledge to produce better decisions in the next step. Hence, vehicle's awareness of environment must constantly be updated due to persistent change and transformation of the situations. This forces the human operator to adopt many cognitive strategies for maintaining SA in highly dynamic environments. For instance, consider a fish as a semi-intelligent agent. A fish moves to different positions and avoids colliding obstacles, while it is often unaware of how its body moves to avoid any collision, but in contrast, it can concentrate on where it wants to go. Automaticity is advantageous to SA because it requires very little conscious attention to process and extricates mental resources for other tasks.

### 4.1.1 The Exercised AUV's Specifications

For the purpose of this research, the REMUS-100 is selected, which is one of the most popular standard classes of AUVs with versatile suite of high-performance sensors and has a proven track record for consistent and reliable operations (depicted by Fig 4.1). This model is a lightweight, compact vehicle with commercial applications and has the following specifications (KONGSBERG website [6]):

− It is a modular AUV with dimensions of $0.2m$ in diameter, $1.6m$ in length, and less than $45kg$ in weight;

− It can operate with a maximum forward velocity of approximately 5 *knots* for maximum mission distances of roughly 55 *km* and can operate up to the depth of 100 *m*;

− It is equipped with a long-range Acoustic DOMPler Current Profiler (ADCP) operating in 75 *kHz* and is able to measure currents profiles up to 1*km* ahead of the vehicle;

− It is equipped with upward-looking downward looking, and sidescan sonar sensors operating in range of 75-540 *kHz* to percept and measure obstacles' features such as obstacles' coordinate and velocity;

− It is additionally equipped with underwater modem communication, GPS navigation, a Seabird CTD, a Wetlabs ECO sensor, and Wi-Fi.

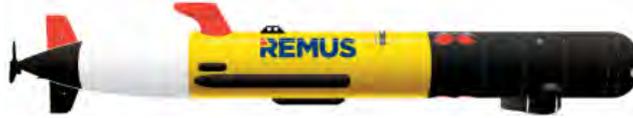

**Fig. 4.1.** REMUS-100: The Industry Standard Compact Man-Portable AUV [6].

## 4.2 Why Underwater Operations Are Challengious?

The underwater is generally a vast 3D volume that barely known in advance [7]. The undersea environment poses several challenges for AUV's deployment such as varying ocean currents, no-fly zones, and mobile/motile obstacles, which usually may not be fully known and characterized at the beginning of a mission. Obstacles may appear suddenly or change behaviour as the vehicle moves through the environment. An AUV operates in a turbulent undersea environment with complex spatiotemporal variability. Water current may have positive or disturbing effects on vehicle's deployment, where an undesirable current is a disturbance itself, and it also can push floating objects across the AUV's path. However, in the most of the previous approaches, the hydrodynamic effects of underwater environment and AUV's limited turning ability is ignored or not sufficiently considered [8]. Current flow sometimes can be very strong that cannot be controlled by the actuator of the AUV, so that avoidance of such zones would be necessary for safe deployment. Ignoring the water current has a detrimental effect on fuel consumption and even may cause serious risks for vehicle's operation. Robustness of AUV to this strong variability is a fundamental element of the motion planning and mission performance.

The vehicle uses acoustic information, imagery observations, and preliminary definitions to make decisions according to the raised situation and takes a matched reaction considering the different value of defined parameters. Existing a restricted information of the later environmental variations suppresses the AUV's robustness and autonomy. On the other hand, high latency and excessively low bandwidth restrict underwater communication. Thus, the vehicle can communicate to an operator and exchange data only before or after a mission. Therefore, it must be highly intelligent and trustable to carrying out the mission accurately [3]. The challenges associated with such an uncertain and unknown environment becomes even more significant in long-range missions (in which the operation area is considerably large).

Another concern about an AUV is its survivability by protecting itself and recovering from faults (if applicable), which helps to maintain its op-erations at near optimal performance. Most of current AUVs' applications are supervised from the support vessel that provides decisions in critical situations and generally takes enormous cost during a mission [9]. Respectively, increasing the vehicle's endurance and autonomy to handle longer missions without human supervision, and improving its hardware/software capabilities to reduce the operation costs, has become the area of interest for many researchers in the filed [10].

### 4.2.1 Challenges in the Scope of AUV's Mission Planning (Task Assignment and Routing)

Considerable attempts have been devoted in recent years to enhancing AUV's capability in robust mission planning in terms of routing, and efficient task/time management (has been discussed in Chapters 2). Although some improvements have been achieved in other autonomous frameworks, there is still a long way toward having a satisfactory level of intelligence and autonomy for underwater vehicles. AUVs ability of handling mission objectives is directly impacted by the performance of the routing and task assigning system. Effective routing and task prioritizing also improves mission timing and performance. As discussed earlier, AUV's task management and autonomous adaptation to a variant environment have not been fully satisfied yet, and it is still necessary for the operators to remain in the loop of considering and making decisions. Assuming that different tasks are distributed over a specific operating area and presented as connected waypoints in a graph-like terrain, some of the existing open problems in this scope are pointed as follows:

- An online mission planner must have a real-time performance to be able to satisfy time constraints of the time-critical missions and provide an online redirection for the AUV. The complexity of the graph topology or in general problem space results in comprehensive computational burden. Developing a fast and efficient mission planning approach for satisfying the real-time requirements of an autonomous operation is still an open area for research.

- A task sequence (in the form of connected waypoints) should be generated during the mission as the vehicle proceeds through the dynamic environment. Majority of previous attempts resulted in developing an offline mode of task-route planning solutions also known as pre-generative strategies. Many conventional AUV missions are limited to executing a list of pre-programmed instructions and completing a pre-defined sequence of tasks. Considering environmental changes and sit-uation updates, re-planning and rescheduling during the mission would be essential for autonomous operations. Such circumstances pose major difficulties on offline pre-generative planning approaches when the system is required to cope actively with the unexpected anomalies.

– The vehicle routing problem is usually studied as the shortest route problem in a network with a particular start and target point, which is known as a fundamental problem in graph theory and is a field of interest of many studies on transportation, communications, network routing, etc., [11, 12]. Complex missions such as logistic applications of an AUV need advance mission scheduling and task prioritizing strategy to maximize the mission productivity in a restricted period. A mission productivity can differ depending on the application, which usually relies on factors such as: decision-making autonomy when facing a new situation; prompt performance of the method to reschedule/re-plan the mission upon request; and efficiency of time and resource management. Therefore, having an intelligent mission planner for fully autonomous operations of AUVs is another area of interest for further research that is not sufficiently fulfilled yet.

– Majority of the existing studies specifically concentrated on task scheduling and target assignment problems while quality of the vehicle's deployment and requirements of a safe motion in presence of the environmental disturbances has not been fully considered. As mentioned earlier, underwater environment poses considerable uncertainties that can cause delay in proceeding the tasks and the overall mission timing. Hence, taking existing uncertainties into account and ensuring safe and efficient deployment is a crucial provision to accurate timing and assuring on-time mission completion.

## 4.2.2 Challenges in the Scope of the AUVs' Path/Motion Planning

The path planning techniques are specifically designed to deal with quality of vehicles' motion encountering environs properties and variations. As hinted earlier, the robustness of a motion planner to underwater disturbances and terrain uncertainties is crucial to AUV's reliable deployment and mission accomplishments. Presence of a priori knowledge about the variability of the current and the terrain facilitates the AUV to reduce the undesirable effects of the environment on its operation. Nevertheless, the existing technology can only estimate limited components of the ocean variability. Deficiency of the information around later conditions of the environment reduces AUVs autonomy in dealing with possible threats. A
remarkable attempt has been accomplished in recent years to enhance the
AUVs motion planning autonomy and extending their capacity for longer
operations. The current state of the art in the UV's guidance and the mo-
tion-planning problem is provided by Chapter 3. The major issues with ex-
isting AUV path planners and also critical hints for improving the level of
motion planning autonomy are clarified below:

- The existing planning studies in AUV platform are predominantly con-
  centrated on short-range local path planning in small-scale areas. Ex-
  tending the operation range and search space dimensions, however, is
  intertwined with new complexities, such as accumulation of data and
  propagation of uncertainty. Therefore, computational burden increases
  for longer missions due to the repetitious computation of an enormous
  data load of terrain updates. This huge data load should be analysed
  continuously every time that replanting is required, which is computa-
  tionally inefficient and unnecessary as awareness of environment in ve-
  hicle's vicinity such that it can react to the changes, is enough. Only a
  limited number of previous studies addressed the problems of AUV's
  long-range motion planning either regarding vehicle routing in a graph-
  like network or the path-trajectory planning. In order to reduce the
  complexity associated with the size of the terrain, some of the recent
  studies approach this problem by reducing the number of waypoints or
  2D implementation of the operation field. Shortcomings of these as-
  sumptions have been discussed earlier in Chapter 3.

- Usually information about obstacles' characteristics (position, velocity)
  is not certain and perfect. An obstacle's position or velocity may
  change over time or it may appear suddenly during vehicle's deploy-
  ment. Obstacles may drift by current force or may have a self-
  motivated speed and direction. The moving obstacles also can be cate-
  gorized as intelligent agents so that in such a case prediction of its be-
  haviour would be impractical. Hence, to carry out collision avoidance
  in such an uncertain environment the vehicle requires real-time capabil-
  ity of path computation. Moreover, current variation can change the be-
  haviour of moving obstacles over the time. Therefore, accurate estima-
  tion of the behaviour of a vast and dynamic terrain, far-off the sensor
  coverage is unreliable and impractical, so any pre-planned trajectory
  upon predicted maps may change to be invalid or inefficient.

- Another issue is that achieving the optimal solutions for NP-hard prob-
  lems is a computationally challenging and hard to solve in practice.
  Meta-heuristics such as evolutionary algorithms are diverse Bio-inspired methods and are known as a new revolution in solving complex problems due to their stochastic and/or deterministic mechanism. However, comparing the previously applied optimization algorithms for AUVs' motion planning is difficult as each of them addresses a specific aspect of this problem. Undoubtedly, there is always a significant requirement for more improvement in the application of these optimization techniques on AUVs' complex motion planning problem.

– There are several publications in the area of AUV path planning; however, only a limited number of them addressed convincing field trial results of AUVs operation in an uncertain spatiotemporal ocean environment. Thus, judgment about efficiency and reliability of the developed technologies in dealing with dynamic ocean environment is difficult and yet is dependent on complex considerations and evaluations. Moreover, a majority of stated motion planners did not accurately address the re-planning procedure as there is a significant difference between offline operation and online motion planning. Operating in a dynamic environment requires an online motion planner to perform an immediate reaction to raised changes. Despite some studies that addressed this issue, online path re-planning in longer-range operations is computationally expensive due to the repeated computation of a massive data load.

## 4.3 Advancing AUV's Autonomy in Mission Management and Robust Deployment

This book aims to advance an AUV's Autonomy by developing an Augmented Reactive Mission Planning Architecture (ARMPA), which provides the vehicle with comprehensive situational awareness for mission management. An AUV has specified battery lifetime and endurance that should be managed efficiently to boost the mission productivity and probability of success. Moreover, it is evident that an individual vehicle is not able to furnish all existent tasks in a single mission. An accurate mission planner and task management system should be able to supply higher-level decision-making for managing its resources, prioritizing the tasks, and performing a resilient operation without human interplay. This is subjected to having a reliable route- task planner, which can select an optimum order of tasks while guiding the vehicle to a defined destination.

Another complementary aspect of having a successful mission is to ensure the quality and security of AUVs deployment. Hence, integrating the system with a robust motion planner to accommodate efficient manoeuvres by adapting the sudden terrain changes and managing the travel time will satisfy the SA requirements of lower level (small-scale) operations. To this end, the ARMPA is designed to handle the given expectations in three main steps:

In the first step a time-efficient mission planning strategy capable of organizing tasks and planning global routes in an accurate manner will be developed in Chapter 5. The mission planner continuously monitors resources and reorder the tasks to maximize the mission performance. The system dynamically tracks the residual time, the unusual lost time, changes in the environmental situations; then adjusts its parameters according to the new situation by re-planning and reordering the tasks to be achievable in the remaining time. Further definition of the mission planning in this study is provided be subsection 4.3.1.

In the next step, which is provided by Chapter 6, a robust local motion planner will be developed to accommodate a reliable motion for the vehicle in dealing with an uncertain environment that sparsely populated with obstacles of static or moving type. In this framework, a battery-efficient path will be drawn along the split area bounded to distance between waypoints, where the persistent variation of the sub-area in the proximity of the vehicle is considered simultaneously. We already have discussed that the water current can perturb vehicle's motion and probably push it to an undesired direction [13]. The applied motion planner in this book comprehensively considers the static and time-varying current map, kinematic of the AUV, and uncertainties as key elements for realistic modelling of an underwater environment and providing safe and optimum operations.

In the last endeavour, this study will develop the augmented structure to encapsulate both implemented motion and mission planners and facilitate them with the capability of reactive re-planning and accurate synchronization. The system will be proposed in two different execution layers of deliberative and reactive (responsive), to satisfy the AUV's autonomy requirements in high and low-level approaches. In the given model, the mission planner operates at the top level as a deliberative layer, and the motion planner operates at reactive layer by concurrent back-feeding of the environmental condition to the system so that it makes decisions according to the raised situation. The operation of each layer may take longer than expectation due to unexpected circumstances. In order to reclaim the missed time, an efficacious synchronization scheme (named "Synchron") is added to the architecture to keep pace of modules in different layers.

The schematic design of the ARMPA process is depicted in the following
diagram by Fig 4.2.

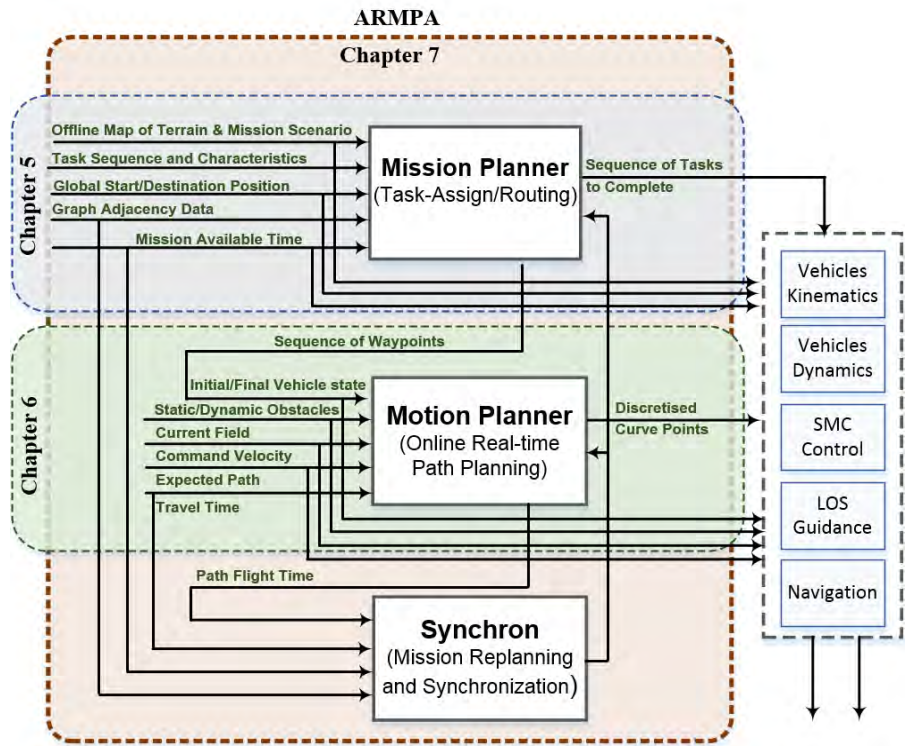

**Fig. 4.2.** The operation diagram of ARMPA and its completion process by each
chapter.

Some of the most famous evolutionary algorithms have been practiced
to investigate the performance of each layer, while minimizing the overall
computational time for the whole system is targeted to be the primary ob-
jective of the proposed approach. Numerical analysis of the simulation re-
sults has been carried out to evaluate the performance of the introduced
augmented architecture in different real underwater scenarios.

### 4.3.1 Mission Planning, Task Assigning and Routing in the Current Research

Generally, the task assignment can be a sub-process in routing or some-
times it can be considered as an individual problem. For example, in a
waypoint navigation problem, routing is the procedure of identifying way-
points to drive the vehicle from its current location to destination before mission times out. In contrast, the task assignment can be some sub-
routines that make sure point-to-point navigation is done properly and
tasks, which are mapped to different geographical areas, are completed ap-
propriately in a specified time. The mission planning in this book (depicted
in Fig 4.3) is performed as a joint problem of:

- Task assignment-allocation analogous to Dynamic Knapsack Problem
  (DKP). This problem has a discrete (synthetic) nature in which combi-
  nation of the solutions is essential.

- Vehicle Routing Problem (VRP) analogous to Traveller Salesman
  Problem (TSP). This problem has a contiguous nature in which order of
  solutions is essential (to guide a vehicle toward the destination).

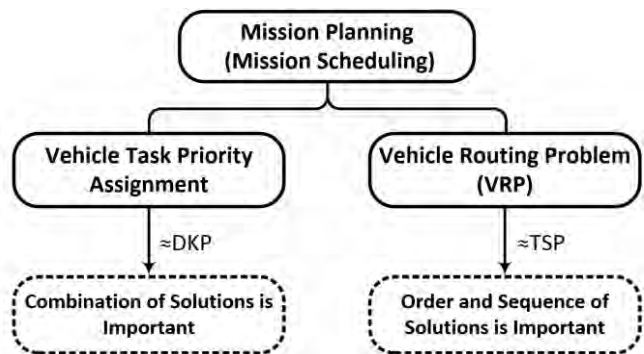

**Fig 4.3.** Mission planning components.

The tasks for any specific mission can have different characteristics
such as priority value, risk percentage, and absolute completion time,
which are known in advance by the AUV. Assuming that different tasks
are distributed over a specific operation area, they can be mapped by a
network (a sample map is illustrated in Fig 4.4). In this way, the complica-
tion of vehicle routing in a graph-like terrain can be addressed straightfor-
wardly. In the given sample in Fig4.4, the AUV is expected to reach on
time to the target waypoint after accomplishing the maximum possible
tasks of the highest priority, which means passing the adequate number of
edges in which any edge ($q_i$) is weighted with the characteristics of the cor-
responding task. The sequence of selected edges is an essential considera-
tion for saving the energy/time, vehicle's guidance, providing efficient
manoeuvres, and increasing mission productivity.

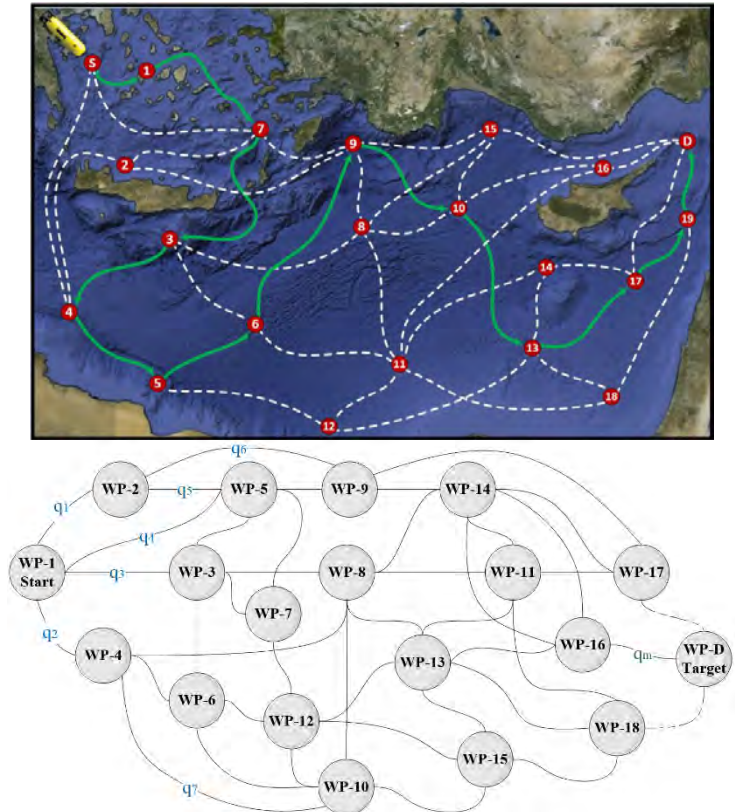

**Fig. 4.4.** Task distribution to an operation area. The tasks are mapped to the distance between waypoints.

Based on given explanations, mission productivity refers to appropriate selection and arrangement of the tasks sequence considering vehicle's availabilities, hardware/software capacities, and battery restrictions. This involves constraint decision-making procedures that usually reflected as a combinatorial optimization problem [14, 15].

## 4.4 Motivation and Contribution

The proposed idea is advantageous from several points of view. It links two diverse aspects of vehicle's autonomy in high-level task/mission management as a decision-making process and online motion planning through the self/environment awareness. The system has a consistent and cooperative mechanism where the deliberative layer at the higher level handles mission scenario and renders a general overview of the terrain by cutting off the operating area to smaller beneficial zones in the feature of a global route (ordered task sequence). Splitting the vast operation area into smaller sections addresses the weakness of motion planning methods associated with large-scale operations. This means search space reduction that leads a significant save to the computational cost; consequently, re-planning a new trajectory needs rendering less information.

Parallel execution of the system's deliberative and reactive layers accelerates the computation process, which is another advantage that reduces the operation time to the range of seconds. Moreover, the modular structure of the ARMPA advances the system to be able to conduct various algorithms with real-time performance.

The main reason behind the efficiency of the proposed approach is the fashion of mixing and matching two strategies from two different perspectives and composing an accurate synchronization between them. Such modular system is privileged to have the capacity of upgrading each module's functionality without any need to manipulating the whole system's structure. This advantage specifically increases the reusability and versatility of the control architecture and eases updating/upgrading AUV's functionalities to be compatible with other applications, and in particular implement it for other autonomous systems such as unmanned aerial, ground or surface vehicles.

## 4.5  Summary of Chapter

It is important for an autonomous vehicle to operate successfully in dealing with continuously changing situations. The main goal of AUV operation is to complete mission objectives while ensuring the vehicle's safety at all times. Although several investigations have been carried out in the scope UVs' motion planning, the existing approaches in the area of underwater operations mainly have covered only one aspect of either task assignment together with time management or path planning along with obstacle avoidance as a safety concern. To the best of authors' knowledge, there is currently no particular research in the scope of AUVs systematically emphasized both vehicle mission planning including task management problem (routing), and robust motion planning. To address the gaps associated with the given strategies, a general overview of the environment at the top level should be provided for the vehicle to make the vehicle more intelligent and robust in managing its availabilities and dynamic task/time management. Consequently, a local motion planner facilitates the system to operate successfully in dealing with uncertainties and environmental anomalies.

Respectively, this research introduces a modular control architecture named ARMPA incorporating two deliberative and reactive execution layers. The model is advanced with reactive re-planning capability in both of the execution layers. The construction of this study is defined in threefold. First attempt will be developing a mission planning (task-assigning and routing framework) to find a time efficient route in a graph-like terrain and appropriate arranging the tasks to ensure the AUV has a plenteous journey and efficient timing. In this framework, the operation network is generated in advance and different tasks are distributed to different parts of the generated map. Then an online local motion planner will be developed in Chapter 6 to ensure the vehicle has a time-efficient safe motion concerning all aspects of a real-world operation including the model of different static-mobile obstacles, time-varying ocean currents, vehicles Kino-dynamic, and geographical map information. Finally, mechanism and structure of this modular architecture will be explained in Chapter 7 and its performance in maximizing mission productivity, mission timing, real-time operation, and handling dynamicity of the terrain will be investigated.

# Chapter 5
# Mission Planning in Terms of Task-Time Management and Routing

S. MahmoudZadeh[1], D.M.W. Powers, R. Bairam Zadeh

[1] Faculty of IT, Monash University, Clayton, VIC 3800, Australia
Email: Somaiyeh.mahmoudzadeh@monah.edu

**Abstract.** Nowadays underwater missions and scenarios have been expanded and this will bold out the demands for having more powerful decision making of the AUV; thus, increasing mission productivity and reliability, concerning time and resource restrictions, are the target of interest in this chapter as the crucial factors of designing an efficient decision-making framework. The problem of accurate mission planning and vehicles ability of task prioritizing in the restricted time that battery capacity allows, and its guidance toward a specific destination is considered as a combination of the DKP and TSP (dynamic knapsack and traveling salesman) problems, will be addressed by this research. Considering the operating terrain as a waypoint cluttered volume in which the points' connections are assigned by specific tasks, the AUV should be able to fulfil a set of these tasks while taking graph routing restrictions into account. In this context, the mission planner in this chapter is designed to furnish the mentioned above expectations by satisfying the following objectives:

- Accommodating an effective route planning for AUV to take a maximum use of time that battery capacity allows;
- Accurate task ordering according to tasks' importance, riskiness, and their time span;
- Guaranteeing on-time completion of a mission while effectively managing the available time for a series of deployments in a long mission;

Attaining these objectives strongly depends on the optimality of the selected route between start and destination. Such a planner is proficient in apportioning the search space to beneficent sections for vehicle's deployment, which reduces the computational burden. Taking the advantages of meta-heuristic in addressing NP-hard graph problems, the mission planner, in this study, employs PSO, DE, and BBO to obtain a quasi-global route with the optimum arrangement of tasks for an underwater mission.

## 5.1 Formulation of the Task-Assign/Routing Problem

For any military, mining, or underwater scientific missions, a series of tasks such as water sampling, seabed habitat mapping, assembling or inspecting pipelines, mine exploration, seafloor mapping, etc., are defined beforehand and get fed to the vehicle in a set of command formats. This study specifies 30 various tasks presented by $\aleph$ in equation (5-1) that are characterized by a priority value (denoted by $\rho_i$), task's absolute risk percentage (denoted by $\xi_i$), and completion duration (denoted by $\delta_i$). The task's parameters are initialized once in advance according to the following uniform distributions.

$$\aleph = \{\aleph_1,...,\aleph_{30}\},$$

$$\forall \aleph_i, \quad \exists \rho_i, \xi_i, \delta_i \Rightarrow \begin{array}{l} \rho_i \sim \mathrm{U}(1,10); \\ \xi_i \sim \mathrm{U}(0,100); \\ \delta_i \sim \mathrm{U}(20,200); \end{array} \qquad (5\text{-}1)$$

To have a realistic model of the marine environment, the Whitsunday island's map in dimensions of **10×10** $km^2$ (*x-y*) and depth of **100** *m* (*z*) is considered to map out a three-dimensional volume (denoted by $\mathbf{\Gamma}_{3D}$). A k-means clustering method has been employed to split up the map into water zone (admissible for AUV's deployment), coastal and uncertain zones. The clustered map is converted to a 2D matrix filled by value of **1** for water-covered sections and a value in **[0, 0.03)** interval for coastal/ uncertain sections. The waypoints (nodes of the network) are distributed in water-covered sections, which are eligible for vehicle's operation. Tasks for a particular mission are placed on distance between waypoints; hence, some of the graph connections (edges) include a task with specific characteristics. Consequently, the operating field is indicated with an undirected connected graph denoted by **G=(P, E)**, where **P** denotes the nodes (waypoints), and **E** indicates the connections (edges) of the graph in which some of them comprise a specific task, formulated as follows:

$$G=(P,E) \Rightarrow \begin{array}{l} |P|=k \\ |E|=m \end{array} \Rightarrow \begin{array}{l} P(G): \{p^1,...,p^k\}; \\ E(G): \{e^1,...,e^m\}; \end{array} \Rightarrow e^{ij} = (p^i, p^j)$$

$$\qquad (5\text{-}2)$$

$$\Gamma_{3D}: 10000_x \times 10000_y \times 100_z$$

$$\forall p^i \in P \Rightarrow \begin{array}{l} p^i_{x,y} \sim \mathrm{U}(0,10000) \\ p^i_z \sim \mathrm{U}(0,100) \end{array} \Rightarrow p^i_{x,y,z} \in \{Map=1\}$$

here, *m* and *k* are the number of edges and nodes, respectively. The position of any arbitrary node in 3D volume of $\mathbf{\Gamma}_{3D}$ is indicated by $p^i_{x,y,z}$ and $e^{ij}$ is the connection between two waypoints of $p^i_{x,y,z}$ and $p^j_{x,y,z}$. The {Map=1} represents the water covered sections of the map, which is obtained from the applied k-means method. Figure 5.1 shows an example

of such a terrain:

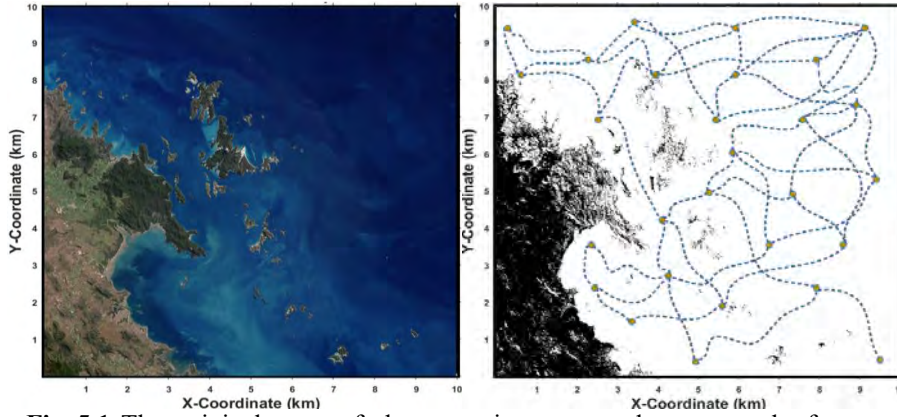

**Fig. 5.1.** The original map of the operating area and an example for graph
representation of the clustered map.

All nodes ($p_{x,y,z}$) are connected but only some of the connections ($e^{ij}$)
comprise a task like $\aleph_{ij}$, where these connections are assigned with a value
greater than one ($w_{ij}$ is weight of the connection $e^{ij}$) calculated by deviation
of the corresponding task's priority $\rho_{ij}$ to its risk percentage $\xi_{ij}$. The $w_{ij}$ is
calculated based on attributes of the corresponding task as follows:

$$\forall e^{ij} \quad \exists \quad w_{ij}, d_{ij}, t_{ij}; \quad w_{ij} = \begin{cases} \dfrac{\rho_{ij}}{\xi_{ij}} & if \ e^{ij} \wedge \aleph_l \\ 1 & otherwise \end{cases}$$

$$\forall e^{ij} \quad \Rightarrow \begin{cases} w_{ij} > 1 & if \quad \exists \quad \aleph_{ij} \\ w_{ij} = 1 & if \quad \exists ! \quad \aleph_{ij} \end{cases} \tag{5-3}$$

$$d_{ij} = \sqrt{\left(p_x^j - p_x^i\right)^2 + \left(p_y^j - p_y^i\right)^2 + \left(p_z^j - p_z^i\right)^2}$$

$$t_{ij} = d_{ij} \Big/ |\upsilon| + \delta_{ij}$$

the $d_{ij}$ and $t_{ij}$ are distance along the $e^{ij}$ and time for travelling the $e^{ij}$. The $|\upsilon|$
is the AUV's water referenced velocity (will be explained in detail in
Chapter 6). The autonomous vehicle is demanded to meet a certain number
of waypoints and to complete some of the existing time-dependent tasks.
For mission planning, the AUV needs to perceive information about
environmental factors, waypoints, terrain, vehicles restrictions, etc.,
comprehend the relation between the perceived information and meaning
of different raised situation in order to compute an optimum global route.
In facing any environmental changes, the mission planner dynamically re-
computes a new global route according to the new situations, and this
process will be repeated until the vehicle reaches the destination.

## 5.2  Shrinking the Search Space to Feasible Set of Tasks

The mechanism of finding an optimum order and number of tasks according to tasks' parameters and battery availability is an underlying problem to be addressed by AUV's mission planner. From SA point of view, some concepts should be regarded such as "what to search?" and "How to search?". Putting some boundaries based on problem's essential criteria would be a necessary step toward shrinking the initial search space to a more reliable and feasible state, and approaching the solutions to the desired outcome. As explained in the previous section, the operation area is mapped by a graph comprising several nodes and links in which some of the links corresponds to a specific task. Ordering and arranging of the tasks should follow a feasible pattern to be able to guide the vehicle toward the target of interest and prevent repeated traveling of a specific edge in the graph. Existing of prior information about terrain, tasks, and time availability will be helpful in producing such a valid pattern. Validity of the generated solutions will be assessed using the following criteria:

- A valid solution is an array of waypoints (a route) that initiated and ended with the index of a predefined start and target waypoints.

- The graph is not complete, so a valid solution does not comprise non-existent edges in the graph.

- A feasible solution does not involve a particular node for multiple times, as it implies wasting time on repeating a task.

- A valid solution does not overpass an edge more than once.

- A valid solution proposes a smaller completion time than the total residual time.

Taking the given criteria into account, MahmoudZadeh et al. [1] introduced a priority based strategy for finding valid routes that greatly improves performance of the optimization process. In this strategy, a randomly initialized vector is attributed to the nodes in the graph. This vector is filled with random positive or negative values in the specified range of [-100,100], where each element of the vector corresponds to existing nodes in the graph. Nodes get added to the array one by one according to their attributed value in the generated random vector and the graph adjacency information. The selected nodes get a large negative value to prevent repetition. The passed edges get eliminated from the graph adjacency matrix. After the process completed, the solutions get fed to the algorithm as the initial population. Figure 5.2 shows a hypothetical example of this process.

| | |
|---|---|
| ***Ad*** | Example of adjacency matrix for a graph with 18 nodes |
| ***n*** | Node index where *n=1* is the start and *n=18* is the destination point |
| $\Re^k$ | Partial route including *k* nodes. |
| $U_i$ | Random non-repeated vector (in range of [-200,100]) |

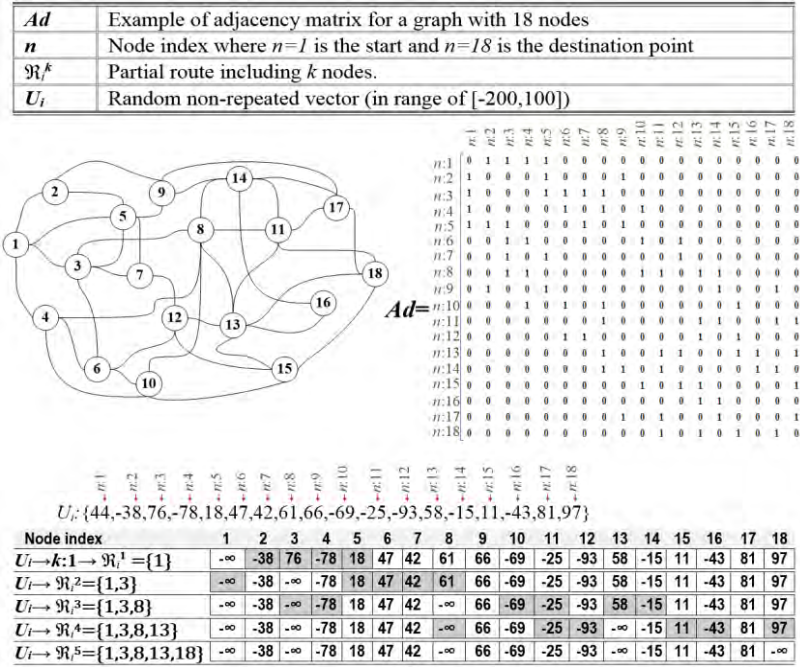

$U_i$:{44,-38,76,-78,18,47,42,61,66,-69,-25,-93,58,-15,11,-43,81,97}

| Node index | 1 | 2 | 3 | 4 | 5 | 6 | 7 | 8 | 9 | 10 | 11 | 12 | 13 | 14 | 15 | 16 | 17 | 18 |
|---|---|---|---|---|---|---|---|---|---|---|---|---|---|---|---|---|---|---|
| $U_i \mapsto k{:}1 \to \Re^1 = \{1\}$ | $-\infty$ | -38 | 76 | -78 | 18 | 47 | 42 | 61 | 66 | -69 | -25 | -93 | 58 | -15 | 11 | -43 | 81 | 97 |
| $U_i \mapsto \Re^2 = \{1,3\}$ | $-\infty$ | -38 | $-\infty$ | -78 | 18 | 47 | 42 | 61 | 66 | -69 | -25 | -93 | 58 | -15 | 11 | -43 | 81 | 97 |
| $U_i \mapsto \Re^3 = \{1,3,8\}$ | $-\infty$ | -38 | $-\infty$ | -78 | 18 | 47 | 42 | $-\infty$ | 66 | -69 | -25 | -93 | 58 | -15 | 11 | -43 | 81 | 97 |
| $U_i \mapsto \Re^4 = \{1,3,8,13\}$ | $-\infty$ | -38 | $-\infty$ | -78 | 18 | 47 | 42 | $-\infty$ | 66 | -69 | -25 | -93 | $-\infty$ | -15 | 11 | -43 | 81 | 97 |
| $U_i \mapsto \Re^5 = \{1,3,8,13,18\}$ | $-\infty$ | -38 | $-\infty$ | -78 | 18 | 47 | 42 | $-\infty$ | 66 | -69 | -25 | -93 | $-\infty$ | -15 | 11 | -43 | 81 | $-\infty$ |

**Fig. 5.2.** Sample of feasible route ($\Re_i$) generation based on topological information
(the value vector $U_i$ and Adjacency matrix ***Ad***).

In this picture, a sample graph with 18 nodes, its Adjacency matrix (*Ad*) a random vector ($U_i$) is demonstrated. To form a feasible sequence based on the given topological information, the first node is selected as the start position. The connected nodes to $n{:}1$ are selected from the Adjacency matrix ($n{:}2$, $n{:}3$, $n{:}4$, $n{:}5$). The node with the highest value (from $U_i$) in this sequence is added to the $\Re_i$ as the next node to be visited. This procedure continues until a legitimate route is built (destination is visited). This mechanism reduces the time and memory consumption, especially in the cases of complex and large graphs.

## 5.3 Mission Planner Optimization Criterion

In such a terrain that is covered by numerous waypoints/tasks, the vehicle should accomplish a maximum possible number of tasks with highest weight in the total available time denoted by $T_{\triangledown}$. To this end, the mission planner tends to find the best fitted $\Re_i$ to $T_{\triangledown}$ collecting a best order of tasks to guide the vehicle toward the destination, where the route total weight is maximized and on-time completion of the journey is assured. The problem incorporates multiple objectives that should be satisfied during the optimization process. One approach to address multi-objective problems is applying multi-objective optimization algorithms. Another alternative is to modify a multi-objective optimization problem into a constrained single-objective problem. Respectively, the objective function for mission planner is defined in the form of the hybrid cost function comprising weighted functions to be maximized or minimized:

$$\Re_k = \left( p_{x,y,z}^S, ..., p_{x,y,z}^i, ..., p_{x,y,z}^D \right)$$

$$\forall e^{ij} = \left( p_{x,y,z}^i, p_{x,y,z}^j \right) \ \exists \ w_{ij}, d_{ij}, t_{ij} \Rightarrow \begin{cases} w_{ij} > 1 & if \ \exists \ \aleph_{ij} \\ w_{ij} = 1 & f \ \exists! \ \aleph_{ij} \end{cases} \qquad (5\text{-}4)$$

$$T_\Re = \sum_{\substack{i=0 \\ j \neq i}}^{n} s_{eij} \times t_{ij} = \sum_{\substack{i=0 \\ j \neq i}}^{n} s_{eij} \times \left( \frac{d_{ij}}{|\upsilon|} + \delta_{ij} \right), \quad s_{eij} \in \{0,1\}$$

$$C_\Re \propto \left| T_\Re - T_\nabla \right| + \left( 1 / \sum_{\substack{i=0 \\ j \neq i}}^{n} s_{eij} \times w_{ij} \right)$$

$$s.t.$$

$$\forall \Re_i \Rightarrow \max(T_\Re) < T_\nabla$$

$$\Lambda_\Re = \max(1 - \frac{T_\nabla}{T_\Re}, 0) \qquad (5\text{-}5)$$

$$C_\Re = \left\{ \Phi_1 \times \left| \sum_{\substack{i=0 \\ j \neq i}}^{n} s_{eij} \times \left( \frac{d_{ij}}{|\upsilon|} + \delta_{ij} \right) - T_\nabla \right| + \Phi_2 \times \left( 1 / \sum_{\substack{i=0 \\ j \neq i}}^{n} s_{eij} \times \frac{\Phi_3 \times \rho_{ij}}{\Phi_4 \times \xi_{ij}} \right) \right\} \times$$

$$\times (1 + \Phi_5 \times \Lambda_\Re)$$

In (5-4), the $\Re$ is a potential route including a set of tasks generated by the mentioned strategy in Section 5.2. The $\Re$ is commenced with the position of the start node $p^s$ and ended at position of the destination node $p^D$. The $s_{eij}$ is a selection variable that equals to **1** for the selected and **0** for unselected edges. The $d_{ij}$ and $t_{ij}$ are distance along the $e^{ij}$ and time for travelling the $e^{ij}$. The generated rout time is denoted by $T_\Re$ that should approach the $T_\nabla$ but not overstep it. The $C_\Re$ is the mission planning cost function that is subjected to total available time of $T_\nabla$. The $\Phi_1$, $\Phi_2$, $\Phi_3$, $\Phi_4$ and $\Phi_5$ are positive numbers (defined by user) that determine amount of participation of each factor in the $C_\Re$ cost calculation. The $\Lambda_\Re$ is the penalty value assigned to the mission planner to prevent solutions from overstepping the time threshold $T_\nabla$. To address the objectives of the mission planner, this study employs three meta-heuristic algorithms of DE, PSO, and BBO, which have shown promising performance in solving NP-hard problems.

## 5.4 Overview of the Applied Meta-Heuristics

Meta-heuristics and evolutionary algorithms are cost-based optimization methods that operate using powerful design philosophy, often captured from nature, to deal with hard and complex problems. As discussed precisely in Chapters 2 and 3, the deterministic methods require considerable computational efforts to deal with vehicle routing or task assignment/planning problems that tend to fail when the problem size grows. To compensate this issue, a majority of the previous approaches emphasized on reducing the network complexity by minimizing the number of nodes or connections and restricting the engaged parameters. Regarding the combinatorial nature of mission planning, which is analogous to both DKP and TSP problems, this problem is characterized as a multi-objective NP-Hard problem often solved by optimization algorithms [1, 2]. Handling the NP-hard problems always associated with a considerable computational effort to achieve the optimal or semi-optimal solutions, which is difficult to solve in practice. Furthermore, producing the exact optimal solution is only possible for the certain situations where the context is fully known which is not the case in this study, as the spatiotemporal underwater environment is highly dynamic and uncertain. Meta-heuristics are appropriate approaches suggested for managing complexities mentioned above. Evolutionary methods have less sensitivity to graph size, so the search time grows linearly with the graph complexity. They are easy to implement and flexible to be implemented on parallel machines with multiple processors [2].

Usually, meta-heuristic algorithms demand accurate and careful modifications to accommodate requirements of a specific problem appropriately. Indeed, theoretical understanding of meta-heuristics is significantly distinct from their peculiarity to different applications in contrast. Therefore, careful adoption of the algorithm and accurate matching of its functionalities and characteristics according to the nature of a particular problem is an essential consideration, which usually is neglected in most of the robotics and engineering frameworks in both theory and practice. Undoubtedly, having a more efficient optimization technique is always a significant requirement for addressing autonomous vehicle's mission planning approach.

Several optimization metrics such as task priority, risk, order, and travel duration, time, safety, etc., are considered in mission planning to be minimized or maximized simultaneously to make the best use of the available time. A considerable amount of uncertainty affects vehicle's travel time; thus, accurate time management is essential to compensate the lost time and to ensure on-time mission completion. This research makes use of DE, PSO, and BBO heuristic search nature to address the mission planning objectives and the stated problems.

**_DE_**: The DE [3] is a stochastic search algorithm, which is known as an improved version of GA that operates based on biological evolution of selection, crossover and mutation. The DE has a discrete nature and has been extensively studied and applied on different realms of routing, task assignment, and the similar applications. Comparing to GA, the DE produces better solutions and faster process due to use of real coding of floating point numbers in presenting problem parameters. This algorithm applies differential mutation and non-uniform crossover operations.

**_PSO_**: The PSO is one of the fastest optimization methods for solving variety of the complex problems widely used in several studies in past decades. The particular issue with the PSO is that it is originally operating in a continuous space which is in contrast with discrete nature of the search space in task assigning; however, the argument for using PSO strong enough as it scales well with complexity of multi-objective problems. This concern has been addressed using the priority based route initialization approach explained earlier. This modification also accelerates the algorithm's performance and prevent stuck in a local optimum.

**_BBO_**: The BBO is another evolutionary method operates according to equilibrium theory of island biogeography concept [4]. The population of candidate solutions in BBO is defined by geographically isolated islands known as habitat. A distinctive feature of the BBO algorithm is that the initial population never get discarded but get modified by migration. This particular characteristic improves the exploitation ability of the algorithm.

### 5.4.1 DE on Mission Planning Approach

Suitable vector coding is the most critical step of DE process that directly impacts the algorithm's performance. This mission planner deals with finding the optimal task sequences through the operation network. Considering the network topology, a solution vector should not include extinct or repeated connections in the network. To keep the solutions feasible, a priority vector is applied to the initialization phase of the algorithm, in which the station sequence in each vector is selected according to their priority values and network's adjacency guiding information. The process is explained in the following steps:

- **_Initialization:_** The initial population of solution vectors $\chi_i$, in DE is initialized with feasible routes (task sequences); hence, the first and last element of the vectors always corresponds to index of the start and destination nodes. Solution space gets improved in each iteration $t$ applying evolution operators. Solution vector is described as $\chi_i$ ($i = 1 \dots i_{max}$), where $i_{max}$ is the number of individuals in DE population, $t_{max}$ is the maximum number of iterations.

- **_Mutation_**: This operator flips multiple elements of a vector to generate new offspring. The effectual modification of the mutation scheme is the main idea behind impressive performance of the DE algorithm, in which a weighted difference vector between two population members to a third one is added to mutation process that is called *donor*. Three different individuals of $\chi_{r1,t}$, $\chi_{r2,t}$, and $\chi_{r3,t}$ are selected randomly from the same iteration $t$, which one of this triplet is randomly selected as the *donor*. So, the mutant solution vector is produced by

$$\acute{\chi}_{i,t} = \chi_{r3,t} + F(\chi_{r1,t} - \chi_{r2,t})$$
$$r1, r2, r3 \in \{1, \dots, i_{max}\} \qquad\qquad (5\text{-}6)$$
$$r1 \neq r2 \neq r2 \neq i, \qquad F \in [0,1+]$$

where $F$ is a scaling factor that controls the amplification of the difference vector ($\chi_{r1,t} - \chi_{r2,t}$). Giving higher value to $F$ promotes the exploration capability of the algorithm. The proper donor accelerates convergence rate that in this approach is determined randomly with uniform distribution as follows

$$donor = \sum_{i=1}^{3}\left(\lambda_i \middle/ \sum_{j=1}^{3} \lambda_j\right)\chi_{ri,t}, \qquad\qquad (5\text{-}7)$$

where $\lambda_j \in [0,1]$ is a uniformly distributed value. This scheme provides a better distribution of the solution vectors. The mutant individual $\acute{\chi}_{i,t}$ and parent individual $\chi_{i,t}$ are then shifted to the crossover operation.

- **_Crossover_**: This operator shuffles sub parts of two parent vectors and generates mixed offspring. The parent vector to this operator is a mixture of individual $\chi_{i,t}$ from the initial population and the mutant individual $\acute{\chi}_{i,t}$. The produced offspring $\acute{\chi}_{i,t}$ from the crossover is described by

$$\begin{cases} \chi_{i,t} = (x_{1,i,t}, \dots, x_{n,i,t}) \\ \acute{\chi}_{i,t} = (\acute{x}_{1,i,t}, \dots, \acute{x}_{n,i,t}) \\ \ddot{\chi}_{i,t} = (\ddot{x}_{1,i,t}, \dots, \ddot{x}_{n,i,t}) \end{cases} \Longrightarrow \ddot{x}_{j,i,t} = \begin{cases} \acute{x}_{j,i,t} & rand_j \leq r_C \vee j = k \\ x_{j,i,t} & rand_j \leq r_C \wedge j \neq k \end{cases} \qquad (5\text{-}8)$$
$$j = 1, \dots, n; \quad n \in [1, i_{max}]$$

where $k \in \{1, \dots, i_{max}\}$ is a random index chosen once for all population $i_{max}$. The second DE control parameter is the rate of crossover $r_C \in [0,1]$ that is set by the user. After offspring are generated, the new generation should be validated.

- **Evaluation and Selection**: The offspring produced by the crossover and mutation operations is evaluated according to (5-9). The best-fitted solutions are selected and transferred to the next generation ($t+1$).

$$\dot{\chi}_{i,t+1} = \begin{cases} \dot{\chi}_{i,t} & C_{\Re}(\dot{\chi}_{i,t}) \leq C_{\Re}(\chi_{i,t}) \\ \chi_{i,t} & C_{\Re}(\dot{\chi}_{i,t}) > C_{\Re}(\chi_{i,t}) \end{cases}; \quad \ddot{\chi}_{i,t+1} = \begin{cases} \ddot{\chi}_{i,t} & C_{\Re}(\ddot{\chi}_{i,t}) \leq C_{\Re}(\chi_{i,t}) \\ \chi_{i,t} & C_{\Re}(\ddot{\chi}_{i,t}) > C_{\Re}(\chi_{i,t}) \end{cases} \quad (5\text{-}9)$$

The efficiency of the offspring and parents are compared for each operator by defined cost function $C_{\Re}$ and the worst individuals are eliminated from the population. Similar to the other EAs, the DE optimization process gets terminated if maximum number of iteration is completed, or if population fitness doesn't change after several iterations and approach to a stall generation.

## 5.4.2 PSO on Mission Planning Approach

The PSO start its search process using a population of $i_{max}$ particles initialized by feasible routes. Particle encoding is very important factor that affects effectiveness of the algorithm. During the search process, the population of routes is directed toward the regions of interest in the search space. Each particle involves a position and velocity that are initialized randomly in a specific range in the search space. The position of each particle (denoted by $\chi_{ij}$; $i \in [1, i_{max}]$; $j \in [1, l]$, where $l$ is the dimension of the search space) is readjusted during the search process through an updated velocity (denoted by $\upsilon_{i,j}$). The particles position and velocity are updated iteratively according to (5-10). The performance of particles is validated according to the defined cost functions. Each particle preserves its best position $\chi^{P\text{-}best}$ and swarm global best position $\chi^{G\text{-}best}$. The current state value of the particle is compared to $\chi^{P\text{-}best}$ and $\chi^{G\text{-}best}$ at each iteration.

$$\begin{aligned} \upsilon_{i,j}(t) &= \omega\upsilon_{i,j}(t-1) + c_1 r_{1,j}\left[\chi_{i,j}^{P\text{-}best}(t-1) - \chi_{i,j}(t-1)\right] + \\ &\quad + c_2 r_{2,j}\left[\chi_{i,j}^{G\text{-}best}(t-1) - \chi_{i,j}(t-1)\right] \\ \chi_{i,j}(t) &= \chi_{i,j}(t-1) + \upsilon_{i,j}(t) \end{aligned} \quad (5\text{-}10)$$

The $c_1$ and $c_2$ are acceleration coefficients, $\chi_{i,j}$ and $\upsilon_{i,j}$ are particle position and velocity at iteration $t$. $r_{1,j}$ and $r_{2,j}$ are two independent random numbers in $[0,1]$. The $\omega$ is an inertia weight and balances the local and global search. More detail about the algorithm can be found in [5]. Figure 5.3 summarizes the process of PSO.

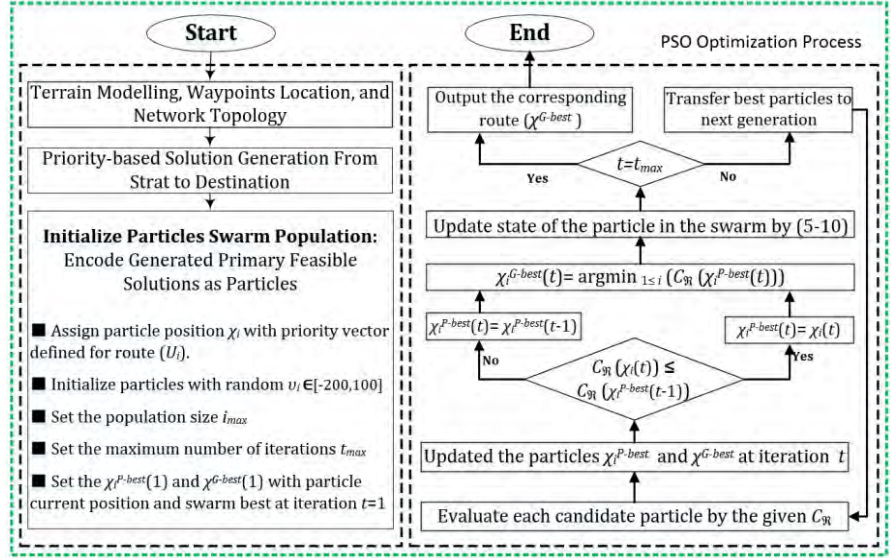

**Fig. 5.3.** Operation diagram of the PSO optimization process.

### 5.4.3 BBO on Mission Planning Approach

The habitat population in BBO is initialized with the generated feasible routes, and each candidate solution holds a quantitative performance index representing the fitness of the solution called Habitat Suitability Index (HSI). High HSI solutions tend to share their useful information with poor HSI solutions. Habitability relies on some qualitative metrics known as Suitability Index Variables (SIVs), which is a random vector of integers. Each solution includes design parameters of SIV, emigration rate ($\mu$), immigration rate ($\lambda$) and fitness value of HSI. Usually, the weak solution has higher immigration rate of $\lambda$ and lower emigration rate of $\mu$. Each given solution (denoted by $h_i$) gets adjusted according to probability of $P_S(t)$ that is the probability of existence of the $S$ species at time $t$ in habitat $h_i$ while $P_S(t+\Delta t)$ is the change in number of species after time $\Delta t$, calculated as follows:

$$P_S(t+\Delta t) = P_S(t)(1-\lambda_S\Delta t - \mu_S\Delta t) + P_{S-1}\lambda_{S-1}\Delta t + P_{S+1}\mu_{S+1}\Delta t \qquad (5\text{-}11)$$

$$\left.\begin{aligned} \lambda_S &= I_r \times \left(1 - \frac{S}{S_{max}}\right) \\ \mu_S &= E_r \times \left(\frac{S}{S_{max}}\right) \end{aligned}\right\} \xrightarrow{if\ E_r=I_r} \lambda_S + \mu_S = E_r \qquad (5\text{-}12)$$

where $I_r$ is the maximum immigration rate and $E_r$ is the maximum emigration rate. $S_{max}$ is the maximum number of species in a habitat. To have $S$ species at time $(t+\Delta t)$ in a specific habitat $h_i$, one of the following conditions must be hold:

$$\forall h_i(t) \quad \exists \lambda_S, \mu_S, P_S(t)$$
$$\begin{cases} \forall h_i(t) : \exists S \quad \Rightarrow \quad \forall h_i(t+\Delta t) : \exists S \\ \forall h_i(t) : \exists S \quad \Rightarrow \quad \forall h_i(t+\Delta t) : \exists S - 1 \\ \forall h_i(t) : \exists S \quad \Rightarrow \quad \forall h_i(t+\Delta t) : \exists S + 1 \end{cases} \quad (5\text{-}13)$$

The number of habitat's species increases by improvement of its suitability index and consequently the immigration rate decreases. In the next step, the mutation is applied, which tends to increase the diversity of the population to propel the individuals toward the global optima. Mutation is required for the solution with low $P_S$, while high-quality solutions are less likely to get mutated. Therefore, the mutation rate $m(S)$ is inversely proportional to probability of $P_S$.

$$m(S) = m_{\max}\left[\frac{1 - P_S}{P_{\max}}\right] \quad (5\text{-}14)$$

In (5-14), the $m_{max}$ is the maximum mutation rate assigned by operator, $P_{max}$ is the probability of habitat with maximum number of species $S_{max}$. The whole procedure is summarized in Fig 5.4.

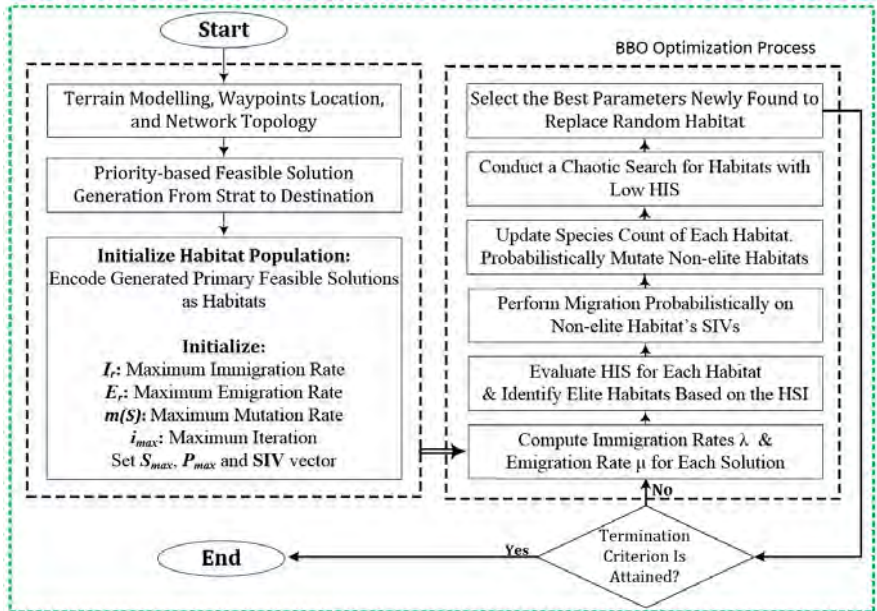

**Fig. 5.4.** Operation diagram of the BBO optimization process.

## 5.5 Mission Planning Simulation Results

Meta-heuristic methods usually get compared due to their similar optimization mechanism. Although they may show distinct performance on different problems, they are all capable of producing feasible solutions after a sufficient number of iterations have completed. An authentic benefit of these algorithms is their ability to handle non-linear cost functions. In real-time platforms, obtaining a prompt near optimal solution is more preferable than taking a long computation time to produce the best possible outcome. Therefore, the computational time is considered as one of the important performance indicators for evaluating the used methods in this chapter. The mission planner in this study uses the prepared information about waypoints location, terrain, tasks parameters, the network's topology, and vehicle's availabilities in terms of actuators and time/battery limitations to compute the most appropriate order of prioritized tasks and guide the AUV towards its destination. Hence, there should be a compromise among the mission available time, maximizing the number of highest priority tasks, and guaranteeing reach the specified target before the vehicle runs out of battery. All algorithms are configured with the population of 70 individuals and $t_{max}$=100 iterations. The BBO, is configured with the maximum mutation rate of $m(S)$=0.5, immigration rate $\lambda$=1-$\mu$, and emigration rate of $\mu$=0.2. The PSO is configured with acceleration coefficients of 1.5 and 2. The inertia weight is set to be decreased iteratively according to $\omega$=($t_{max}$ -$t$)/$t_{max}$. For DE, the lower and upper bounds of scaling factor were set on 0.2 and 0.8, and the crossover probability was fixed on 20% of the whole population for this experiment.

The problem is defined as a constraint routing in a graph-like terrain in which each connection represents a task. Consequently, raising the number of tasks increases the problem's complexity. The CPU time and cost variations of the applied methods are investigated on different network topologies, in which network complexity and size increases incrementally from 30 to 150 nodes, presented in Fig 5.5. Concerning the mission planning cost function (given by (5-5)), an optimum solution corresponds to a mission that takes maximum use of available time by maximizing the number of completed tasks and route weight, while respects the upper bound time threshold denoted by $T_{\overline{v}}$.

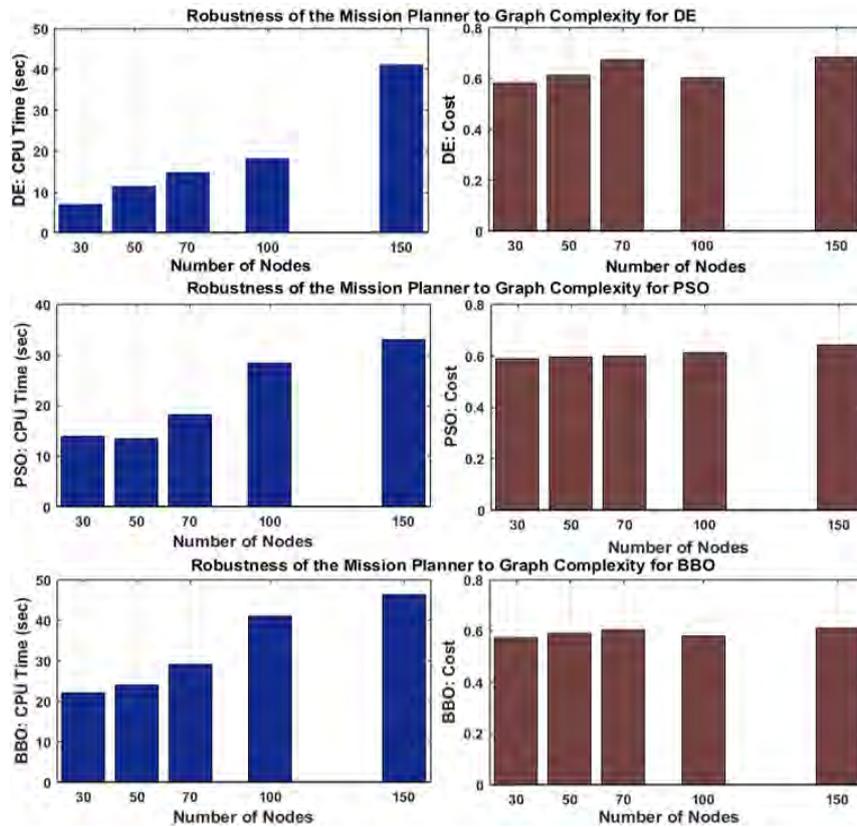

**Fig. 5.5.** The variations of mission cost and CPU time for different graph complexities.

As shown in Fig 5.5, the applied algorithms accurately manage the increasing complexity of the network as the cost value for all topologies lies in the similar range over the 100 iterations, while the computational time linearly increases with the growth of the graph nodes. The CPU time for all samples remained within suitable bounds for a real-time solution. The cost and CPU-Time variations indicate that the performance of the model is almost independent of both size and complexity of the graph, whereas this is recognized as a problematic issue in many of the outlined approaches.

Accordingly, the performance of the DE, BBO, and PSO algorithms is investigated quantitatively through the 150 Monte Carlo simulation runs (shown in Figs 5.6 to 5.8). Having quantitative Monte Carlo simulation, enhance the confidence on robustness and efficiency of the method in dealing with the random transformation of the graph size and topology. In the performed simulation, number of nodes is set to differ between 30 to 50 waypoints, connection between nodes and overall graph topology is randomly changed (using a Gaussian distribution) on the problem search space in each execution, while the time threshold is set on $3.42×10^4$ (*sec*).

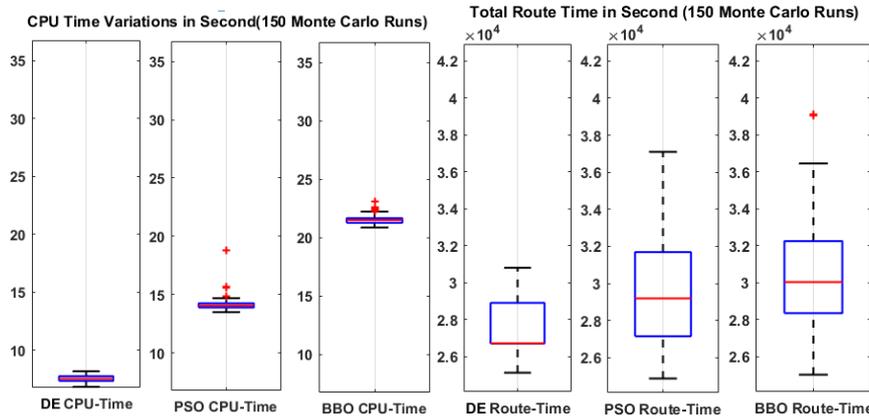

**Fig. 5.6.** Variations of CPU time and the route time ($T_\Re$) over the 150 Monte Carlo simulations, where the graph complexity of is changing with a Gaussian distribution in each execution.

Figures 5.6, 5.7 and 5.8 compare the functionality of the applied algorithms in dealing with problem's space deformation. It is evident from CPU-Time variations in Fig 5.6 that all algorithms are capable of satisfying real-time requirement of the proposed problem as the all variations are placed in a very narrow boundary in range of seconds; despite, the fastest operation belongs to DE and then PSO, BBO, in order.

The best possible performance for the mission planner is to produce routes with maximum travel time limited to given threshold. It is outstanding from route time variations in Fig 5.6, all algorithms accurately manage the route time to approach defined time threshold, but it seems that DE acts more cautious in constraining the route time as the variation range for DE is strictly kept far below the threshold. However, fluctuation of PSO and BBO almost placed in a similar range, while no outlier is seen in PSO route time variations, which is a good point for this algorithm.

Figure 5.7 indicates the performance of the proposed mission planner in managing the number of completed tasks and total obtained weight for each algorithm.

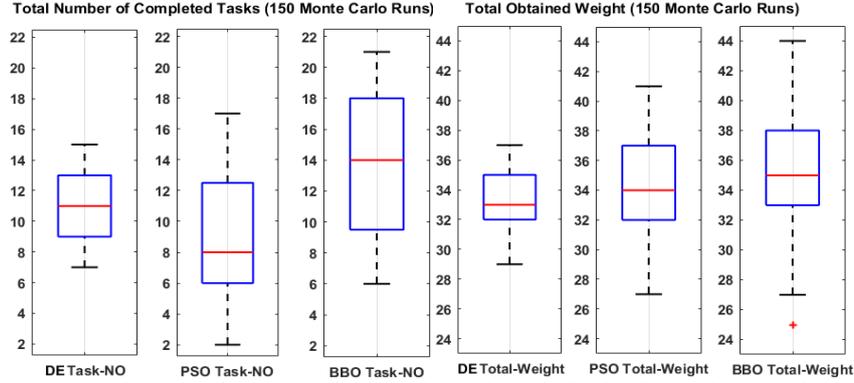

**Fig. 5.7.** Variations of the completed tasks and total obtained weight over the 150 Monte Carlo simulations.

It is shown in Fig 5.7 that the range of total obtained weight and completed tasks by DE appeared in a smaller interval comparing to the outcome of two other algorithms. It is also notable that PSO and BBO propose almost similar performance in the quantitative measurement of route obtained weight and completed tasks; however, being more specific, BBO acts more efficiently as the produced results by BBO dominates two others.

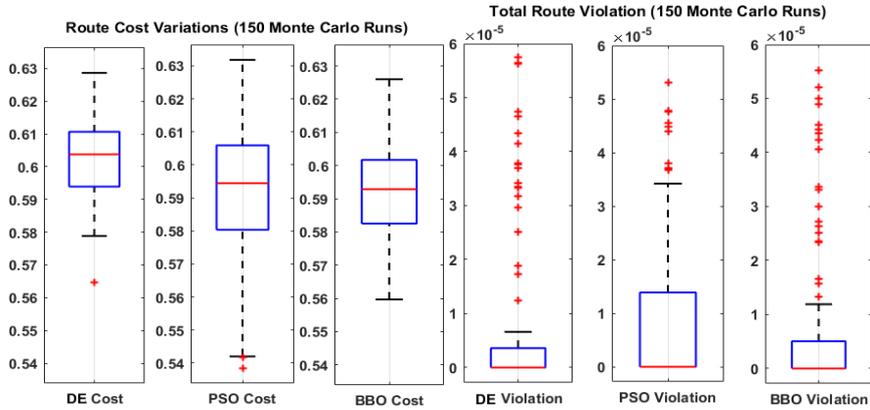

**Fig. 5.8.** Average variations of $C_\Re$ (route cost) and $\Lambda_\Re$ (violation) over the 150 Monte Carlo simulations.

To keep the solutions in a valid time range and assure on-time mission completion, a route (task sequence) gets a penalty value of $\Lambda_\Re$, when the $T_\Re$ oversteps the total mission available time $T_\nabla$. The quantitative variations of route cost of $C_\Re$ and violation of $\Lambda_\Re$ over the 150 Monte Carlo simulations are depicted in Fig 5.8. The $C_\Re$ is roughly varying in a similar range of [0.57 - 0.62] for all three algorithms, while the average variation of $\Lambda_\Re$ for Monte Carlo executions approaches zero. This analysis proves the efficiency of the applied algorithms in satisfying mission planning constraints and confirms their persistency against problem space deformation. As addressed earlier, the primary goal is to demonstrate the efficiency and real-time performance of meta-heuristics for this mission planner rather than criticizing or comparing them. However, among all three applied meta-heuristics, DE reveals better performance in fast computation and eliminating the route violation. A benchmark table is provided (Table 5.1) to summarize the performance of the DE, BBO, and PSO on satisfying the given performance metrics that quantitatively analyzed by Figs 5.6 to 5.8.

**Table 5.1.** Summarize of the DE, BBO, and PSO performance in satisfying the performance metrics.

|  | *Route Weight* | *#Tasks* | *CPU Time* | $T_\Re$ | $C_\Re$ | $\Lambda_\Re$ |
|---|---|---|---|---|---|---|
| ***Best to Worst*** | BBO | BBO | DE | PSO | BBO | DE |
|  | PSO | DE | PSO | BBO | PSO | BBO |
|  | DE | PSO | BBO | DE | DE | PSO |

Although PSO applies three principals of the transient state, local and global updates, it suffers the stagnation problem when the swarm traps in local optima. The particle swarm will eventually find an optimal or near optimal solution despite the fact that no estimate can be given on convergence time. The DE suffers the similar deficiency of the GA algorithm, where there is no guarantee in a random method for fast optimal solutions and when the complexity of graph increases the length of solution vectors increases accordingly so that mission-planning process becomes slower. However, taking the advantages of the applied priority based route generating method, all exercised algorithms operate in a competitive CPU time, unlike the conventional heuristic and deterministic approaches. Compared outcome of the applied algorithms in routing and mission planning process declares that they have very competitive performance with a very slight difference in the produced results.

## 5.6 Summary of Chapter

An AUV should be able to complete the best subset and ordering of the provided tasks while biasing to maximum priority in a restricted mission time. Realistically, it is impossible for a single vehicle to cover all tasks in a single mission in a large-scale operation area. Reaching to the destination is another critical factor for the mission planner that should be taken into consideration. All tasks are distributed in the operating field and mapped in the feature of a connected network, in which some edges of the network are assigned by tasks. The tasks should be prioritized in a way that selected edges (tasks) of the network can govern the AUV to the destination. Hence, the mission planner tends to find the best collection of tasks and order them in a way that AUV gets guided to the destination while it is respecting the upper time threshold (mission available time) at all times during the mission. Selecting the optimum route based on these criteria is a very challenging problem, especially when time restrictions forces overall operation performance. The problem involves multiple objectives that should be satisfied during the optimization process (given by (5-5)).

The planner tends to compromise among the mission objectives and constraints, while supporting vehicle's guidance toward the predefined destination, which is analogous to both TSP and DKP NP-hard problems. Relying on competitive performance of the meta-heuristics in handling computational complexity of NP-hard problems, the BBO, DE, and PSO methods are carefully chosen and configured according to expectation from the corresponding problem; and then used to validate the performance of the planner in finding correct and near-optimal solutions in competitive CPU time. Proper coding of the initial population is the most essential step in implementing all evolutionary algorithms. Thus, the individual population for all DE, PSO, and BBO are initialized with feasible and valid routes ($\Re$) provided by an accurate proven mechanism (discussed in Section 5.2). The efficiency and stability of the model in satisfying the given performance metrics is investigated in a quantitative manner through 150 Monte Carlo simulation runs with the initial condition analogous to real underwater mission scenarios. The study aims to prove the robustness of proposed model employing any algorithms with real-time performance, rather than criticizing or comparing the mechanism of the algorithms. The analysis of the results shows that all algorithms reveal very competitive computational performance.

This framework is a foundation for improving vehicle's autonomy in higher levels of decision making in prioritizing tasks, time management, and mission management-scheduling, which are core elements of autonomous underwater missions. However, the marine environment poses a considerable variability and uncertainty and the proposed mission planner is not capable of handling sudden environmental changes, but it gives a general overview of the area that an AUV should fly through (general route). Hence a regional online motion planner will be developed in Chapter 6 to accompany the mission planner in providing safe and effective deployments. The importance of addressing the environmental factors and their influence on vehicle's performance will be investigated through the next chapter.

# Chapter 6

# AUV Online Real-Time Motion Planning

S. MahmoudZadeh[1], D.M.W. Powers, R. Bairam Zadeh

[1] Faculty of IT, Monash University, Clayton, VIC 3800, Australia
Email: Somaiveh.mahmoudzadeh@monah.edu

**Abstract.** Robust motion planning is a complicated NP-hard problem that is considered as an essential characteristic of autonomy. This multi-objective problem considers the environmental disturbances and the possibilities for vehicles deployment during a mission. Although the recent advancements in embedded processors and sensor technology have opened new opportunities in underwater motion planning and facilitated AUVs to handle long-range operations, the inaccuracy of existing knowledge on uncertain spatiotemporal environment extends the complexity of motion planning problem. The class of underwater vehicles still have major challenges in dealing with uncertain ocean current variability that can strongly affect their motion, battery usage and mission duration. Current variations can also drift moving objects across the vehicle's trajectory; therefore, the planned path may turn to be invalid or inefficient. Another challenge is that having a precise estimation of the behaviour of such an uncertain/dynamic environment in long-range operations, outside the vehicle's sensor coverage, is usually unreliable and impractical. The robustness of a vehicle's path planning to this strong environment variability is a crucial consideration in vehicle's safety and mission performance.

No polynomial time algorithm exists to solve an NP-hard problem of even moderate size. On the other hand, obtaining an absolute optimum solution is only applicable in fully known and certain environments. The modelled underwater environment in this chapter corresponds to a highly dynamic uncertain environment. To address the challenges associated with the path planning across a dynamic large-scale geographical area, which has been discussed precisely in Chapters 3 and 4, this chapter aims to develop an online real-time motion planning strategy that enhances a vehicle's ability to cope with variations of surroundings and render a reliable trajectory for vehicle's transmission. The following objectives are introduced to furnish the mentioned above expectations:

- Avoid colliding static and uncertain mobile-motile objects;

- Avoid entering no-flying zones (e.g., coastal shallow areas and strong turbulence);
- Detecting anomalies and adapting/coping adverse current flow;
- Using accordant water current for saving energy;
- Prompt re-planning when an anomaly is detected;

To address these objectives, this study employs meta-heuristics of DE, PSO, and BBO in the core of the proposed local motion planner and investigates their performance of guiding the vehicle from an initial loitering point towards the destination through a comprehensive simulation study. To emulate a realistic ocean environment, the operating field in this study is modelled to be matched with real-world concerns and possibilities.

## 6.1 Modelling of the Operational Environment

Practical modeling of a real underwater environment needs strong mathematical modeling and assumptions that closely match with the real-world situations. This study encounters a three dimensional volume of $\mathbf{\Gamma}_{3D}$:{$3.5 \times 3.5$ $km^2$ ($x$-$y$), $100$ $m(z)$} presented by an offline geographical map, time-varying ocean current modeled with numerical estimation methods, and static/buoyant uncertain objects that provide an inclusive and extensive testbed for assessing the planner in various probable situations.

### 6.1.1 Offline Geographical Map

Having a prior knowledge about the environmental characteristics advances AUV's capability in robust motion planning. A sample map of the area of the Whitsunday Islands (as shown in Fig 6.1), has been conducted by this research to model a realistic marine environment. The K-means clustering method is employed as a popular method of data partitioning to cluster the coasts and authorized water zones into separate regions [1]. The method is initialized as $k$ cluster centers and the clusters are then iteratively refined. It converges when a saturation phase emerges or when there is no further chance for changes in assignment of the clusters. The method aims to minimize a squared error function in (6-1):

$$\arg\min_{C} \sum_{i=1}^{k} \sum_{\partial \in C_i} \left\| \partial - \mu_i \right\|$$  (6-1)

where $\partial$ belongs to a series of observations ($\partial_1$, $\partial_2$, …, $\partial_n$). Any $\partial_i$ corresponds to a d-dimensional real vector. The $\mu$ denotes the center of clusters $C=\{C_1,\dots, C_k\}$. The algorithm first takes the original map as an image of size 350-by-350 pixels in which each pixel corresponds to $10\times10\ m^2$. The algorithm recognizes the blue sections of the image as the water covered zones (eligible for AUV's deployment) and identifies the darker brown sections as the coastal or uncertain zones. Afterward, the geographical map will be divided into the black and white ranges corresponding to eligible and forbidden regions of operation. The next step is transforming the clustered map into a matrix format in which the matrix size is same to the image's pixel density. The matrix fills with the following digits:

– Value of (0,0.03] corresponding to uncertain shallow regions that is shown by grey color in Fig 6.1.

– Value of 0 corresponding to coastal regions that is shown by black color in Fig 6.1.

– Value of 1 corresponding to the water covered area that is shown by white color in Fig 6.1.

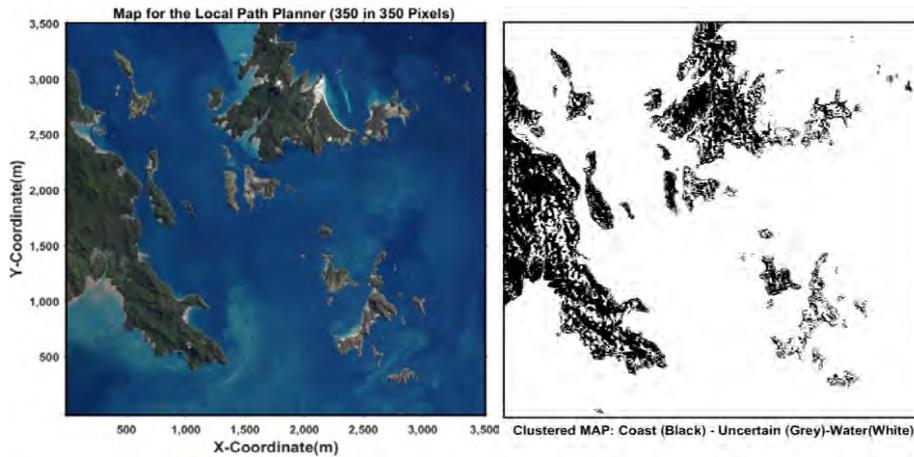

**Fig. 6.1.** The geographical map and clustered image used for simulating the operating field.

The utilized color-sensitive clustering method is able to cluster any alternative map efficiently depending on the color ranges of the image. The map analysis in this study is assumed in a 2D scale, in which depth of water is not considered in the map clustering process. However, the third dimension is taken into account in AUV's deployment, and the probable challenges in the 3D volume is simulated through the 3D modeling of the ocean current and obstacles.

## 6.1.2 Mathematical Model of Time-Varying Current Field

The water current in this study is modeled using a numerical estimator that uses the Navier-Stokes equations and multiple Lamb vortices [2] to mimic the three-dimensional behavior of the deep ocean current. The AUV's motion in the vertical plane is negligible due to its small scale comparing to vehicle's motion in the horizontal plane [3]. To implement a 3D dynamic current field, a multiple layer structure of 2D current map is employed, where the water circulations patterns gradually change in depth. Consequently, to estimate layers continuous circulation patterns, a recursive form of Gaussian noise is applied to the parameters of the 2D model. A probability density function of the multivariate normal distribution is utilized to compute the vertical profile $w_c$ of the 3D current $\boldsymbol{V}_C = (u_c, v_c, w_c)$, which is mathematically described as follows:

$$V_c : \begin{cases} u_c(\vec{S}) = \left( -\Im(y - y_0) + \Im e^{\frac{-(\vec{S} - \vec{S}^O)^2}{\ell^2}} \right) \Big/ 2\pi(\vec{S} - \vec{S}^O)^2 \\[2ex] v_c(\vec{S}) = \left( \Im(x - x_0) - \Im e^{\frac{-(\vec{S} - \vec{S}^O)^2}{\ell^2}} \right) \Big/ 2\pi(\vec{S} - \vec{S}^O)^2 \\[2ex] w_c(\vec{S}) = \gamma \Im e^{\frac{-(\vec{S} - \vec{S}^O)^T}{2\Delta_w(\vec{S} - \vec{S}^O)}} \Big/ \sqrt{\det(2\pi\Delta_w)} \end{cases}$$

$$\Delta_w = \begin{bmatrix} \ell & 0 \\ 0 & \ell \end{bmatrix}; \quad X^{wc} \sim N(S^o, \Delta_w) \tag{6-2}$$

$$V_c : (u_c, v_c, w_c) = f\left(\vec{S}^O, \Im, \ell\right)$$

The $S$ in (6-2) denotes a 2D span in *x-y* coordinates, the center, strength, and radius of a turbulent are respectively shown by $S^O$, $\Im$, and $\ell$. The $\Delta_w$ is a covariance matrix based on radius of the turbulent. The $\gamma$ is a parameter to scale the vertical profile of $w_c$ from the horizontal components. For modeling the dynamicity of the current, Gaussian noise is recursively applied to $\ell$, $\Im$, and $S^O$ parameters as follows:

$$S_t^o = A_1 S_{t-1}^o + A_2 X_{(t-1)}^{S_x} + A_3 X_{(t-1)}^{S_y}$$
$$\ell(t) = A_1 \ell(t-1) + A_2 X_{(t-1)}^{\ell}$$
$$\Im(t) = A_1 \Im(t-1) + A_2 X_{(t-1)}^{\Im} \tag{6-3}$$
$$A_1 = \begin{bmatrix} 1 & 0 \\ 0 & 1 \end{bmatrix}, A_2 = \begin{bmatrix} U_R^C(t) \\ 0 \end{bmatrix}, A_3 = \begin{bmatrix} 0 \\ U_R^C(t) \end{bmatrix}$$

The current field update rate is denoted by $U_R{}^C(t)$ and $X^{\delta x}{}_{(t-1)} \sim \mathbf{N}(0, \sigma_{Sx})$, $X^{\delta y}{}_{(t-1)} \sim \mathbf{N}(0, \sigma_{Sy})$, $X^\ell{}_{(t-1)} \sim \mathbf{N}(0, \sigma_\ell)$, $X^{\Im}{}_{(t-1)} \sim \mathbf{N}(0, \sigma_{\Im})$ are Gaussian normal distributions.

## 6.1.3 Mathematical Model of Uncertain Buoyant and Static Obstacles

Along with collision boundaries of an offline geographical map, different types of uncertain static and buoyant objects exist in the operational field which can have unpredictable behavior and should be taken into account of motion planning. An AUV is equipped with sonar sensors to measure the velocity, coordinates, and diameters of the objects with a level of uncertainty. To simulate different possibilities of the real world situations, the obstacles in this study are modeled in the following two categories.

***Uncertain Static Objects:*** These objects are introduced with a fixed-center of $\Theta_p$ and located between AUV's starting and destination point in the given map. They are considered with an uncertain radius of $\Theta_r$ that varying over the time according to (6-4).

$$\forall \Theta_{x,y,z}, \quad \exists \left( \Theta_p, \Theta_r, \Theta_{Ur} \right)$$
$$\Theta_p^i \in \left[ p_{x,y,z}^S, p_{x,y,z}^D \right] - \Theta_{r_{x,y,z}}^i$$
$$\Theta_p \sim \mathrm{N}(0, \sigma_1) \tag{6-4}$$
$$\Theta_{Ur} \sim \mathrm{U}(0, \sigma_2)$$
$$\Theta_r \sim \mathrm{N}(\Theta_p, \Theta_{Ur})$$

The $\Theta_{Ur}$ denotes uncertainty over the radius of $\Theta_r$. The $(p^S_{x,y,z})$ and $(p^D_{x,y,z})$ are the location of start and destination, respectively.

***Uncertain Mobile and Buoyant Objects***: These objects are considered to be buoyant that are affected with the water current flow or modelled to have a self-motivated velocity to a random direction (to emulate unpredictable behavior of some unknown objects):

$$U_R^C = |V_C| \sim \mathrm{N}(0, 0.3)$$
$$X_{(t-1)} \sim \mathrm{N}(0, \sigma_3)$$
$$\Theta_p(t) = \Theta_p(t-1) \pm \mathrm{N}(\Theta_{p_0}, \Theta_{Ur}) \tag{6-5}$$
$$\Theta_r(t) = B_1 \Theta_r(t+1) + B_2 X_{(t-1)} + B_3 \Theta_{Ur}$$
$$B_1 = \begin{bmatrix} 1 & U_R^C(t) & 0 \\ 0 & 1 & 0 \\ 0 & 0 & 1 \end{bmatrix}, B_2 = \begin{bmatrix} 0 \\ 1 \\ 1 \end{bmatrix}, B_3 = \begin{bmatrix} 0 \\ 0 \\ U_R^C(t) \end{bmatrix}$$

The $U_R{}^C(t)$ denotes the impact of current velocity ($V_C$) on objects motion. $\Theta_{Ur} \sim \sigma$ is the rate of change in objects position, and $X_{(t-1)} \sim N(\Theta_p, \sigma_0)$ denotes the Gaussian normal distribution that assigned to each obstacle and gets updated over the time $t$.

## 6.2 Formulation of the Regional Motion-Planning Problem

In order to compute a safe and time-optimum path, AUV's current state information and the environmental factors affecting its motion are taken into account while specific objectives are satisfied respecting vehicle's properties and constraints. An optimum path takes the minimum travel time/distance and is safe enough to avoid collision boundaries. The motion planner should be capable of extracting valid areas of a geographical map to recognize the conceivable space of deployment. Another concern is the water current. A severe current can act as a disturbance by pushing the vehicle to an unwanted direction, or it can drive the buoyant objects across the AUV's trajectory. In contrast, accordant water flow can motivate vehicle's motion, which results in saving energy/time and considerably diminish the total operating costs. Therefore, accurate adaption to the current variations and assessing its influence on AUV's Kino-dynamic is an important consideration for planning an optimal trajectory. The AUV has a freely deploying rigid body with six degrees of freedom providing a 3D motion capability, in which its state variables of body frame $\{b\}$ and **NED** (North-East-Depth) frame $\{n\}$ are defined based on a set of ordinary differential equations given by (6-6) and (6-7):

$$\begin{cases} \{n\} \rightarrow \eta : (X, Y, Z, \varphi, \theta, \psi) \\ \{b\} \rightarrow \upsilon : (u, v, w, p, q, r) \end{cases} \tag{6-6}$$

$$\begin{bmatrix} \dot{X} \\ \dot{Y} \\ \dot{Z} \end{bmatrix} = \begin{bmatrix} {}_b^n R \end{bmatrix} \begin{bmatrix} u \\ v \\ w \end{bmatrix}; \quad \begin{bmatrix} {}_b^n R \end{bmatrix} = \begin{bmatrix} \cos\psi\cos\theta & -\sin\psi & \cos\psi\sin\theta \\ \sin\psi\cos\theta & \cos\psi & \sin\psi\sin\theta \\ -\sin\theta & 0 & \cos\theta \end{bmatrix} \tag{6-7}$$

The $\eta$ and $\upsilon$ representing AUV's state in the $\{n\}$ frame and velocity in the $\{b\}$ frame, which affirm vehicle's dynamics and kinematics over the time [4]. $X,Y,Z$ denotes AUV's position along the generated path, and $\varphi, \theta, \psi$ are the Euler angles of roll, pitch, and yaw, respectively. AUV's directional velocities of surge ($u$), sway ($v$), heave ($w$) and rotational velocities of $p$, $q$, $r$ describe vehicle deployment in x-y-z axis. Ultimately, rotation matrix of [$_b{}^n R$] is defined in (6-7) to transforms the $\{b\}$ frame into the $\{n\}$ frame. AUV's and water current coordinates are depicted in Fig 6.2.

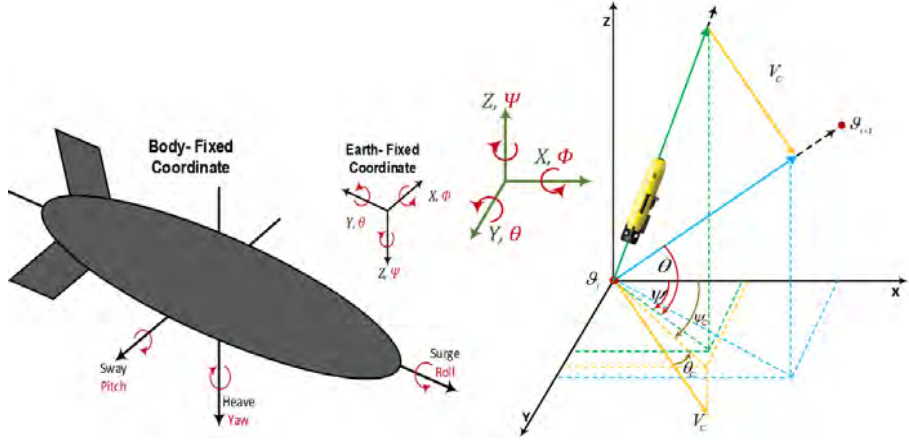

**Fig. 6.2.** AUV's and water current coordinates in the $\{b\}$ and $\{n\}$ frame.

The initial potential path of $\wp_i$ in this study is designed using the basis spline method in which the path curves is captured by locating a set of control points $\vartheta: \{\vartheta^1_{x,y,z}, ..., \vartheta^i_{x,y,z}, ..., \vartheta^n_{x,y,z}\}$ as described in (6-8) and (6-9). The motion planning is a constraint optimization problem that aims to render the safest and quickest trajectory between two points. The placement of these control points plays a substantial role in optimality of the solutions. All $\vartheta$ are defined to be located in a valid space limited to predefined Upper-Lower bounds of $\vartheta \in [U_\vartheta, L_\vartheta]$ in Cartesian coordinates. The process of path generation using the basis spline method is as follows:

$$L_\vartheta \equiv p^a_{x,y,z} \quad \& \quad U_\vartheta \equiv p^b_{x,y,z}$$

$$\begin{cases} \vartheta^i_x = L^i_{\vartheta(x)} + Rand^x_i [U^i_{\vartheta(x)} - L^i_{\vartheta(x)}] \\ \vartheta^i_y = L^i_{\vartheta(y)} + Rand^y_i [U^i_{\vartheta(x)} - L^i_{\vartheta(x)}] \\ \vartheta^i_z = L^i_{\vartheta(z)} + Rand^z_i [U^i_{\vartheta(x)} - L^i_{\vartheta(x)}] \end{cases} \tag{6-8}$$

$$\left.\begin{array}{l} X = \sum_{i=1}^n \vartheta^i_x \times B_{i,K} \\ Y = \sum_{i=1}^n \vartheta^i_y \times B_{i,K} \\ Z = \sum_{i=1}^n \vartheta^i_z \times B_{i,K} \end{array}\right\} \mapsto \wp_{x,y,z} = \sum_{x_s,y_s,z_s}^{|\wp|} \sqrt{\Delta X^2 + \Delta Y^2 + \Delta Z^2} \tag{6-9}$$

The lower and upper bounds of $L_\vartheta$ and $U_\vartheta$ in (6-8) are respectively set with the location of the start and destination points ($L_\vartheta \equiv p^a_{x,y,z}$; $U_\vartheta \equiv p^b_{x,y,z}$). This will keep the path in the valid search space. The $B_{i,K}$ is a blending function used to slice the curve, and $K$ is the order of the curve's smoothness (further information about basis spline method can be found in [5]). The AUV should be oriented along the path segments generated in (6-10) to (6-12). Water currents constantly affect the vehicle's motion; so the

AUV's angular velocity components along the path curve $\wp$ is calculated by:

$$\psi = \tan^{-1}\left(\frac{\left|\vartheta_y^{i+1} - \vartheta_y^i\right|}{\left|\vartheta_x^{i+1} - \vartheta_x^i\right|}\right) = \tan^{-1}\left(\frac{\Delta Y}{\Delta X}\right)$$

$$\theta = \tan^{-1}\left(\frac{-\left|\vartheta_z^{i+1} - \vartheta_z^i\right|}{\sqrt{\left(\vartheta_y^{i+1} - \vartheta_y^i\right)^2 + \left(\vartheta_x^{i+1} - \vartheta_x^i\right)^2}}\right) = \tan^{-1}\left(\frac{-\Delta Z}{\sqrt{\Delta X^2 + \Delta Y^2}}\right) \quad (6\text{-}10)$$

$$\begin{cases} u_c = \left|V_C\right|\cos\theta_c\cos\psi_c \\ v_c = \left|V_C\right|\cos\theta_c\sin\psi_c \end{cases} \Rightarrow \begin{aligned} u &= \left|\upsilon\right|\cos\theta\cos\psi + \left|V_C\right|\cos\theta_c\cos\psi_c \\ v &= \left|\upsilon\right|\cos\theta\sin\psi + \left|V_C\right|\cos\theta_c\sin\psi_c \\ w &= \left|\upsilon\right|\sin\theta \end{aligned} \quad (6\text{-}11)$$

$$\langle X(t), Y(t), Z(t)\rangle \approx \left\langle \sum_{i=1}^{n}\left(\vartheta_x^i(t), \vartheta_y^i(t), \vartheta_z^i(t)\right)\right\rangle$$

$$\wp_{x,y,z}^i = \sum_{i=1}^{|\wp|}\sqrt{(\vartheta_x^{i+1} - \vartheta_x^i)^2 + (\vartheta_y^{i+1} - \vartheta_y^i)^2 + (\vartheta_z^{i+1} - \vartheta_z^i)^2} \quad (6\text{-}12)$$

$$\wp(t) = [X(t), Y(t), Z(t), \psi(t), \theta(t), u(t), v(t), w(t)]$$

The $|V_C|$ in (6-11) denotes of the current magnitude, $\psi_c$ and $\theta_c$ are the current vector in horizontal and vertical planes, respectively. The schematic diagram of the AUV path planning process is depicted in Fig 6.3.

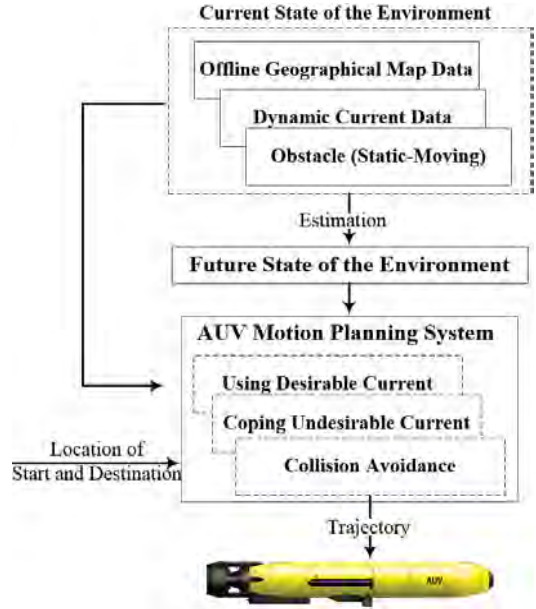

**Fig. 6.3.** The operation diagram of the AUV local motion planning.

### 6.2.1 Path Optimization Criterion

As an NP-hard optimization problem, the motion planner in this study aims to guide the vehicle towards a specific location encountering dynamics of the ocean and kinematics of the vehicle; hence, the following performance indicators need to be addressed:

– **To minimize the travel duration:** to this purpose the shortest path between two points with minimum travel time should be computed while avoiding collision boundaries and adapting ocean current are taken into account. This study assumes AUV to move with constant thrust power, which means the battery consumption for a path is a constant function of the time and distance travelled. Therefore, performance of the computed path is evaluated according to path travel time ($T_\wp$) and distance ($L_\wp$). Assuming that the vehicle moves with a constant water referenced velocity of $|v|$, the $T_\wp$ is estimated as follows:

$$\forall \wp_{x,y,z}^i$$

$$L_\wp = \sum_{i=p_{x,y,z}^s}^{|\wp|} \sqrt{\left(X_{i+1}(t)-X_i(t)\right)^2 + \left(Y_{i+1}(t)-Y_i(t)\right)^2 + \left(Z_{i+1}(t)-Z_i(t)\right)^2} \qquad (6\text{-}13)$$

$$T_\wp = \sum_1^{|\wp|} \frac{L_\wp}{|v|}$$

– **To address the path constraints:** having a safe and reliable deployment is another concern of path optimality in a vast and uncertain environment. The path constraints in this study are associated with bounds on the vehicle states due to the limits of AUV's actuators, and also the environmental factors. Hence, the resultant path should avoid crossing the collision borders of obstacles and forbidden zones of map denoted by ($\sum M, \Theta$). The current force can cause drift to vehicle's desired motion. AUV's directional velocity components of $u$, $v$ should be restricted to $u_{max}$, $[v_{min}, v_{max}]$, and the vehicle should be oriented with $\psi \in [\psi_{min}, \psi_{max}]$, while current magnitude $|V_c|$ and direction $\psi_c$, $\theta_c$ are encountered in all states along the path. The path cost function is defined by (6-14).

$$\wp(t) = [X(t), Y(t), Z(t), \psi(t), \theta(t), u(t), v(t), w(t)]$$

$$\Lambda_{\Sigma M, \Theta} = \begin{cases} 1 & \wp_{x,y,z}(t) = Coast : Map(x, y) = 1 \\ 1 & \wp_{x,y,z}(t) \cap \bigcup_{N\Theta} \Theta(\Theta_p, \Theta_r, \Theta_{Ur}) \\ 0 & Otherwise \end{cases}$$

$$\Lambda_{\wp} = \begin{cases} \Phi_{z\min} \times \min(0; Z(t) - Z_{\min}) \\ \Phi_{z\max} \times \max(0; Z(t) - Z_{\max}) \\ \Phi_u \times \max(0; u(t) - u_{\max}) \\ \Phi_v \times \max(0; |v(t)| - v_{\max}) \\ \Phi_\psi \times \max(0; |\dot{\psi}(t)| - \psi_{\max}) \\ \Phi_{\Sigma M, \Theta} \times \Lambda_{\Sigma M, \Theta} \end{cases}$$

(6-14)

$$C_{\wp} = L_{\wp} + \sum_{i=1}^{n} Q_i f(\Lambda_{\wp})$$

The $Q_i f(\Lambda_{\wp})$ in (6-14) is a weighted violation function that respects the AUV Kino-dynamic constraints of surge ($u$), sway ($v$), yaw ($\psi$); depth constraint of ($Z$), to prevent the path from straying outside the vertical operating borders; and collision constraints of ($\Lambda_{\Sigma M, \Theta}$) to prevent the path from collision danger. The $\Phi_{z\min}$, $\Phi_{z\max}$, $\Phi_u$, $\Phi_v$, $\Phi_\psi$, and $\Phi_{\Sigma M, \Theta}$ respectively denote the impact of each constraint violation in calculation of total path cost $C_{\wp}$.

## 6.2.2 Online Re-planning based on Previous Solution

The motion planner entails a heuristic search for refining the path considering situational awareness of environment; thus it needs to accommodate the unforeseen anomalies. Having an online path re-planning strategy advances the vehicle to incorporate dynamic and reactive behaviors in dealing with prompt changes of the operation field. In this process, the vehicle receives the existing situation of the environment and refines the path until the specified target waypoint is met. Advancing the path planner with such a re-planning mechanism enables adapting the new current patterns. Thereafter, using a penalty function, an augmented cost function embodied all constraints as given by (6-14). In the utilized mechanism, when the path re-planning flag is triggered, the new optimal path is calculated from current position to the destination by refining the previous solution and replacing the current states of the vehicle with the new boundary conditions. In fact, this strategy provides a closed-loop guidance configuration; that can be supported with a minimal computational burden, as there is no need to compute the path from the scratch, as opposed to reactive planning strategy [6]. This process is summarized in Figs 6.4 and 6.5.

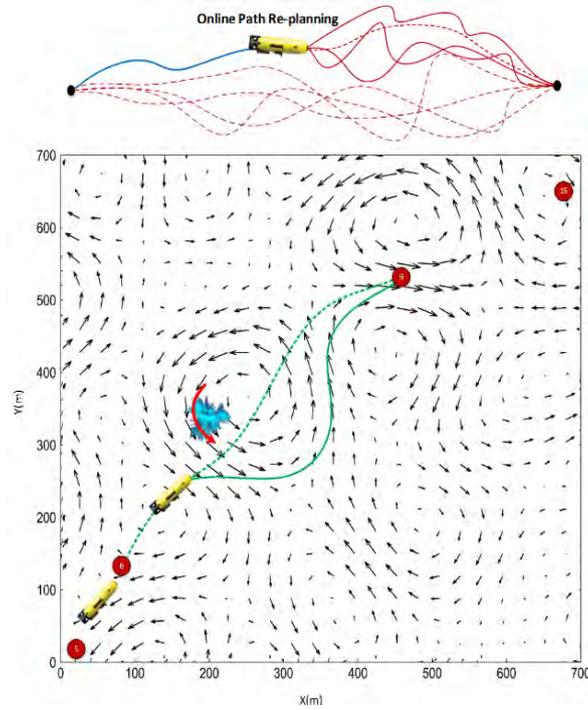

**Fig. 6.4.** Online path re-planning, which reuses the subset of the past solutions.

**Online Path Re-Planning Mechanism**
– Get the position of start and target points
– Generate the initial optimum path using Eq.(6-9) to (6-12)
– Evaluate the generated path by defined $C_{iP}$ in Eq.(6-14)
– Check Path-Replanning Flag continuously
**BEGIN**
  *If Flag ==1*
    – Apply the current states of the vehicle as the initial condition
    – Set the current location as the start point for local path planning process
    – Use the current solution as an initial solution or initial guess for the proposed evolutionary method
    – Generate a new path
    – Check the problem optimality conditions Eq.(6-14)
    *If* Eq.(6-14) is satisfied
      Follow the new path
      Terminate the planning process
      *else*
      Follow the pervious path
      Terminate the planning process
      Back to **BEGIN**
    *End*
  *End*
**END**

**Fig. 6.5.** Online path re-planning pseudo-code.

## 6.3 Meta-Heuristics Used for Online Motion Planning

There is a significant distinction between theoretical understanding of evolutionary algorithms and their properness for being applied to a problem due to the difference in problem's size, complexity, and nature. As mentioned before, we are dealing with a kind of constraint optimization problem, while the basis spline curves are exploited to parameterize the initial path solutions. Accordingly, the path curve is captured from a set of control points like $\vartheta = \{\vartheta_1, \vartheta_2, ..., \vartheta_i..., \vartheta_n\}$ in the problem space with Cartesian coordinates of $\vartheta_1:(x_1,y_1,z_1),...,\vartheta_n:(x_n,y_n,z_n)$, where $n$ corresponds to the number of control points. Appropriate placement of these points is a substantial part of the optimization problem that affects optimality of solutions. Selecting a proper set of algorithm and precise setting of their functionalities according to the properties of a particular problem is the main step of the implementation. Similar to the Chapter 5, the same meta-heuristics of DE, PSO, and BBO are employed for dealing with the NP-hard complexity of motion planning in this chapter. A brief overview of the applied algorithms has been provided by Chapter 5. Figure 6.6 explains the mechanism of the DE, PSO, and BBO on motion planning.

- **DE:** The DE is a modified version of the GA algorithm and applies similar evolution operators (mutation, crossover, selection) [7]. The population of $\wp^i_{x,y,z}$ is initialized with solution vectors $\chi_i$, $(i=1,..., i_{max})$, in which any path control point of $\vartheta_i$ correspond to elements of $\chi_i^{x,y,z}$ vector. The DE employs floating-point real numbers, non-uniform crossover and differential mutation that improves quality of the solution and provides faster computation. The mutation operator in DE picks three random solution vector like $\chi_{r1,t}$, $\chi_{r2,t}$, and $\chi_{r3,t}$ in each iteration and uses their difference $(\chi_{r1,t}\text{-}\chi_{r2,t})$ to mutate any candidate solution vector. The *donor* is one of the randomly selected triplets gets used in mutation to accelerate the convergence rate. The new generation gets evaluated applying the cost function in (6-14), and best-fitted candidate solutions with the minimum cost get transferred to the next generation. This procedure is repeated iteratively until converging the population to the optimum solution [8]. The average fitness of the population gets improved at each iteration by adaptive heuristic search nature of the DE. The operation is ended when the iterations get completed, or when no dramatic change is observed in population evolution.

- **PSO:** The PSO is well suited to solve continuous natured problems (general explanation of PSO can be found in Chapter 5). In the motion planning problem, the particles are coded by the potential paths. The position and velocity parameters of the particles correspond to the co-ordinates of the spline control points ($\vartheta$) utilized in path generation. The process of PSO optimization is summarized in Fig 6.6. The most attractive benefit of the PSO is that it is easy to implement and it requires less computation to converge to the optimum solution.

– **BBO:** The BBO is an evolution-based technique that mimics the equilibrium pattern of inhabitancy in biogeographical island [9]. This algorithm has strong exploitation ability as the habitats never are eliminated from the population but are improved by migration. In the proposed BBO based motion planner, each habitat $h_i$ corresponds to the coordinates of the spline control point's $\vartheta_i$, where $h_i$ defined as a parameter to be optimized ($P_s$:($h_1, h_2, ..., h_{n-1}$)). Habitats are improved iteratively as explained in Fig 6.6.

---

**Procedure of DE on Motion Planning**

- Initialize population of solution vectors randomly $\chi_i^{x,y,z}$ with the control points ($\theta_i^x, \theta_i^y, \theta_i^z$)
- Set the maximum number of iteration $t_{max}$
- Choose appropriate parameters for the population size $i_{max}$
- Set the crossover coefficient $r_C \in [0,1]$
- Set the scaling factor of $S_f$ to keep the balance in difference vector ($\chi_{r1,t}, \chi_{r2,t}$)

   **For** $t = 1$ **to** $t_{max}$
      **For** $i = 1$ **to** $i_{max}$

         Reconstruct a path according to:   $\chi_{i,t}\left(\chi_{i,t}^x, \chi_{i,t}^y, \chi_{i,t}^z\right) = \left(\theta_i^x(t), \theta_i^y(t), \theta_i^z(t)\right)$

         Evaluate the path using the cost function $C_\wp$ given in (6-14)

         Determine the *donor:*    $donor = \sum_{j=1}^{3}\left(\lambda_j \Big/ \sum_{j=1}^{3}\lambda_j\right)\chi_{r3,G}; \quad \lambda_j \in [0,1]$

         Apply mutation using:    $\check{\chi}_{i,t} = \chi_{r3,t} + S_f(\chi_{r1,t} - \chi_{r2,t})$
                                 $r1, r2, r3 \in \{1,..., i_{max}\}, \quad r1 \neq r2 \neq r3 \neq i, \; S_f \in [0,1+]$

         **For** $j = 1$ **to** $i$
            Apply crossover using $\chi_{i,t}$, and the mutant solution $\chi'_{i,t}$ to get the $\chi''_{i,t}$

                $\check{\chi}_{j,i,t} = \begin{cases} \check{\chi}_{j,i,t} & rand_j \leq r_C \vee j = k \\ \chi_{j,i,t} & rand_j \leq r_C \wedge j \neq k \end{cases} \begin{cases} j = 1,..., n \\ n \in [1, t_{max}] \end{cases}$

            Reconstruct a path according to $\chi_{i,t}, \chi'_{i,t}$ and $\chi''_{i,t}$
            Evaluate the corresponding paths to $\chi_{i,t}, \chi'_{i,t}$ and $\chi''_{i,t}$
            **if** $C_\wp(\chi_{i,t}) \leq C_\wp(\chi'_{i,t})$
                $\chi'_{i,t+1} = \chi_{i,t}$
                **if** $C_\wp(\chi_{i,t}) \leq C_\wp(\chi''_{i,t})$
                    $\chi'_{i,t+1} = \chi_{i,t}$
                **else**
                    $\chi'_{i,t+1} = \check{\chi}''_{i,t}$
                **end (if)**
            **else**
                $\chi'_{i,t+1} = \chi'_{i,t}$
            **end (if)**
         **end (For)**
      **end (For)**
      Select the best solutions to transfer to next iteration
   **end (For)**
Output best solution and the corresponding paths

---

**Procedure of PSO on Motion Planning**

Initialize each particle by random velocity and position in following steps:

- Assign Spline control points $\vartheta_i$ as particle position $\chi_i$
- Initialize each particle with random velocity $\upsilon_i$ in range of predefined bounds $\beta_\vartheta = [U_\vartheta \& L_\vartheta]$.
- Choose appropriate parameters for the population size $i_{max}$
- Set the number of control-points $n$ used to generate the Spline path
- Set the maximum number of iterations $t_{max}$
- Initialize $\chi_i^{P\text{-}best}(1)$ with current position of each particle at first iteration $t=1$.
- Set the $\chi^{G\text{-}best}(1)$ with the best particle in initial population at $t=1$.

*For* $t=1$ *to* $t_{max}$

  Evaluate each candidate particle according to given cost function in (6-14)

  *For* $i=1$ *to* $i_{max}$

    Updated the particles $\chi_i^{P\text{-}best}$ and $\chi^{G\text{-}best}$ at iteration $t$

    *if* $C_\psi(\chi_i(t)) \leq C_\psi(\chi_i^{P\text{-}best}(t\text{-}1))$

      $\chi_i^{P\text{-}best}(t) = \chi_i(t)$

      *else*

      $\chi_i^{P\text{-}best}(t) = \chi_i^{P\text{-}best}(t\text{-}1)$

    *end (if)*

    $\chi^{G\text{-}best}(t) = \underset{1 \leq i}{\arg\min}\, C_\psi(\chi_i^{P\text{-}best}(t))$

    Update the state of the particle in the swarm

    $\upsilon_i(t) = \omega \upsilon_i(t\text{-}1) + c_1 r_1 \left[\chi_i^{P\text{-}best}(t\text{-}1) - \chi_i(t\text{-}1)\right] + c_2 r_2 \left[\chi_i^{G\text{-}best}(t\text{-}1) - \chi_i(t\text{-}1)\right]$

    $\chi_i(t) = \chi_i(t\text{-}1) + \upsilon_i(t)$

    Evaluate each candidate particle $\chi_i$ according to given cost function $C_\psi(\chi_i(t))$

  *end (For)*

  Transfer best particles to next generation

*end (For)*

Output $\chi^{G\text{-}best}$ and its correlated path as the optimal solution

---

**Procedure of BBO on Motion Planning**

Initialize a set of solutions as initial habitat population

- Assign Spline control points $\vartheta_i$ as habitat $h_i$
- Choose appropriate parameters for the population size $i_{max}$
- Set the number of control-points $(n)$ that used to generate the Spline path
- Set the maximum number of iteration $t_{max}$
- Assign maximum immigration and emigration rate $(I_r, E_r)$
- Assign maximum mutation rate $m_{max}(S)$
- Set $S_{max}$ and SIV vector

*For* $t=1$ *to* $t_{max}$

  Compute immigration rates $\lambda$ and emigration rate $\mu$ for each solution

  $\lambda_S = I_r \times (1 - (S/S_{max}))$,    $\mu_S = E_r \times (S/S_{max})$

  Evaluate the fitness (HSI) of each habitat and identify Elite Habitats based on HIS

  Modify habitats based on $\lambda$ and $\mu$ (Migration):

  $P_S(t) = P_S(t\text{-}1)(1 - \lambda_S(t) - \mu_S(t)) + P_{S\text{-}1}\lambda_{S\text{-}1}(t) + P_{S+1}\mu_{S+1}(t)$

  *For* $i=1$ *to* $i_{max}$

    Use $\lambda_i$ to probabilistically decide whether immigrate to habitat $h_i$

    *if* $rand(0,1) < \lambda_i$

    *For* $j=1$ *to* $i_{max}$

      Select the emigrating habitat $h_j$ with probability $\propto \mu_j$

      *if* $rand(0,1) < \mu_j$

        Replace a randomly selected SIV variable of $h_i$ with its corresponding value in $h_j$

      *end (if)*

    *end (For)*

    *end (if)*

  *end (For)*

  Carry out the probability-based mutation: $m(S) = (m_{max} - P_S m_{max})/P_{max}$

  Transfer the best solution in the population from one generation to the next

*end (For)*

Output the best habitat and its correlated path as the optimal solution

---

**Fig. 6.6.** Pseudo-code of DE, PSO and BBO based motion planning.

## 6.4  Motion Planning Simulation Results

With respect to the defined path cost function $C_\wp$, the optimum solution corresponds to quickest and safest path that makes use of desirable currents and cope with adverse current flows to increase battery lifetime, while respecting the vehicular and collision boundaries. The simulation results obtained for the online motion-planning problem will be analyzed in this section. The efficiency of the applied meta-heuristics in satisfying the given objectives and performing adaptive maneuverability of the vehicle in a cluttered dynamic operating field will be investigated through two different scenarios. Each scenario comprises specific aspects of uncertain environment, while the missions get complicated in the second scenario to validate the planner's ability in dealing with different real-world situations.

***Simulation Setup***:

The properties of REMUS underwater vehicle is settled for the purpose of this research, with the 1.6$m$ in length and dimensions of 0.2$m$ in diameter. This vehicle operates with the forward velocity of almost 5 *knots* and dives up to maximum depth of 100$m$. It uses a set of on-board sonars such as Horizontal Acoustic Doppler Current Profiler (HADCP) to measure the updates of operating field including specifications of obstacles and water current. The vehicle is able to carry out a mission distances of approximately 55 $km$. The dynamic current fields in study is computed from a multilayered random distribution of 5 to 8 Lamb vortices in a 350×350 grid, in which the grid size is set according to the pixel density of the clustered geographical map.

The PSO is configured with the population of $i_{max}$=120 particles, acceleration coefficients of *1.8* to *2.5*, and the varying inertia weight in range of [0.5, 1.5]. The BBO is set with the habitats population of $i_{max}$=120; the emigration rate of $\mu \in$[0,1], immigration rate of $\lambda$=1- $\mu$; and the maximum mutation rate of 0.1. Population size of DE is set to be 120, the scaling factor is set on $0.2 \leq S_f \leq 0.8$, and the crossover rate of $r_C$ =40. Iteration number is set on $t_{max}$=100 for all algorithms and number of the path control points ($\vartheta$) for each spline is defined to be $n$=5. The implementation was performed in MATLAB® R2016a on a desktop PC with an Intel i7 3.40 GHz quad-core processor.

***Scenario-1: Motion Planning Considering Ocean Current***

The performance of the proposed motion planners in adapting water currents variations is depicted by Fig 6.7. The water current update rate is set on $U_R^C$=4. The current pattern varies within 4 time steps to model three-dimensional volume of $\Gamma_{3D}$. In order to change the current parameters, Gaussian noise, in a range of 0.15~0.7, is randomly applied to current parameters of $S_i^o$, $\ell$, and $\Im$ given in (6-3). The update of current pattern is illustrated through four subplots in Fig 6.7 (labeled by time steps 1 to 4). The vehicle starts to travel from the red circle toward the target position indicated by a red square.

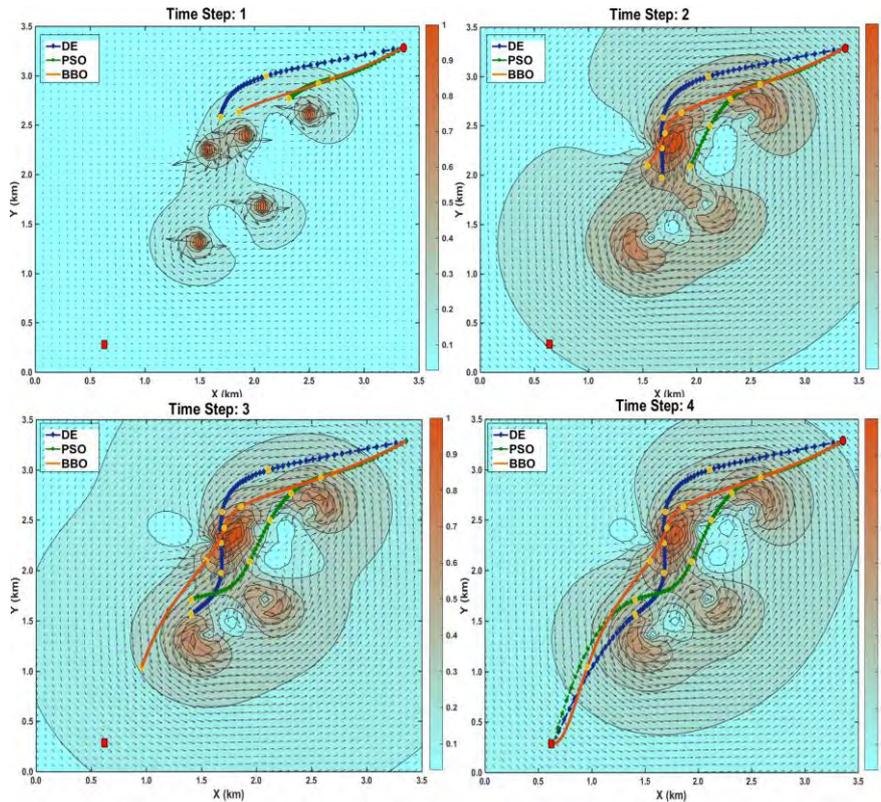

**Fig. 6.7.** Path adaption to the water current variations in 4 time steps.

When the current pattern is updated, the online motion planner receives the new data from HADCP sensors. The path formed between the starting position and the final state in Fig 6.7 shows a great adaptivity to water current behavior in the given time steps for all three algorithms, while the DE performs comparatively better flexing along the favorable current flow. In the second rank, the path generated by PSO shows significant flexibility in coping with current change specifically when the current magnitude gets sharper (is more clear in Time Step: 3). The trend of path orientation in each subplot shows that the planner autonomously adapts current variations and refines the trajectory to avoid severe adverse flow.

### Scenario-2:  Motion Planning Considering Time Varying Current, Geographical Map, and Uncertain Static-Buoyant Obstacles

In the second scenario, the goal is to thoroughly simulate a highly uncertain cluttered environment encountering all possible barriers to assess the performance of the online planner. In this scenario, the clustered map information is used as an underlying environment embodied with variable ocean flows and uncertain buoyant and static objects. The AUV starts to transmit from the starting position denoted by "○"on Fig 6.8 and tends to take the shortest collision-free trajectory to the destination point marked by "□". The current update is shown by two black and blue arrows that contain similar features. The expected time for passing this distance is assumed $T_{\bar{\varepsilon}}$=1800 (sec). The AUV water-reference velocity is set on υ=2.5 (m/s). The collision boundaries represented as black circles around the obstacles indicating a confidence of 98% that the obstacle is located within this area. The gradual increment of collision boundary of each obstacle is presented in Fig 6.8 (b) in which the uncertainty propagation is assumed to be linear with time.

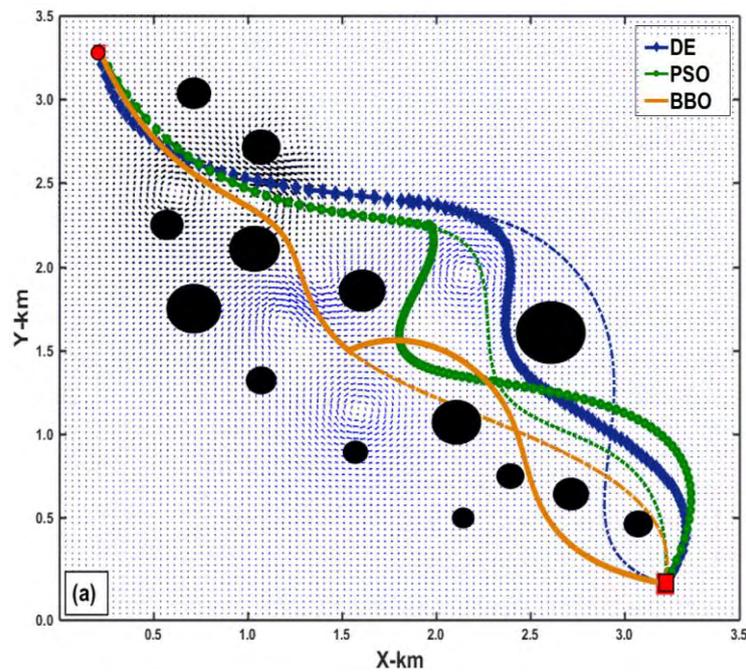

**Fig. 6.8 (a)** The path behaviour and re-planning process in coping current flow and obstacles avoidance encountering a random number of static obstacles;

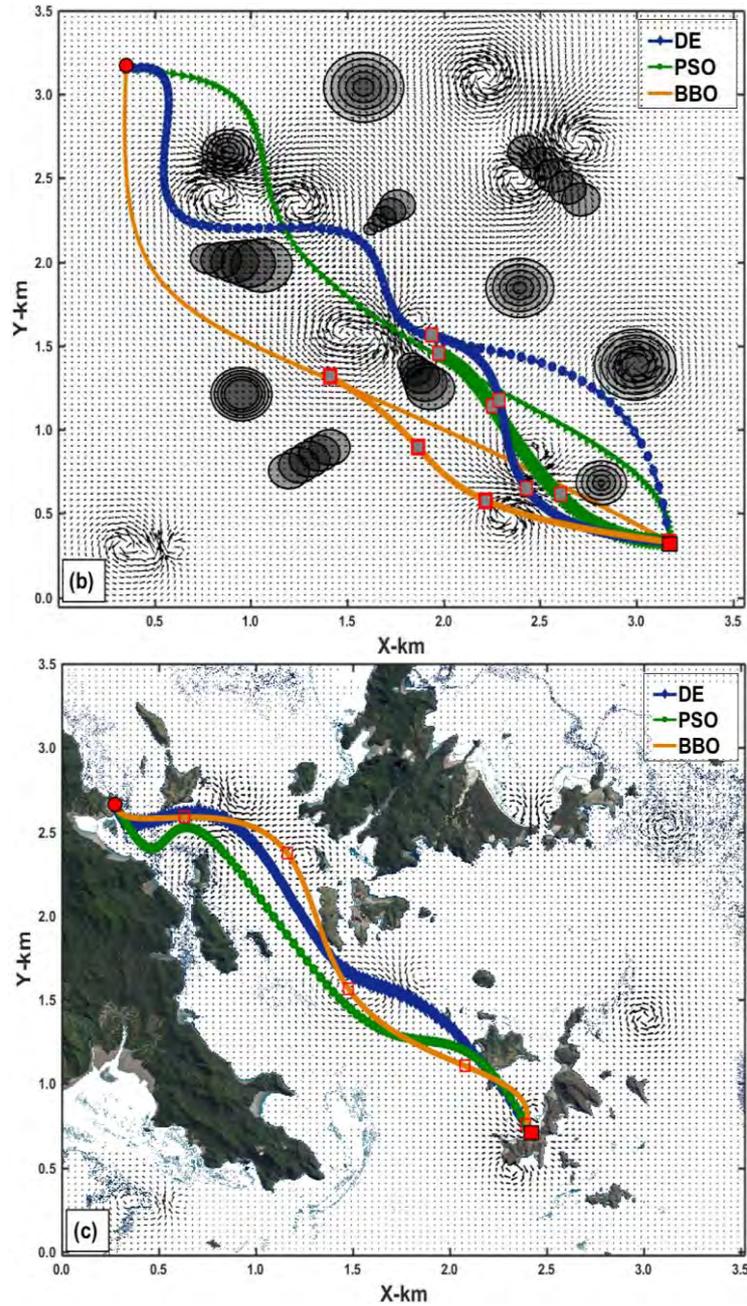

**Fig. 6.8 (b)** Path deformation in handling collision avoidance according to current map updates encountering uncertain static-buoyant objects; **(c)** Accuracy of the proposed path planners in recognizing the map coastal sections.

Increasing the number of obstacles and adding uncertainty into account, increases the problem's complexity; however, it is derived from Fig 6.8 all aforementioned algorithms are able to efficiently carry out the collision avoidance regardless of the terrain complexity. As displayed in Fig 6.8 (b), path re-planning procedure re-generates the alternative trajectory according to the latest update of the current map, in which the refined path takes a detour according to previously generated optimum path (the initial paths are presented by thinner lines and new paths are shown by thicker lines). Further, it is noted from Fig 6.8 (b) the path generated by DE shows superior flexibility in coping with current change. In the implementation of the *k*-means clustering method, the given map is classified into 3 subsections of uncertain risky area, coastal area, and water covered area as valid zone for deployment. It is derived from simulation results in Fig 6.8 (d) that there is only slight difference comparing the results produced by all four algorithms and all of them accurately avoid crossing into the coastal sections of map even when the disturbing current pushes the vehicle to the undesired directions, also they are accurate and resistant against current deformations in the given operation window. To have a clear visualization, Fig 6.9 illustrates 3D-path generated by the evolutionary methods, where in this figure, just the final refined path is indicated. Due to the linear propagation of the uncertainty with time, there exist a collision boundary encircling the objects. The safe trajectory is achieved if the vehicle maneuver does not have any intersection with the proposed obstacle boundary, which is greatly satisfied by all algorithms as can be seen in Fig.9.

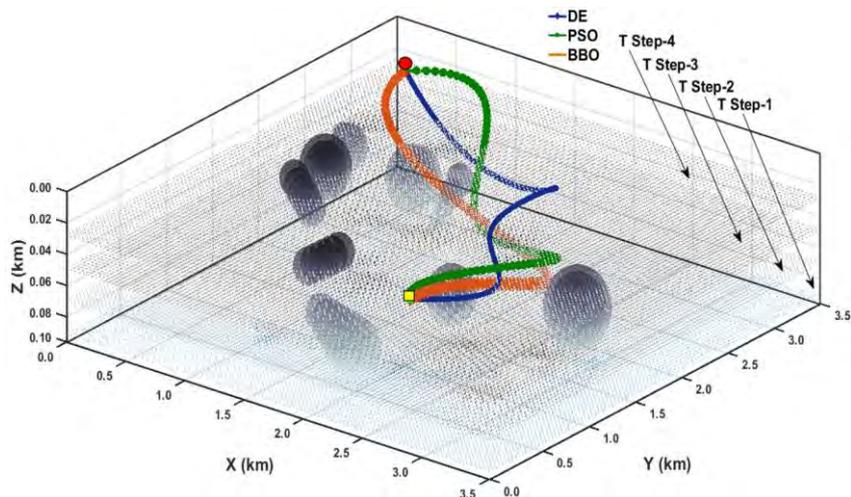

**Fig. 6.9.** Evolution curves of DE, PSO, and BBO algorithms with four step current update in three-dimensional space.

For making a better analysis, the algorithm's cost variation and the rate of solution convergence is shown by Fig 6.10 (a). This plot shows the expedient convergence of the methods in finding optimal solution and sharp rate of eliminating collision violation.

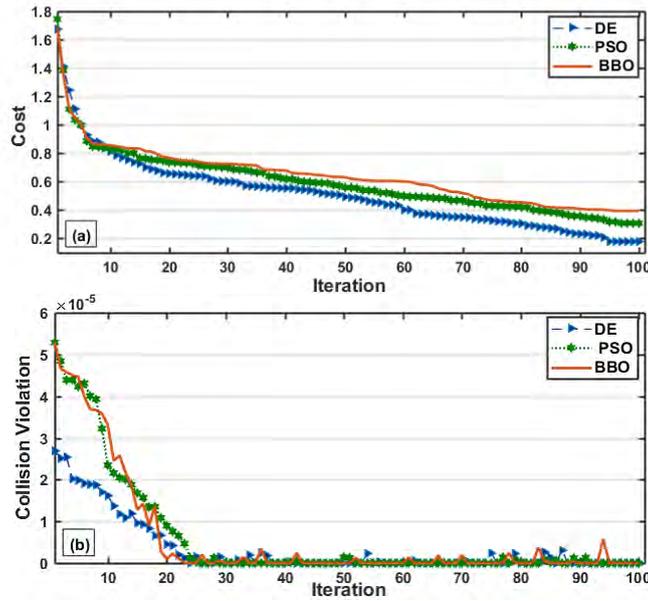

**Fig 6.10 (a)** Iterative cost variations; **(b)** Collision violation for DE, BBO, and PSO algorithms over the 100 iterations.

The cost variations of all four algorithms (given by Fig 6.10(a)) declares that the DE shows superior performance in reducing the cost within 100 iterations, while PSO and BBO show roughly similar performance in optimization. It is also noted from Fig 6.10 (b), the violation of all three algorithms diminishes iteratively, but again the DE starts eliminating the violation earlier than others. It is noteworthy to mention that setting lower rate for Gaussian noise on current vector parameters (i.e., in range of 0.1 to 0.8) leads towards better fitness values for the generated paths by all algorithms. Moreover, by setting higher current update rate (where Gaussian noise parameters get greater value than 0.8), the path fitness value reduces and practically other conditions remaining constant in path execution process.

Figure 6.11 compares the efficiency of the proposed planners to meet the pre-set threshold time ($T_\varepsilon$). The final path time ($T_{fp}$) is different stemming from the inherent differences in structure and mechanism of the proposed methods; however, all of them properly satisfy the assigned upper time threshold of $T_\varepsilon$.

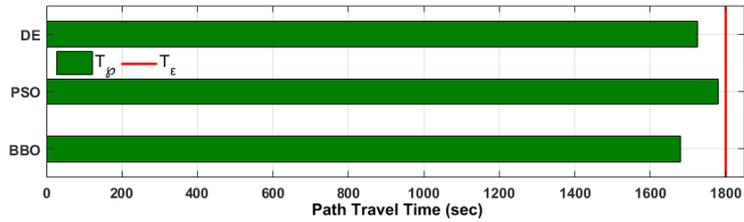

**Fig 6.11.** Performance of the algorithms in respecting the time restriction.

Another concern in motion planning is satisfying the vehicular constraints. The path curvature generated by basis spline should respect the vehicle's radial acceleration and angular velocity constraints, which is investigated thoroughly for all three algorithms and presented in Fig 6.12.

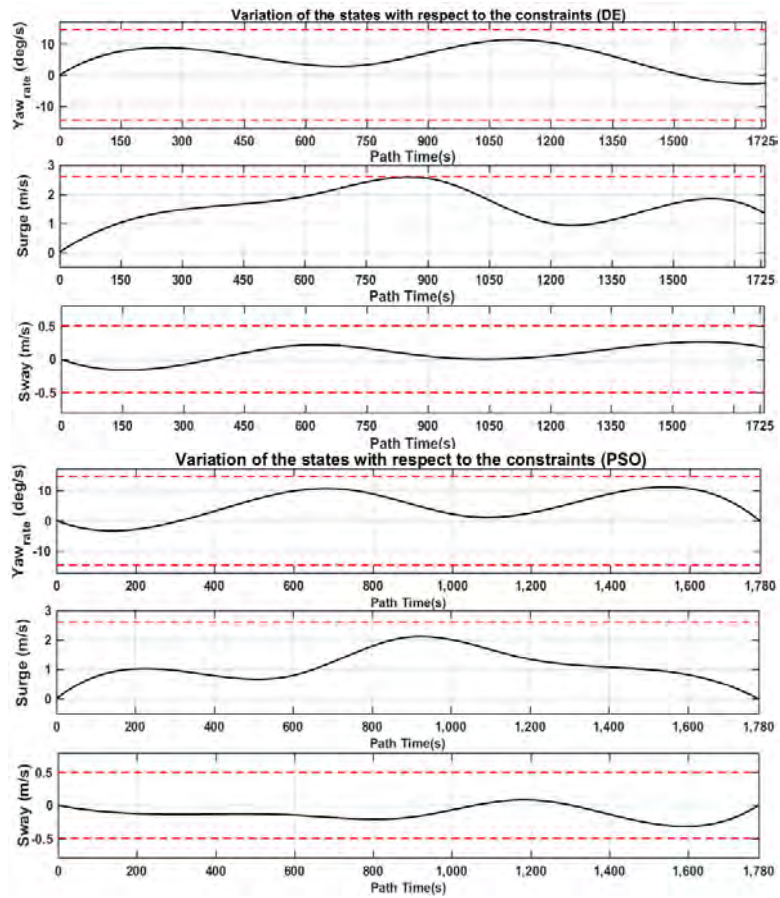

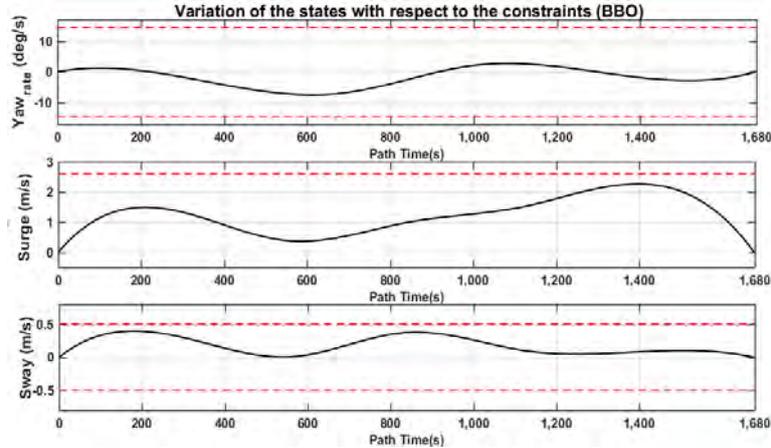

**Fig. 6.12.** Variations of the Surge, Sway, and Yaw rates for all the states across the generated path, where the given vehicular constraints are depicted with red dashed lines.

It is evident in Fig 6.12, all paths accurately satisfy the defined constraints as the variations of surge, sway, and yaw rate along the generated paths by DE, PSO and, BBO are placed between the defined bounds shown with the red dashed line. It is outstanding from the simulation results given by Figs 6.7 to 6.12, all DE, BBO and PSO based motion planners perform almost similar cost and great fitness encountering environmental and vehicular constraints. Being more specific, it is notable that DE acts more efficiently in finding time optimum current-resistant collision-free solutions and performs faster convergence with lower cost comparing to other algorithms. Albeit, it is also noteworthy to mention that performance and optimality of all generated paths by three algorithms are competitive and very similar to each other.

## 6.5 Summary of Chapter

Efficient motion planning is identified as a critical factor to enhance vehicle's autonomy and its resistance against various disturbances. This chapter aims to develop and evaluate an online real-time motion planner as the lower level autonomy provider. On the one hand, existence of buoyant objects may lead to change the path conditions, as time goes on, and on the other hand, variability of ocean current components has considerable impact in drifting the vehicle from desired trajectory. In such a situation, adaptability of the motion planner to environmental variabilities is a key element to carry out the mission safely and successfully. An efficient tra-jectory produced by the motion planner enables an AUV to cope with adverse currents as well as exploit desirable currents to enhance the operation speed that results in considerable energy saving. To have an expedient adaptation with the variations of the underlying environment, an accurate online re-planning mechanism is developed and implemented. We mathematically formulate the realistic scenarios that AUV usually is dealing with in a large-scale undersea terrain. This provides with use of a geographical offline map containing current and obstacles information; furthermore, the uncertainty of environment and navigational sensor suites are carefully taken into account by simulating imperfectness of sensory information in dealing with different types of buoyant objects. The online motion planner simultaneously tracks the measurements of the terrain status and tends to generate the shortest collision-free trajectory by extracting eligible areas of map for vehicles deployment, carrying out obstacle avoidance and coping water current disturbance. The proposed planner is capable of refining the original path considering the update of current flows, uncertain static and moving obstacles. This refinement is not computationally expensive as there is no need to compute the path from scratch and the obtained solutions of the original path is utilized as the initial solutions for the employed methods. This leads to reduction of optimization space and acceleration of the searching process.

Having significant flexibility for approximating complex trajectories, basis spline curves are utilized to parameterize the desired path. Applying the evolutionary methods namely PSO, BBO, and DE indicate that they are capable of satisfying the motion planning problem's conditions. Considering the results of simulation, performance comparison between all applied methods is undertaken. From the results given by Section 6.4, it can be concluded that all DE, BBO and PSO based planners perform with almost similar cost and great fitness encountering collision, vehicular, and all other constraints for the corresponding numbers of iteration. From comparison point of view, the DE-based planner shows better performance in terms of making use of favorable current flow and maneuverability in collision avoidance, while BBO acts better in terms of time optimality. In summary, the simulation results confirm that the proposed online planning approach using all three meta-heuristic is efficient and fast enough in generating optimal and collision-free path encountering dynamicity of the uncertain operating field that results in leveraging the autonomy of the vehicle for having a successful mission.

In the next chapter, an augmented reactive mission planning architecture will be designed to accommodate the autonomy in both high-level decision-making and low-level action generator by simultaneous mission scheduling (ordering the tasks), and efficient motion planning to provide safe and reliable maneuver.

# Chapter 7

# Augmented Reactive Mission Planning Architecture

S. MahmoudZadeh[1], R. Bairam Zadeh

[1] Faculty of IT, Monash University, Clayton, VIC 3800, Australia
Email: Somaiyeh.mahmoudzadeh@monah.edu

**Abstract.** Advancing the decision autonomy is a real challenge in the development of today AUVs as their operation is still restricted to very particular tasks that usually supervised by the human operator(s). Having a robust decision-making system along with an accurate motion planning mechanism facilitates a single vehicle to manage its restricted energy resources and endurance times toward accomplishing various complex tasks in a single mission while accompanying any immediate changes of a highly uncertain environment. The proceeding approach builds on recent two chapters towards developing a comprehensive structure for AUV mission planning, task-time managing, routing, and synchronic online motion planning adaptive to sudden changes of the time-variant marine environment. To this end, the following objectives are defined to approach the mentioned above expectations:

- To augment the mission planner with a real time motion planner;

- To accommodate a concurrent operation and synchronization among mission and motion planners;

- To split a large-scaled terrain to smaller efficient operational windows, which results in reducing the computational burden of motion planning system;

- To detect anomalies and compensate any lost time during the motion re-planning process;

- Advancing the system with a synchronous re-scheduling mechanism to manage mission time and reprioritizing the tasks;

This chapter introduces an "Augmented Reactive Mission Planning Architecture" (ARMPA) and exercises DE meta-heuristic algorithm in layers of the proposed control architecture to investigate the efficiency of the structure in addressing the given objectives and ensuring the stability of ARMPA performance in real-time task-time-threat management. Numeri-cal simulations for analysis of different situations of the real-world environment is accomplished separately for each layer and also for the entire ARMPA model at the end.

## 7.1 Motivation: Why and How the ARMPA Can Leverage the Level of Autonomy?

With respect the challenges and shortcomings associated with the mission-motion planning strategies, which have been comprehensively discussed in Chapters 2 and 3, many technical issues still remained unaddressed and the existing approaches mainly addressed only a specific aspect of autonomy either in task-time management together with routing or the motion planning with safety considerations. The proposed ARMPA control architecture augments a reactive mission planner with a local real-time motion planning approach and some specific configurations to produce a reliable time optimum mission plan including maximum possible completed tasks and to adapt unforeseen situations of an uncertain terrain, which aims in addressing the shortcomings associated with each of the above-mentioned strategies.

The mission planning module at the top layer is responsible for the operation timing and the task-assign process in a confined time interval while guiding the AUV towards a target of interest. The motion planning module at the inner layer is in charge of generating safe and optimal trajectories between the waypoints and continuously refining the path to cope with local changes and uncertainties. According to the performance of the inner layer, the mission planner decides to re-organize tasks order and sequence to reactively compensate any possible lost time in the current state of the operation. As a result, the vehicle would be able to undertake the maximum number of tasks with a certain degree of maneuverability having situational awareness of the operating field. Constant interaction is streaming between high-low level modules and the ARMPA decides whether to carry out the motion re-planning or mission re-planning procedure according to the raised situation. The proposed re-planning mechanism in both of the layers improves the robustness and reactive-ability of the AUV in dealing with unexpected circumstances and accurate mission timing. The operation diagram of the ARMPA is depicted in Fig 7.1.

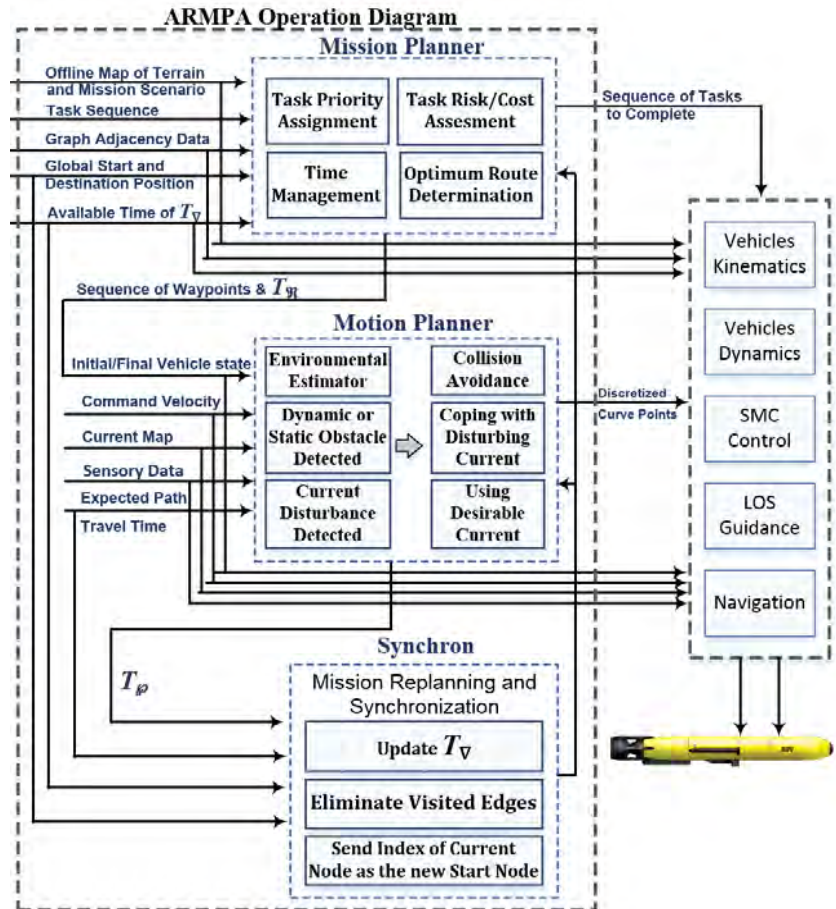

**Fig. 7.1.** The ARMPA operation diagram and relation between the engaged modules.

The ARMPA offers a degree of flexibility for employing diverse sorts of algorithms subjected to real-time performance of them. Both of the planners operate individually and concurrently while sharing their information. Parallel execution of the planners accelerates the computation process. The operating field for the motion planner is split into smaller spaces between pairs of waypoints; thus, re-planning a new trajectory requires rendering and re-computing less information. Moreover, the modular structure of the ARMPA provides a reusable and versatile framework that, at first, is easily upgradable and secondly is applicable to a broad group of autonomous vehicles.

## 7.2 The 'Synchron' Module and ARMPA Modular Mechanism

The ARMPA is designed in separate modules running concurrently. The modules interact simultaneously by back feeding the situational awareness of the surrounding operating field; accordingly, the system decides to re-plan the mission, or just refine the trajectory, or continue the current operation. Hence, the "Synchron" module is designed to manage the lost time within the layers concurrent process and reactively adapt the system to the latest updates of surroundings and the decision parameters (e.g., residual operation time). This process is constantly replicated until the AUV reaches the endpoint (success) or runs out of battery (failure).

After a mission plan is generated in the context of tasks/waypoints sequence, the order of engaged waypoints and the terrain information are passed to the motion planning module, and this module starts generating safe and time-efficient trajectories between the waypoints while the AUV concurrently executes the corresponding tasks assigned to the distance between waypoints. The planned trajectory is shifted to the guidance controller to produce the guidance commands. During deployment between two waypoints, the inner layer can incorporate any possible environmental changes. Base on the received sensory information, the motion planner repeatedly calculates the trajectory between vehicles current position and its specified target location and the total path time of $T_{\wp}^{ij}$ gets compared to the expected travel time of $T_{\varepsilon}^{ij}$ after a waypoint is visited. The $T_{\varepsilon}^{ij} \approx t_{ij}$ is the expected time for traversing the corresponding distance of $d_{ij}$, which is extracted from the mission time of $T_{\Re}$ given by (5-4). In a case that the $T_{\wp}^{ij}$ is found to be smaller than $T_{\varepsilon}^{ij}$, the current order of tasks would be valid, and the vehicle can continue its operation. In contrast, when the $T_{\wp}^{ij}$ exceeds the $T_{\varepsilon}^{ij}$, means that a certain amount of the total time $T_{\nabla}$ is spent for handling any probable issue in motion planning process; thus, the previously defined mission scenario cannot be valid to the time constraint anymore. In such a conditions mission re-planning would be necessary, which this process also has a computational burden. Moreover multiple traversing a specific distance means repeating a task for multiple times which is a time dissipation itself. To handle these complications, the following steps should be carried out at mission re-planning:

- Update $T_{\nabla}$;

- Discard the passed distances from the operation network (so the search space shrinks);

- Set the location of the current waypoint as the new starting point for both planners;

- Recall mission planning module to reorder the tasks and to prepare a new mission scenario based on the updated situation and the new network topology. A computation cost is to be encountered any time that mission re-planning is required.

- This process is to be continued until the vehicle arrives in the defined destination.

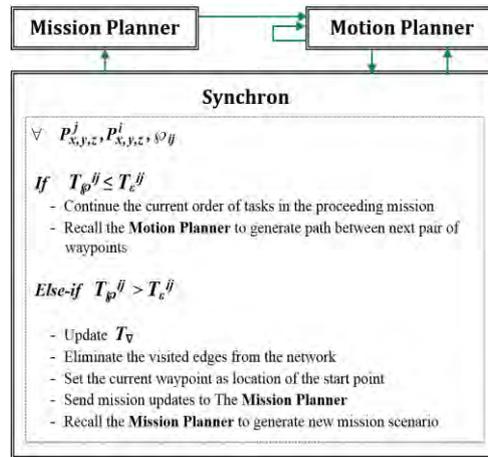

**Fig 7.2.** The synchronization process.

The trade-off between the mission objectives and total available time is a critical process that should be adaptively carried out by the mission planner. Hence, the ARMPA should be fast enough to track the changes, understand the situation, propose a solution to cope with the raised situation and carry out a prompt mission/motion re-planning based on the request. The proposed synchronous architecture is evaluated according to criterion given in the next sub-section.

## 7.2.1 ARMPA Optimization Criterion

Fast and concurrent operation of the subsystems is the most critical concern in preserving the stability and consistency of the proposed ARMPA architecture that prevents each of planning modules from dropping behind the other. Any delay in operation of each layer disrupts the concurrency of the entire system. The motion planner operates concurrently in background of the mission planning scheme. The generated local path is assessed according to the path cost function of $C_{\wp}$ and is directly used in the context of total mission cost calculation ($C_\Re$). The ARMPA aims to guide the vehicle toward endpoint by selecting the best order of highest priority tasks subjected to the time threshold of $T_\nabla$ while guaranteeing a safe collision-free operation. The entire ARMPA is searching for a beneficial solution in the sense of the best combination of task, path, and time as given by (7-1) and (7-2):

$$\forall\, \wp_{x,y,z}^{j} = [X(t),Y(t),Z(t),\psi(t),\theta(t),u(t),v(t),w(t)]$$

$$C_\wp \approx T_\wp = \sum_{i=P_{x,y,z}^{g}}^{|\wp_{x,y,z}|} \frac{\sqrt{(X_{i+1}-X_i)^2+(Y_{i+1}-Y_i)^2+(Z_{i+1}-Z_i)^2}}{|v|}$$

$$s.t.$$
$$[X(t),Y(t),Z(t)] \cap \Sigma_{M,\Theta} = 0$$
$$u(t) \leq u_{\max} \quad \& \quad v_{\min} \leq v(t) \leq v_{\max}$$
$$\theta(t) \leq \theta_{\max} \quad \& \quad \dot\psi_{\min} \leq \dot\psi(t) \leq \dot\psi_{\max}$$
$$d_{ij} \propto \wp \quad \& \quad t_{ij} \propto T_\wp$$
$$T_\Re = \sum_{\substack{i=0 \\ j \neq i}}^{n} s_{eij} \times t_{ij}$$

$$C_\Re \propto |T_\Re - T_\nabla| = \left| \sum_{\substack{i=0 \\ j \neq i}}^{n} s_{eij} e_{ij} \times (C_\wp^{ij} + \wp_{\mathrm{CPU}}) - T_\nabla \right| + \sum_{\substack{i=0 \\ j \neq i}}^{n} s_{eij} \times \frac{e_{ij}}{w_{ij}}; \quad s_{eij} \in \{0,1\}$$

$$s.t.$$
$$\max(T_\Re) < T_\nabla$$

The $C_\wp$ is subjected to vehicles Kino-dynamic and collisions boundaries and is encountered in total cost ($C_\nabla$) computation. The mission re-planning criteria (given by Figure 7.2) is investigated after the completion of the motion planning process. Therefore, the computation time is also taken into account of total cost computation as follows:

$$C_\nabla = C_\Re \times f(C_\wp) + \sum \wp_{\mathrm{CPU}} + \sum_{1}^{rep} T_{compute} \times (\Re_{\mathrm{CPU}}) \tag{7-3}$$

The *rep* and $T_{compute}$ denote the mission re-planning repetition number and computation time, respectively. The DE algorithm is employed by planning modules to evaluate the performance of the designed ARMPA architecture in this chapter. The meta-heuristics usually have two inner loops through the population size of $i_{max}$ and iterations of $t_{max}$ (refer to pseudocodes in previous chapters); thus, in the worst case, their computational complexity is O($i_{max}^2 \times t_{max}$), which is linear with time. The performance of applied algorithm is statistically analyzed through the series of Monte Carlo simulations.

## 7.3 ARMPA Simulation Results

The designed model aims to make the best use of the total available time, which is computed according to the vehicles battery capacity, for accommodating the best and efficient order of tasks, and guaranteeing a secure deployment and on-time arrival to the destination. For preserving the coherence of the system and the components involved, an accurate and consonant collaboration should be provided between the components of the model so that each module can accomplish its responsibility synchronously. Assumptions play an essential role in the performance of the model which is discussed in the following sections.

### 7.3.1 Assumptions and Simulation Setup

A model of the maritime environment is implemented to assess the performance of the proposed framework in providing a comprehensive control on mission timing, multi-tasking, safe and efficient deployment in a waypoint-based, cluttered, uncertain, and varying underwater environment. For this purpose, a 3D volume in dimensions of $\{10{\times}10\ km\ (x\text{-}y),\ 100\ m(z)\}$ is considered based on a realistic example geographical map, where a $K$-means clustering is applied to classify the coast, water and uncertain zones of the map. The map is fed to the clustering method as an image of $1000{\times}1000$ pixels corresponding to $10{\times}10\ km$ square area for the mission planning module and a sub-map of $350{\times}350$ pixels corresponding to the area of $3.5{\times}3.5\ km$ square for the motion planning module, where each pixel of the image corresponds to $10{\times}10\ m$ square space. The clustered image is then transformed to a 2D matrix in the same size of image pixel density. The matrix is filled with values in [0, 1] interval, in which the corresponding coastal sections filled by $\{0\}$, the uncertain sections filled by a value in $(0,0.8]$, and the water sections filled with $\{1\}$.

The operation field is populated by a number of waypoints and tasks are distributed to the waypoints interval. In the produced network, every connection (waypoints' interval) is assigned with a weight and cost calculated according to tasks' priority, completion time, length of the connection and the expected time for passing the interval. Quantity of the waypoints is set with a random number of nodes in [30, 50] that is calculated with a uniform distribution and the topology of the network also transforms randomly with a Gaussian distribution on the problem search space. The position of the waypoints are randomized according to $\{p^i_{x,y}{\sim}\mathbf{U}(0,10000)$ and $p^i_z{\sim}\mathbf{U}(0,100)\}$ in the water-covered area. The speci-fied tasks are randomly assigned to some of the connections so that those connections that assigned with no task are weighted with value of {1} to be neutral in cost calculation.

In addition to offline map, different types of static and buoyant obstacles are taken into account in order to model different possibilities of the real world situations. Obstacle's velocity, dimension and coordinates can be measured by the sonar sensors with a specific uncertainty that is modelled with a Gaussian distribution; hence, each obstacle is presentable by its position ($\Theta_p$), dimension (diameter $\Theta_r$) and uncertainty ratio. Position of the obstacles are initialized using normal distribution of $\mathbf{N} \sim (0, \sigma^2)$ bounded to position of two targeted waypoints by the path planner (e.g. $p^a_{x,y,z} < \Theta_p < p^b_{x,y,z}$), where $\sigma^2 \approx \Theta_r$. Refer to chapter 6 for further details on modelling of different obstacles.

Beside the uncertainty of operating field, water current influence on motion planning needs to be addressed thoroughly in accordance with the type and the range of the mission. For the purpose of this study, a 3D turbulent time-varying current is modelled using a multiple layered 2D current maps, in which the current circulation patterns gradually change in depth. It is assumed the AUV is traveling with constant velocity $|v| \sim 5$ *knots* while the boundary conditions on AUV actuators and state also are taken into account. Therefore, the path violation function is defined as a combination of the vehicle's depth, surge, sway, yaw and collision violations as follows: the $z_{min}=0(m)$; $z_{max}=100(m)$; $u_{max}=2.7(m/s)$; $v_{min}=-0.5(m/s)$; $v_{max}=0.5(m/s)$; $\psi_{min}=-17$ (*deg*) and $\psi_{max}=17$ (*deg*). The proposed autonomous/reactive system is implemented in MATLAB®2016 on a desktop PC with an Intel i7 3.40 GHz quad-core processor and its performance is statistically analysed.

### 7.3.2 Performance of the ARMPA on Mission Time Management for Testing the Reliability of the Operations

The performance of the model is independent employing a diverse set of meta-heuristics algorithms and the ARMPA can perfectly synchronize components to have a stable outcome. The model owe this advantage to joining two disparate perspective of vehicle's autonomy in high-level task-time management and low-level self/environment awareness. The DE is selected among tested algorithms in the previous chapters, and is employed by mission-motion planning modules to test the overall performance of the ARMPA. In this context, the higher layer of mission planner is expected to be capable of generating a time efficient route with best sequence of tasks to ensure the AUV has a beneficial and on-time operation. A beneficial operation is a mission that covers maximum possible number of tasks in a manner that total obtained weight by a route and its travel time is maximized but not exceed the time threshold. In the inner layer, the motion planner is expected be fast enough to rapidly react to prompt changes of the environment and generate an alternative trajectory that safely guides the vehicle through the specified waypoints. Hence, the performance and stability of the model in satisfying the metrics of "obtained weight", "number of completed tasks", and "cost of the $C_\wp$, $C_\Re$" is investigated and statistically analyzed through the 30 Monte Carlo simulation runs (presented by Figures 7.3 to 7.7).

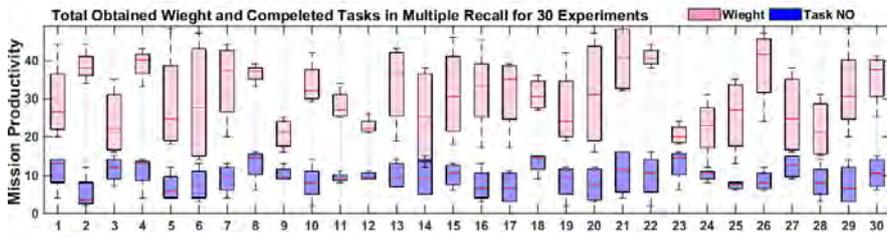

**Fig. 7.3.** Statistical analysis of the model in terms of mission productivity in 30 Monte Carlo simulations.

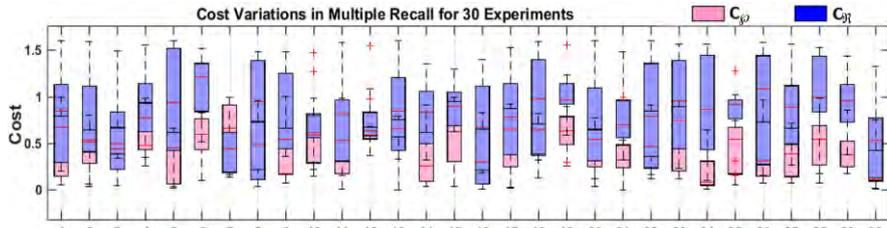

**Fig. 7.4.** The cost variation for both mission and motion planners in 30 Monte Carlo simulations.

The cost variation for both mission and motion planners is shown by Fig 7.4 as a metrics of the models robustness and stability in producing optimal solutions. It is outstanding from Fig 7.4, the cost variation for several recalls of the both planners stands in a specific range for all 30 experiments (mission). It is also notable from the analysis of the results in Fig 7.3 that the generated solutions have a consistent distribution of number of completed tasks and the total obtained weight despite of the problem space deformation. This indicates the system's stability in maintaining beneficence of the mission in a reasonable range when dealing with environmental changes and random deformation of the topology of operation network.

Another concern in keeping the coherency of the system is to have proper coordination between modules. Hence, having a short computation-al time for both of the planners is essential to provide a concurrent synchronization. To address this performance metric a numerical analysis of the computational time is provided over the 30 experiments as shown in Fig 7.5.

The synchronism of the whole system is dependent on the compatibility of the path time ($T_{\wp}$) and the expected time ($T_{\varepsilon}$) in multiple operations of the motion planner. Therefore, a reasonable correlation between $T_{\wp}$ and $T_{\varepsilon}$ values should exist to prevent any interruption in cohesion of the whole system, as it is critical for recognizing the requisition for mission replanning. The performance of the model in satisfying the given performance indicator is quantitatively investigated and depicted in Fig 7.6.

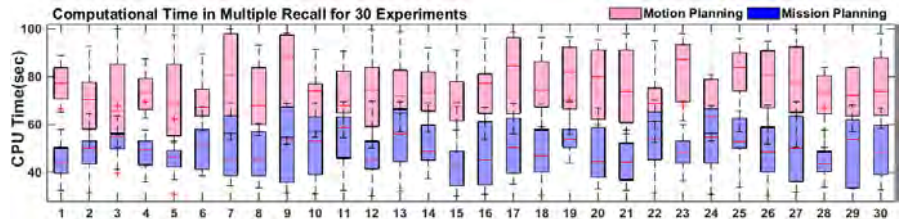

**Fig. 7.5.** Statistical analysis of the model in terms of computational time for both mission and motion planners in 30 Monte Carlo simulations.

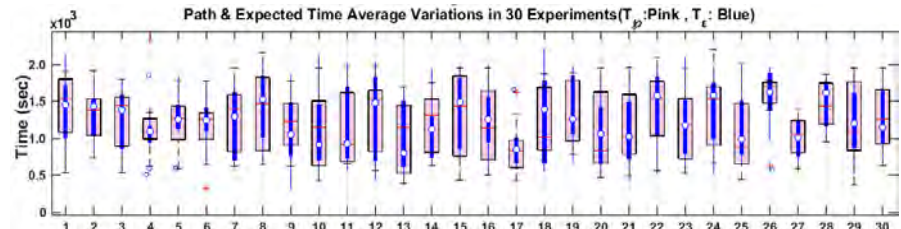

**Fig. 7.6.** Real-time performance of the model, and compatibility of the $T_{\wp}$ and $T_{\varepsilon}$ in multiple operations of the motion planner.

It is outstanding in Fig 7.6 that variations of the $T_{\wp}$ (presented with pink transparent boxplot) and the $T_{\varepsilon}$ (is shown with blue compact boxplot) are settled in a similar range very close to each other. Figure 7.5 also presents the relative proportions of CPU time used for multiple runs of the motion planner compared to CPU time for mission planner operations. Analysis of the given results in Fig 7.5 declares that the modules take a very short CPU time for all experiments that makes it highly appropriate for real-time application. The variations of CPU time for both planners is almost in similar range in all experiments that proves the inherent stability of the model's real-time performance.

Although the system aims to take the maximum use of the total available time ($T_\nabla$) to enhance the mission productivity, the system should be able to manage tasks' timing to prevent overstepping the time threshold of $T_\nabla$ and to guarantee on-time mission completion. To this end, a quantitative analysis of the ARMPA behavior of time management, as the most important indicator for systems robustness and stability, is provided over the 30 individual missions with the same initial condition that closely matches actual underwater scenarios, presented in Fig 7.7.

$$T_{Remained} = T_\nabla - T_\Re \qquad (7\text{-}4)$$

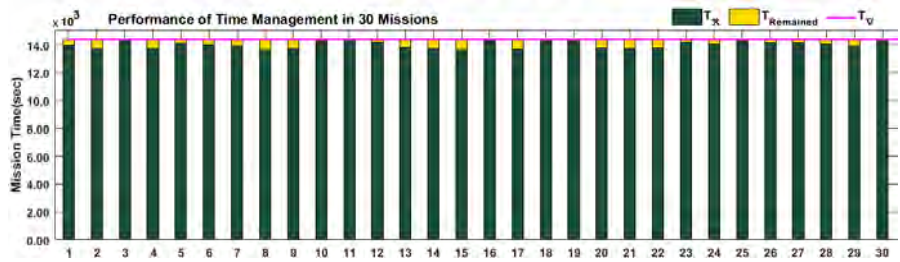

**Fig 7.7.** ARMPA performance of mission time management in 30 missions.

The most effective performance indicator for this model is its accuracy in mission time management. The mission time ($T_\Re$) is the summation of time taken for passing through the waypoints and completion time for included tasks in a mission. The mission should be completed with a minimum residual time ($T_{Remained}$) which has a conversely linear relation to $T_\Re$ according to (7-4). As can be seen in Fig 7.7, the proposed reactive ARMPA system is capable of making full use of the available time as the $T_\Re$ in all experiments approaches the $T_\nabla$ and met the above constraints denoted by the upper bound of $T_\nabla=14.4\times10^3(sec)$ (presented by the pink line). Obviously, minimization of the $T_{Remained}$ represents that how much of the $T_\nabla$ is used for completing different tasks in each mission. The $T_{Remained}$ should not cross the $T_\nabla$ line that is accurately satisfied in all 30 missions. Considering the fact that reaching the destination is a big concern for AUV's safety, a big penalty value is assigned to strictly prevent operations from taking longer than $T_\nabla$.

Fig 7.8 provides a visualized dynamic mission planning process in one experiment in which rearrangement of the tasks/stations order due to the updated $T_\nabla$ and terrain deformations is shown through the multiple mission re-planning procedures. The summary of the outcomes in Fig 7.8 is provided by Table 7.1.

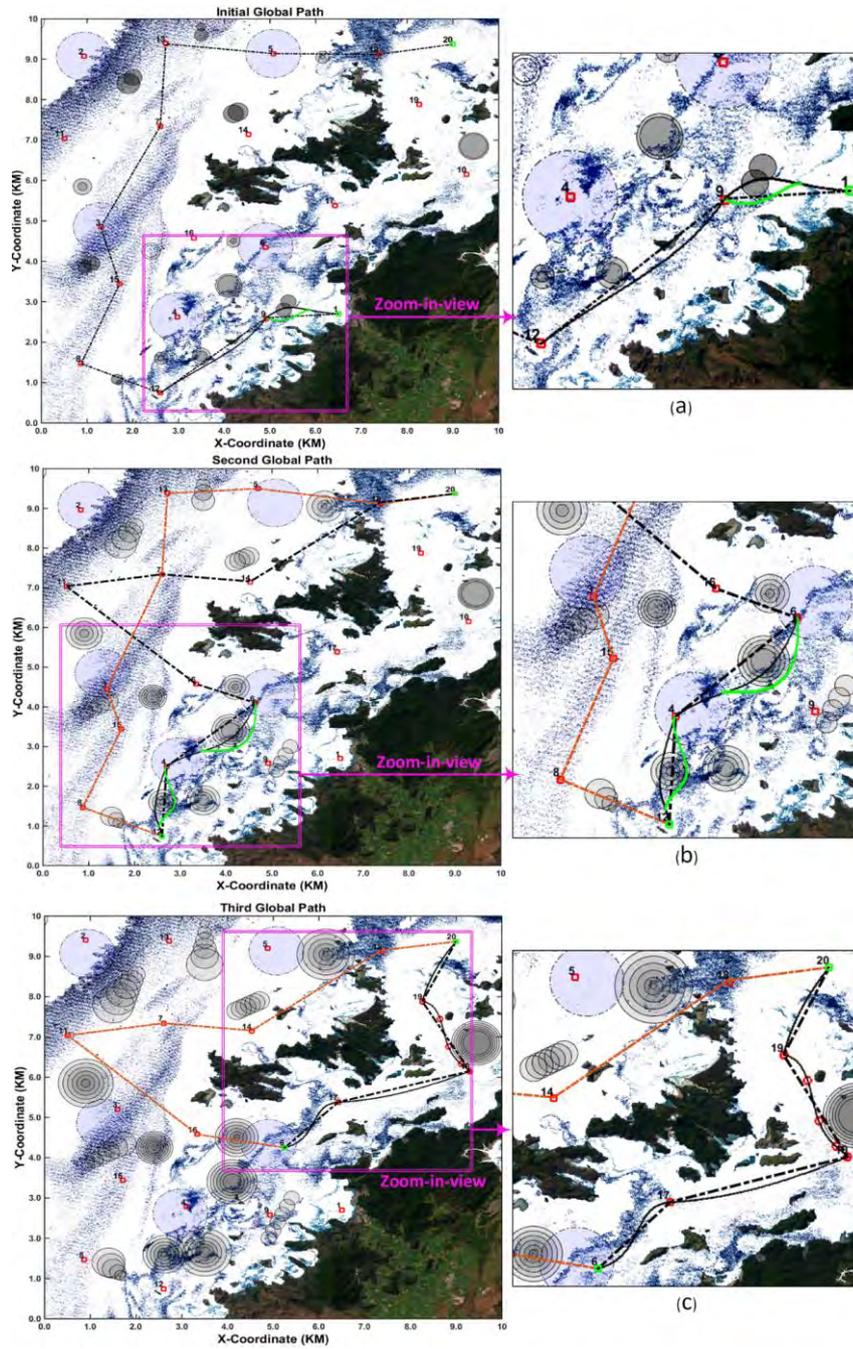

**Fig. 7.8.** The motion re-planning and waypoints rearrangement based on updated $T_{Remained}$ and changes of the terrain.

**Table 7.1.** The performance the framework in a single mission depicted in Fig 7.8.

| Mission | $T_\Re$ (sec) | $T_\nabla$ (sec) | # Nodes | Total $C_\nabla$ | CPU Time (sec) |
|---|---|---|---|---|---|
| **First** | 13983 | 14,400 | 11 | 1.093 | 283 |
| **Second** | 11,002 | 11,347 | 9 | 1.149 | 240 |
| **Third** | 658 | 856 | 5 | 1.071 | 183 |
| **Resultant Route** | 14,187 | 14,400 | 9 | 1.031 | 266 |

Referring to Fig 7.8, the initial order of waypoints in the first mission plan includes 11 stations and takes $T_\Re$=13,983($sec$) duration, which is less than $T_\nabla$=14,400 ($sec$). In the second step, the motion planner in the inner layer tends to find a collision-free trajectory between the waypoints. Referring to Fig 7.8 (a), it can be seen that the motion planner refines the path between waypoints {1-9} to avoid crossing the edges of a buoyant obstacle, which cause a time dissipation. Whenever the motion planner completes its execution, the network deformation is considered, the path time of $T_\wp$ gets compared to the expected time of $T_\varepsilon$, and then $T_\wp$ is reduced from the total existent time of $T_\nabla$. When the $T_\wp^{ij}$ takes bigger value than the $T_\varepsilon^{ij} \approx t_{ij}$, the mission re-planning flag is triggered to compensate the time dissipation, which is the case in the second pair of waypoints {$i$=9, $j$=12} in the initial mission. Therefore, the mission planner module is recalled to generate new efficient set of tasks/stations fitted to the updated $T_\nabla$. The second mission plan in Fig 7.8(b) is commenced from waypoint {12} and comprises overall 9 stations of {12-4-6-16-11-7-14-18-20}. Referring to the given information in Table 7.1, the second route takes $T_\Re$=11,002 ($sec$), which is again less than the updated $T_\nabla$=11,347 ($sec$). Likewise, the waypoint sequence is passed to the inner layer and the process of local motion planning is replicated until the $T_\wp^{ij}$ exceeds the $T_\varepsilon^{ij}$. The process of the motion planner gets completed at waypoint {4}; however, due to having a delay on vehicles travel between {6} to {16} the mission re-planning would be required. Thus, another waypoint sequence is re-planned by the motion planner to cover the loos of time, which is shown in Fig 7.8(c).

Ultimately, the third mission involves 5 waypoints {6-17-10-19-20}. As shown in the zoom-in-view of Fig7.8 (c), the motion planner accurately avoids colliding obstacles. It operates continuously without any delay and the vehicle reaches on-time to the target waypoint {20} by remaining time of 213 ($sec$). The purple circles around some of the stations in Fig 7.8 represent the uncertainty over the exact location of some of the waypoints and the black dashed lines denote the current valid route (waypoint sequence) while the orange dashed line refers to the discarded route. Likewise, the black thick lines represent the local paths generated by the inner layer, and the green thick lines are the refined paths. The transparent black circles around the obstacles represent the collision boundaries with the confidence of 98% that the object is located within this area with an uncertainty propagation (presented by gradual increment of the circles). The overall mission involves 3 mission and 3 motion re-planning process. No collision occurred over the operation. The mission is completed successfully with no delay (with remaining time of $T_\nabla$=14400-14187=213 ($sec$)), and the final resultant route involves 9 stations of {1-9-12-4-6-17-10-19-20}. The given data in Table 7.1 declares the feasibility of the resultant route, while it is satisfying the time constraint and the total time is efficiently used to serve maximum possible tasks/waypoints.

The results of quantitative simulations support that the proposed architecture is reliable and robust, particularly in dealing with uncertainties.

## 7.4  Summary of Chapter

After having discussed on different aspects of mission and motion planning in the previous chapters, this chapter introduced a synchronous augmented architecture to provide a comprehensive control on mission time management and performing beneficent mission with the best sequence of tasks fitted to available time, while safe deployment is guaranteed at all stages of the mission. The system incorporates two different execution layers, deliberative and reactive, to satisfy the AUV's high and low level autonomy requirements, in which two previously developed mission and motion planning modules are integrated to operate in a synchronous and concurrent manner in the context of ARMPA. In this respect, the mission planning module is in charge of prioritizing the available tasks in a way that the selected waypoints in a graph-like terrain direct the vehicle toward destination in a predefined restricted time. Subsequently, in the inner layer the motion planner handles the vehicle's safe motion along the selected direction, where the severe environmental disturbances and uncertainties constantly perturb the vehicle's deployment. Precise and concurrent synchronization of the higher and lower level modules is the critical requirement to preserve the accretion and cohesion of the system, which is provided through designing of the 'Synchron module'. This module recognizes whether to carry out the mission rescheduling or motion re-planning.

A quantitative analysis is provided in order to examine the capability and efficiency of the proposed framework performed in MATLAB® 2016b. Synchronization between higher and lower level modules is effi-ciently configured to manage the mission time and to guarantee on-time termination of the mission. Real map data and different uncertain buoyant/static objects are encountered to model a realistic marine environment. Moreover, the impact of the time variant water current on the AUV's motion is incorporated. Ultimately, performance and stability of the ARMSP architecture in maximizing mission productivity, real-time operation, mission timing, and handling dynamicity of the terrain was evaluated using DE algorithm. The obtained quantitative and qualitative results from different simulated missions indicate model's consistency and effectiveness in providing safe and beneficial operation and confirm system's robustness against the problem space deformation. Knowing the fact that existing approaches are only able to cover a part of problem of ether motion/path planning or task assignment based on offline map, the provided framework is a remarkable contribution to improve an AUV's autonomy due to its comprehensive mission-motion planning capability that covers shortcomings associated with previous strategies in this scope.

Developing such a modular architecture is advantageous as it can be upgraded or modified in an easier way that promotes reusability of the system to be implemented on other robotic platforms such as ground or aerial vehicles. In this way, the architecture also can be developed by adding new modules to handle vehicles inner situations, fault tolerant or other consideration toward increasing vehicle's general autonomy.